\documentclass[11pt]{article}
\usepackage[T1]{fontenc}
\usepackage{lmodern}
\usepackage[letterpaper,margin=1in]{geometry}
\usepackage{microtype}
\usepackage{booktabs,tabularx,array}
\usepackage{enumitem}
\usepackage{placeins,needspace}
\usepackage{xcolor}
\usepackage{colortbl}
\usepackage{graphicx}
\usepackage{natbib}
\AtBeginDocument{\setcitestyle{round,semicolon}}
\usepackage{amsmath,amssymb,amsthm,mathtools,mathrsfs}
\usepackage{algorithm}
\usepackage{algpseudocode}
\usepackage{authblk}

\usepackage{hyperref}
\hypersetup{
  colorlinks=true,
  allcolors=blue!60!black,
  pdfborder={0 0 0},
  pdftitle={Lower Bounds for Smooth Nonconvex Zeroth-Order Optimization: The Price of Determinism under Low Curvature Trace},
  pdfauthor={Yuxing Peng, Zhiqing Tang, Weijia Jia},
  pdfkeywords={zeroth-order optimization, lower bounds, exact function values,
    derivative-free optimization, oracle complexity, randomization, curvature trace,
    effective rank, nonconvex optimization}
}
\usepackage[capitalize,noabbrev]{cleveref}
\newif\ifphaseinappendix\phaseinappendixfalse
\newcommand{\phasefigwidth}{0.62\textwidth}
\newcommand{\leanavailability}{The formalization is available at
  \url{https://github.com/Pyx9876/zo-lower-bounds-lean} and in the ancillary
  files of this arXiv entry.}
\newtheorem{theorem}{Theorem}[section]
\newtheorem{lemma}[theorem]{Lemma}
\newtheorem{proposition}[theorem]{Proposition}
\newtheorem{corollary}[theorem]{Corollary}
\theoremstyle{definition}

\newcommand{\R}{\mathbb R}

\newcommand{\E}{\mathbb E}
\newcommand{\Prob}{\mathbb P}
\newcommand{\eps}{\varepsilon}
\newcommand{\norm}[1]{\left\lVert#1\right\rVert}
\newcommand{\ip}[2]{\left\langle#1,#2\right\rangle}
\newcommand{\Lip}{\operatorname{Lip}}

\newcommand{\cF}{\mathcal F}
\newcommand{\cP}{\mathcal P}
\providecommand{\cA}{\mathcal A}

\providecommand{\tr}{\operatorname{tr}}

\newcommand{\cB}{\mathcal B}

\newcolumntype{Y}{>{\raggedright\arraybackslash}X}

\title{Lower Bounds for Smooth Nonconvex Zeroth-Order Optimization:\\
The Price of Determinism under Low Curvature Trace}
\author[1]{Yuxing Peng}
\author[2,*]{Zhiqing Tang}
\author[2,3]{Weijia Jia}
\affil[1]{Faculty of Arts and Sciences, Beijing Normal University, Zhuhai 519087, China}
\affil[2]{Institute of Artificial Intelligence and Future Networks, Faculty of Arts and Sciences, Beijing Normal University, Zhuhai 519087, China}
\affil[3]{Guangdong Key Lab of AI and Multi-Modal Data Processing, Beijing Normal-Hong Kong Baptist University, Zhuhai 519087, China}
\affil[*]{Corresponding author: \texttt{zhiqingtang@bnu.edu.cn}}
\date{}

\begin{document}
\maketitle

\begin{abstract}
Low curvature trace, an effective rank condition, is a common explanation for
why randomized zeroth-order methods scale to language model fine-tuning. Can
deterministic methods achieve the same savings? We study finding an
$\varepsilon$-stationary point of an $L$-smooth nonconvex function $f$ on
$\mathbb{R}^d$ from exact function values when, on the set where $f$ is below
its initial value $f(0)$, the Hessian is dominated by a fixed matrix of trace at
most $LE$. For an initial
gap $\Delta$, let $T=L\Delta/\varepsilon^2$. We show that the Gaussian two-point
method used in fine-tuning needs only $O(ET)$ calls if it accepts only the steps
that decrease $f$, and we prove that every randomized method requires at least
$\Omega(ET)$ calls when $ET\le d$, so this cost is optimal up to constants. In
that range, every deterministic method requires at least
$\widetilde\Omega(d\min\{T,\sqrt{ET}\})$ calls, and adaptive subspace search
attains $O(d\min\{T,\sqrt{ET}\})$ calls. A deterministic method pays about $d$ calls
for each direction it discovers, and the trace budget allows about $\sqrt{ET}$
directions. Determinism therefore multiplies the cost by a factor of order
$d/(E+\sqrt{ET})$, up to logarithmic factors. Once $E$ exceeds $T$ by a
logarithmic factor, deterministic methods require $\Omega(dT)$ calls, just as they
do without any trace budget. Our lower bounds hold for $C^\infty$ functions that satisfy
the matrix bound on all of $\mathbb{R}^d$, even when each call also returns
projections of the gradient, charged by their rank.

\end{abstract}

\section{Introduction}
\label{sec:introduction}

In zeroth-order optimization, an algorithm optimizes an objective function with
access only to evaluations of its value. Such methods are the tool of choice
whenever gradients are unavailable or too costly to compute, as when attacking
or explaining a model only through its outputs \citep{chen2017zoo}, tuning
simulations, hyperparameters, and engineering designs
\citep{larson2019derivative}, or searching over policies in reinforcement
learning \citep{salimans2017evolution}; see \citet{liu2020primer} for a survey.
Large language models have brought renewed interest to this setting. Because
fine-tuning with forward passes alone requires about the same memory as
inference, zeroth-order methods allow a single GPU to tune models that
backpropagation cannot fit in memory
\citep{malladi2023mezo,zhang2024revisiting}, while seed-based zeroth-order
updates reduce communication in federated fine-tuning \citep{qin2024federated}.

\begin{table*}[t]
\caption{Query complexity of finding an $\eps$-stationary point and the
algorithms that attain the upper bounds. D and R denote deterministic and
randomized algorithms, NC means nonconvex, and $T=L\Delta/\eps^2$. All results
use exact function values unless the setting names another oracle.
\textbf{Bold entries are new in this paper.} ``Matched'': whether the two
bounds agree up to constant factors (yes) or up to the stated factor; a dash
means that only one side is known.}
\label{tab:related}
\centering
\footnotesize
\renewcommand{\arraystretch}{1.15}
\setlength{\tabcolsep}{4pt}
\begin{tabularx}{\textwidth}{>{\raggedright\arraybackslash}p{0.3\textwidth}>{\centering\arraybackslash}p{0.05\textwidth}>{\raggedright\arraybackslash}p{0.255\textwidth}Y>{\centering\arraybackslash}p{0.07\textwidth}}
\toprule
Setting & Alg. & Bounds & Upper bound attained by & \makebox[\linewidth][c]{Matched}\\
\midrule
\multicolumn{5}{@{}l}{\emph{Low curvature trace: $\nabla^2f\preceq H$ with $\operatorname{tr}H\le LE$ below $f(0)$ (this work)}}\\
\rowcolor{black!6}$ET\le d$, Theorem~\ref{thm:main-random} & R & $\boldsymbol{\Omega(ET)}$, $\boldsymbol{O(ET)}$ & \textbf{Gaussian two-point with an acceptance test}\,$^{a}$ & yes\\
$ET\le d$, Theorem~\ref{thm:main-det} & D & $\boldsymbol{\widetilde\Omega(d\min\{T,\sqrt{ET}\})}$, $\boldsymbol{O(d\min\{T,\sqrt{ET}\})}$ & \textbf{restarted subspace search}, adapting \citet{cosson2022lowrank} & up to $\sqrt{\log}$\,$^{b}$\\
\rowcolor{black!6}$T\log(e+T)\le E\le d/T$, Corollary~\ref{cor:main-window} & D vs R & \textbf{D: $\boldsymbol{\Omega(dT)}$; R: $\boldsymbol{\Omega(ET)}$}; D: $O(dT)$; R: $O(ET)$ & coordinate finite differences; Gaussian two-point & yes\\
\midrule
\multicolumn{5}{@{}l}{\emph{Smooth NC, no trace budget}}\\
\rowcolor{black!6}\citet{xia2026global} & R & $\Omega(dT)$\,$^{\S}$, $O(dT)$ & Gaussian two-point \citep{nesterov2017random} & yes\\
$1\le T\le d/\log(ed)$, Theorem~\ref{thm:unrestricted} & D & $\boldsymbol{\Omega(dT)}$\,$^{c}$, $O(dT)$ & coordinate finite differences & yes\\
\midrule
\multicolumn{5}{@{}l}{\emph{Other settings}}\\
\rowcolor{black!6}Effective rank \citep{malladi2023mezo,yue2023beyond} & R & $d$ replaced by an effective dimension & random finite differences & --\\
$\|\nabla^2f\|_*\le L_*$ \citep{anonymous2026tracenorm} & R & $O(L_*\Delta/\eps^2)$ & Gaussian two-point & --\\
\rowcolor{black!6}First-order oracle, large $d$ \citep{carmon2020lower} & D, R & $\Omega(T)$, $O(T)$ & gradient descent & yes\\
Lipschitz, local oracle \citep{jordan2023deterministic,kornowski2026exponential} & D vs R & Goldstein points: D pays $(1/\eps)^{\Omega(d)}$ & -- & --\\
\bottomrule
\end{tabularx}
\par\vskip 0.04in
\raggedright{\footnotesize \mbox{$^{\S}$\,For $T\ge c$ and $d\ge CT\log(2+T)$, with absolute constants $c,C>0$.}\quad
\mbox{$^{a}$\,Theorem~\ref{b:thm:monotone}; without the test, $O(ET+T\log(ET))$ (Theorem~\ref{g2:thm:sublevel}).}\quad
\mbox{$^{b}$\,$\widetilde\Omega$ hides only a factor $\sqrt{\log(e+T)}$.}\quad
\mbox{$^{c}$\,At least $dT/500$, also with gradient projections charged by rank.}\par}
\vskip -0.1in
\end{table*}

For an $L$-smooth nonconvex function $f:\R^d\to\R$ with initial gap
$f(0)-\inf f\le\Delta$, random finite differences find an $\eps$-stationary
point, a point $\widehat x$ with $\|\nabla f(\widehat x)\|\le\eps$, and the
classical analysis bounds their cost by $O(dT)$ function values, where
$T=L\Delta/\eps^2$ \citep{nesterov2017random,ghadimi2013stochastic}. In high
dimension, this rate cannot be improved on the full smooth class even with
exact function values. Concurrent and independent work of \citet{xia2026global}
proves that randomized algorithms, and hence deterministic ones as well,
require at least $\Omega(dT)$ exact value calls when $d\gtrsim T\log T$. For deterministic algorithms, we prove the
explicit lower bound $dT/500$ for $1\le T\le d/\log(ed)$, which matches
coordinate finite differences up to constants
(Theorem~\ref{thm:unrestricted}). In high-dimensional tasks such as language model fine-tuning, however, randomized
zeroth-order methods succeed at far lower query cost. Effective rank analyses
attribute this speedup to a low curvature trace
\citep{malladi2023mezo,yue2023beyond}, consistent with the low intrinsic
dimension of fine-tuning \citep{li2018intrinsic,aghajanyan2021intrinsic}. Does
this saving come from problem structure alone, or does it require randomness?

This question also applies to practical implementations of randomized methods.
Memory-efficient zeroth-order fine-tuning regenerates each perturbation from a
stored random seed \citep{malladi2023mezo}. Once its seeds are fixed in
advance, such a procedure is a deterministic algorithm, so deterministic lower
bounds apply to it on functions chosen after the seeds are set; we defeat such
a fixed-seed run experimentally in Section~\ref{sec:empirical}. Classical
derivative-free optimizers, including coordinate and pattern search
\citep{hooke1961direct}, the Nelder--Mead method \citep{nelder1965simplex}, and
Powell's interpolation methods \citep{conn2009introduction,larson2019derivative},
are deterministic as well. This raises the question of whether a deterministic
algorithm can learn the active subspace of curvature along its trajectory and
search within it, paying only for the curvature trace rather than for the
ambient dimension $d$. Whereas \citet{jordan2023deterministic} and
\citet{kornowski2026exponential} establish a price of determinism for nonsmooth
Goldstein stationarity (Section~\ref{sec:related}), we study smooth first-order
stationarity in high dimension and show that randomness is essential to fully
exploit a low curvature trace.

To compare deterministic and randomized methods on the same functions, we
consider an $L$-smooth objective $f$ with initial gap $\Delta$ whose positive
curvature has bounded trace on the initial sublevel set
$\{x\in\R^d:f(x)<f(0)\}$, where descent methods operate. For almost every $x$
with $f(x)<f(0)$, we assume that the Hessian $\nabla^2f(x)$ is dominated by one
fixed positive semidefinite matrix of operator norm at most $L$ and trace at
most $LE$. Because $E$ bounds only the trace of this dominating matrix,
positive curvature may spread across many directions. Since the matrix bounds
the Hessian only from above, negative curvature is not charged at all. Section~\ref{sec:compare} compares this class with prior
curvature assumptions. \citet{malladi2023mezo} assume such a dominating
matrix near each iterate of a trajectory and derive an effective-rank upper
bound for zeroth-order SGD, whereas our condition is fixed on the initial
sublevel set before the run begins.

Suppose $1\le E\le d$, $T\ge1$, and
$ET\le d$, and let $N_{\rm det}$ and $N_{\rm rand}$ denote the deterministic
and randomized worst-case query complexities on this class
(Section~\ref{sec:model}). Randomized methods pay only for the curvature trace.
We show that the Gaussian two-point method
attains $O(ET)$ calls when it accepts only steps that decrease $f$, and that
every randomized method requires at least $\Omega(ET)$ calls. A deterministic
method, by contrast, pays roughly $d$ calls for each active direction it
discovers, while the trace budget allows at most $O(\sqrt{ET})$ such
directions.
We show that adaptive subspace search attains $O(d\min\{T,\sqrt{ET}\})$ calls,
while our hidden-chain construction forces $\widetilde\Omega(\min\{T,\sqrt{ET}\})$
directions. We prove (Theorems~\ref{thm:main-random} and~\ref{thm:main-det})
\[
\boxed{\begin{aligned}
 \widetilde\Omega\bigl(d\min\{T,\sqrt{ET}\}\bigr)
 &\le N_{\rm det}\le O\bigl(d\min\{T,\sqrt{ET}\}\bigr),\\[2pt]
 \Omega(ET)&\le N_{\rm rand}\le O(ET),
\end{aligned}}
\]
where $\widetilde\Omega$ hides only a factor $\sqrt{\log(e+T)}$; Section~\ref{sec:results}
states the bounds with this factor written out. For $ET\le d$, the price of
determinism $N_{\rm det}/N_{\rm rand}$ is therefore bounded below by
$\widetilde\Omega(d/(E+\sqrt{ET}))$ and above by $O(d/(E+\sqrt{ET}))$, growing
linearly with $d$ when $E$ and $T$ are fixed. Once $E\ge T\log(e+T)$ within
this range, so that $T\log(e+T)\le E\le d/T$, deterministic methods require at
least $\Omega(dT)$ calls, just as without a trace budget. Our upper and lower
bounds then match up to constant factors for both algorithm classes, and their
costs differ by a factor of order $d/E$ (Corollary~\ref{cor:main-window}).
Table~\ref{tab:related} compares these bounds with the closest prior ones.

Why can exact real replies still force $\Omega(d)$ calls per hidden direction?
When each reply carries a bounded number of bits, lower bounds follow by
counting the revealed information \citep{basu2025binary}. An exact real reply,
however, can encode arbitrarily much discrete information, so counting returned
scalars gives no lower bound. Against a deterministic algorithm, we instead
combine the gated chains of \citet{carmon2020lower} with a resisting oracle
\citep{nemirovski1983problem}, so that the replies agree \emph{exactly} with those of a
single function in the class along the adaptive transcript. The oracle fixes each hidden direction only when its
stage closes, after $\Omega(d)$ units of cost, and keeps an unfinished direction
orthogonal to the output, even if the output was never queried. The resulting
function produces every reply of the run, and its gradient at the output has norm
above $\eps$. Multiplying the chain by a perspective
weight rules out small gradients far from the origin and keeps the Hessian
dominated by one fixed matrix on all of $\R^d$ (Section~\ref{sec:techniques}).
Against randomized algorithms, the function must be chosen before the seed,
and Yao's principle \citep{yao1977probabilistic} applied to a quadratic with an
unknown center gives $\Omega(ET)$.

\paragraph{Contributions.}
\begin{itemize}[leftmargin=1.2em,itemsep=1pt,topsep=2pt]
\item \textbf{Randomized complexity scales with the trace rather than the dimension.}
We show that for all $1\le E\le d$ and $T\ge1$, the Gaussian two-point method
of \citet{nesterov2017random} attains $O(ET)$ calls when it accepts only steps
that decrease $f$. Without this acceptance test, we prove that it attains
$O(ET)$ calls when the matrix bound holds on all of $\R^d$ and
$O(ET+T\log(ET))$ calls on our sublevel-set class. Conversely, we prove that
when $ET\le d$, every randomized algorithm requires at least $\Omega(ET)$ calls
(Theorem~\ref{thm:main-random}).
\item \textbf{Quantifying the deterministic cost under a low curvature trace.}
For $ET\le d$, we adapt the subspace search scheme of \citet{cosson2022lowrank}
to find an $\eps$-stationary point in $O(d\min\{T,\sqrt{ET}\})$ calls, and we
construct $C^\infty$ hard functions satisfying the matrix bound on all of $\R^d$
to prove that every deterministic algorithm requires at least
$\widetilde\Omega(d\min\{T,\sqrt{ET}\})$ calls (Theorem~\ref{thm:main-det}).
Across $T\log(e+T)\le E\le d/T$, we show that deterministic algorithms require
at least $\Omega(dT)$ calls, losing any savings from the trace budget, so
determinism multiplies the worst-case query cost by a factor of order $d/E$
(Corollary~\ref{cor:main-window}).
\item \textbf{Optimality of coordinate finite differences without a trace budget.}
On the unrestricted $L$-smooth class, we prove that for
$1\le T\le d/\log(ed)$, every deterministic algorithm requires at least
$dT/500$ calls, even when each call may also return projections of the
gradient, charged by their rank (Theorem~\ref{thm:unrestricted}). For exact
function values alone, the randomized lower bound of \citet{xia2026global}
already implies a deterministic $\Omega(dT)$ bound when $d\gtrsim T\log T$; our
result adds an explicit constant and covers these gradient projections,
which shows that coordinate finite differences are
optimal up to constants under both oracles.
\end{itemize}

\section{Setting}
\label{sec:model}

\paragraph{Function classes.}
For $L,\Delta>0$, we consider the smooth nonconvex class
\begin{equation}
\begin{aligned}
 \cF^d_{L,\Delta}:=\bigl\{f\in C^{1,1}(\R^d):\ &\Lip(\nabla f)\le L,\\
 &f(0)-\inf f\le\Delta\bigr\}.
\end{aligned}
 \label{eq:function-class}
\end{equation}
The goal is a point $\widehat x\in\R^d$ with $\norm{\nabla f(\widehat x)}\le\eps$,
and we write $T:=L\Delta/\eps^2$. For $1\le E\le d$, the main class
$\cB^d_{L,\Delta,E}$ consists of the functions $f\in\cF^d_{L,\Delta}$ for which
some matrix $H\succeq0$, unknown to the algorithm, with $\norm H_{\rm op}\le L$ and
$\operatorname{tr}H\le LE$ satisfies
\begin{equation}
 \nabla^2f(x)\preceq H
 \qquad\text{for almost every $x$ with $f(x)<f(0)$}.
 \label{eq:main-class}
\end{equation}
Matrix bounds of this form, with $\operatorname{tr}H/\norm H_{\rm op}$ called
an effective rank, are used to explain why zeroth-order fine-tuning is cheap
\citep{malladi2023mezo}. We impose \eqref{eq:main-class} only on the initial
sublevel set $U=\{x:f(x)<f(0)\}$, because this is where descent methods
operate. Above $f(0)$ the class asks only for $L$-smoothness, and since $H$
bounds the Hessian only from above, negative curvature is not charged, so
functions in the class may be strongly nonconvex.

The trace of $H$ bounds the total positive curvature by $LE$, so $E$ plays the
role of a dimension: $E=d$ adds nothing to $L$-smoothness, whereas a small $E$
forces the curvature to concentrate in a few directions or to spread thinly
over many. A single matrix for the whole sublevel set is a structure that a
deterministic method can learn and reuse (Section~\ref{sec:techniques}),
whereas randomized methods need only a bound at each point.

Our lower bounds hold on the smaller global fixed-matrix class
$\cF^d_{L,\Delta,E}\subseteq\cB^d_{L,\Delta,E}$
(Lemma~\ref{p:lem:inclusions}), which requires $\nabla^2f(x)\preceq H$ for
almost every $x\in\R^d$. For $C^{1,1}$ functions, this is the quadratic upper
model
\begin{equation}
 f(x+v)\le f(x)+\ip{\nabla f(x)}{v}+\tfrac12v^\top Hv
 \qquad(x,v\in\R^d),
 \label{eq:trace-class}
\end{equation}
and we call $H$ an upper model of $f$. For $E=d$, every function in
$\cF^d_{L,\Delta}$ belongs to $\cF^d_{L,\Delta,E}$. Our randomized upper bound,
on the other hand, needs neither a common matrix nor smoothness above $f(0)$:
it holds on the local pointwise class of Appendix~\ref{app:classes}, which
asks for a locally $L$-Lipschitz gradient on $U$ and for the positive
eigenvalues of $\nabla^2f(x)$ to sum to at most $LE$ at almost every point of
$U$.

\paragraph{Oracles and costs.}
At each call $t$, the algorithm chooses a point $x_t\in\R^d$ based on the
earlier replies, and the oracle returns the exact value $f(x_t)$. The output
$\widehat x$ need not be one of the queried points. The deterministic
complexity $N_{\rm det}$ is the smallest worst-case number of calls of a
deterministic algorithm that succeeds on every function in the class. A
randomized algorithm must succeed with probability at least $2/3$ on
\emph{each} function while making at most $N$ calls on every run (a cost cap
of $N$), and $N_{\rm rand}$ is the smallest such cap. Algorithms know $d$,
$L$, $\Delta$, $\eps$, and $E$; computation is free and arithmetic is exact, and
randomized policies are jointly measurable in their seed and the transcript
(Appendix~\ref{a:sec:random}).

All our lower bounds also hold for a stronger oracle that additionally
returns projections of the gradient and charges for their rank: at call $t$,
the algorithm may choose a matrix $V_t\in\R^{d\times s_t}$ and receive
\begin{equation}
 \mathcal O_f(x_t,V_t)=\bigl(f(x_t),V_t^\top\nabla f(x_t)\bigr)
 \quad\text{at cost }1+\operatorname{rank}V_t.
 \label{eq:linear-oracle}
\end{equation}
A rank-one call returns, besides $f(x_t)$, a forward-mode directional
derivative \citep{baydin2022gradients}, and $V_t=I_d$ gives the full gradient. We write
$N^{\rm val}$ and $N^{\rm lin}$ when the oracle matters
($N^{\rm lin}\le N^{\rm val}$). The lower bounds also hold for the worst-case
expected cost.

\subsection{Comparison with prior curvature assumptions}
\label{sec:compare}
How does our condition relate to the conditions under which earlier
zeroth-order methods improve their dependence on $d$?
\begin{itemize}[leftmargin=1.2em,itemsep=1pt,topsep=2pt]
\item \textbf{Sparse and latent structure} \citep{wang2018stochastic,golovin2020gradientless}.
\citet{wang2018stochastic} assume sparsity, and \citet{golovin2020gradientless}
study objectives with a low-dimensional latent structure. A globally
$L$-smooth objective with $f(0)-\inf f\le\Delta$ that depends only on a fixed $k$-dimensional subspace
lies in $\cF^d_{L,\Delta,k}$, and our trace bound also allows Hessians of
full rank.
\item \textbf{Local effective rank} \citep{malladi2023mezo}. \citet{malladi2023mezo}
assume that near each iterate $x_t$ of a run, $\nabla^2f\preceq H_t$ with
$\operatorname{tr}H_t\le r\norm{H_t}_{\rm op}$. Letting $H_t$ change along the
run is more general than our single $H$, but the condition then depends on the
algorithm, whereas a lower bound needs a class that is fixed before the run.
Every $f\in\cF^d_{L,\Delta,E}$ satisfies their condition with $r=E+1$: take
$H_t=H+(L-\norm H_{\rm op})vv^\top$ for a top unit eigenvector $v$ of $H$, so
that $\norm{H_t}_{\rm op}=L$ and $\operatorname{tr}H_t\le L(E+1)$. Our lower
bounds therefore apply in their setting.
\item \textbf{Effective dimension and nuclear norm}
\citep{yue2023beyond,anonymous2026tracenorm}. These conditions sum powers of
the absolute Hessian eigenvalues, or bound $\norm{\nabla^2f}_*\le L_*$, at
every point of $\R^d$. Our condition is weaker in that it charges only
positive curvature and only below $f(0)$, so a strongly nonconvex function can
have a small $E$ but large values of these measures. It is stronger in that it
asks for one common matrix, the structure that deterministic subspace search
exploits.
\item \textbf{Parameter symmetries} \citep{anonymous2026symmetry}. \citet{anonymous2026symmetry}
derive positive trace bounds from symmetries and analyze zeroth-order SGD
under a Polyak--{\L}ojasiewicz condition; we assume neither symmetries nor that
condition.
\end{itemize}
Our randomized upper bound holds on the local pointwise class of
Appendix~\ref{app:classes}, which for $L=L_*$ and $E=1$ contains the nuclear
norm class of \citet{anonymous2026tracenorm}, and our lower bounds hold on
functions of $\cF^d_{L,\Delta,E}$ with Hessian nuclear norm at most $LE$.
The separation therefore holds on every class between these hard functions and
the local pointwise class, including that nuclear norm class.

\section{Main Results}
\label{sec:results}

We now state our bounds. Throughout, $d\ge1$, $1\le E\le d$, and $T\ge1$, and
$c,C>0$ denote absolute constants. Unless another class is named, costs refer
to the main class $\cB^d_{L,\Delta,E}$, and all bounds hold for both oracles.
We prove all lower bounds on the global fixed-matrix class
$\cF^d_{L,\Delta,E}\subseteq\cB^d_{L,\Delta,E}$
(Appendix~\ref{gl:sec:global}), and the randomized upper bound holds even
under the pointwise trace bound of Appendix~\ref{app:classes}.

\paragraph{1. Randomized complexity.}
\begin{theorem}[Randomized cost]
\label{thm:main-random}
For all the above parameters, $N_{\rm rand}\le C\,ET$. If $ET\le d$, then
$N_{\rm rand}\ge c\,ET$.
\end{theorem}
Random directions replace the dimension $d$ by the trace budget $E$. The upper
bound is attained by the Gaussian two-point method \citep{nesterov2017random},
whose estimator is used in zeroth-order fine-tuning \citep{malladi2023mezo}, with
step size $1/(L(E+2))$. If the method accepts a step only when $f$ decreases,
it uses at most $12ET+26T+3$ calls, still with two function values per step
(Theorem~\ref{b:thm:monotone}). Without this test, it uses at most
$34(E+2)T+2$ calls on $\cF^d_{L,\Delta,E}$ (Theorem~\ref{g2:thm:global}). On
$\cB^d_{L,\Delta,E}$, a long step may leave the sublevel set, so the method
needs a smaller step and uses $O(ET+T\log(ET))$ calls
(Theorem~\ref{g2:thm:sublevel}). The stochastic three-point method
\citep{bergou2020stochastic} attains $O(ET)$ calls under the pointwise trace bound
(Theorem~\ref{x:thm:stp}). Our lower bound holds for every randomized method,
including all these variants, and uses quadratics whose unknown center lies in
a subspace of dimension at most $\min\{d,ET\}$ that is known to the algorithm.

\paragraph{2. The price of determinism.}
\begin{theorem}[Deterministic cost]
\label{thm:main-det}
If $ET\le d$, then
\[
 c\,d\min\Bigl\{T,\sqrt{\tfrac{ET}{\log(e+T)}}\Bigr\}
 \le N_{\rm det}\le
 C\,d\min\bigl\{T,\sqrt{ET}\bigr\}.
\]
\end{theorem}
In $\widetilde\Omega$ notation, the two bounds are
$\widetilde\Omega(d\min\{T,\sqrt{ET}\})$ and $O(d\min\{T,\sqrt{ET}\})$, and
they agree up to a factor $\sqrt{\log(e+T)}$. When $E<T$, adaptive subspace
search beats coordinate finite differences, which use $O(dT)$ calls, because it
pays about $d$ calls for a full gradient estimate only when its current
subspace is exhausted. The common matrix $H$ of $\cB^d_{L,\Delta,E}$ limits
the number of these subspace expansions to $O(\sqrt{ET})$, even though the search receives neither $E$
nor $H$ as input (Section~\ref{sec:techniques}, Algorithm~\ref{alg:subspace},
and Theorem~\ref{u:thm:main}). Our lower bound uses $C^\infty$ hard
functions whose Hessian nuclear norm is at most $LE$ on all of $\R^d$ and whose
Hessian Lipschitz constant is finite (Appendix~\ref{gl:sec:global}).

\begin{corollary}[A window with matching bounds]
\label{cor:main-window}
If $T\log(e+T)\le E\le d/T$, then
\[
 \begin{aligned}
 c\,dT&\le N_{\rm det}\le C\,dT,\\
 c\,ET&\le N_{\rm rand}\le C\,ET.
 \end{aligned}
\]
For the same oracle, the cost ratio is bounded below by $\Omega(d/E)$ and above by
$O(d/E)$.
\end{corollary}
Once $E\ge T\log(e+T)$, a threshold set by the accuracy alone, deterministic
methods lose their savings from the trace and pay $\Omega(dT)$. The window of
Corollary~\ref{cor:main-window} is nonempty when $T^2\log(e+T)\le d$; for $d=10^6$ it allows $T$ up to about
$400$. Over the whole range $ET\le d$, the cost ratio lies between
$\widetilde\Omega(d/(E+\sqrt{ET}))$ and $O(d/(E+\sqrt{ET}))$
(Corollary~\ref{b:cor:aligned}); this ratio is at least
$\widetilde\Omega(\sqrt d)$ whenever $E\le T$. Appendix~\ref{a:sec:main} gives the constants explicitly,
and Theorems~\ref{thm:main-random} and~\ref{thm:main-det} and
Corollary~\ref{cor:main-window} are formalized in Lean~4 on
$\cB^d_{L,\Delta,E}$ (Appendix~\ref{app:lean}).

\paragraph{3. Without a trace budget.}
\begin{theorem}[Unrestricted smooth class]
\label{thm:unrestricted}
On $\cF^d_{L,\Delta}$, if $1\le T\le d/\log(ed)$, then
\[
 \frac{dT}{500}\le N_{\rm det}\le10\,dT.
\]
\end{theorem}
For exact values, the randomized lower bound of \citet{xia2026global} already
gives $\Omega(dT)$ for deterministic methods when $d\gtrsim T\log T$. We add the
explicit constant $1/500$, and our bound also holds when each call may
return gradient projections, charged by their rank
(Theorem~\ref{thm:linear-lower}); for that oracle a randomized $\Omega(dT)$
bound remains open. Coordinate finite
differences are therefore optimal up to constant factors for
$1\le T\le d/\log(ed)$ under both oracles. The hard functions of
Theorem~\ref{thm:unrestricted} also separate deterministic from
randomized methods: a randomized batching method minimizes each of them
exactly with probability at least $5/6$ using $O(d\sqrt{T\log(e+T)})$ function values
(Proposition~\ref{rb:prop:parity}).

\ifphaseinappendix\else\begin{figure}[t]
\centering
\includegraphics[width=\phasefigwidth]{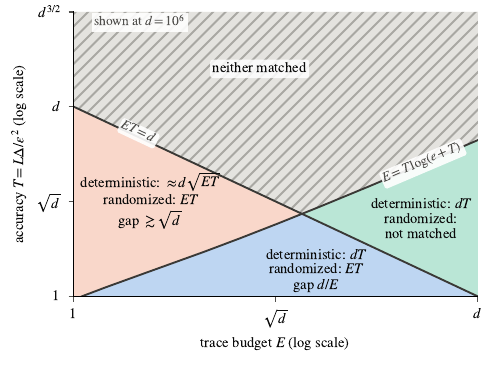}
\vskip -0.08in
\caption{Which of our bounds match? The curves $E=T\log(e+T)$ and $ET=d$ divide
the parameter plane, drawn for $d=10^6$. In each region, the figure gives the
deterministic and randomized costs where our lower and upper bounds match up to
constant factors, or up to $\sqrt{\log(e+T)}$ where marked with $\approx$;
``not matched'' marks regions where our lower and upper bounds differ. Theorem~\ref{a:thm:main} and Corollary~\ref{gl:cor:global-window} give
the conditions.}
\label{fig:phase}
\vskip -0.1in
\end{figure}
\fi

\paragraph{Beyond these ranges.}
Figure~\ref{fig:phase}\ifphaseinappendix{} in Appendix~\ref{a:sec:main}\fi{} maps the parameter plane. For exact values, the lower bound of
\citet{xia2026global} widens the window of Corollary~\ref{cor:main-window} to
$T\log(e+T)\le E\le d$ when $T\le cd/\log(ed)$
(Corollary~\ref{rb:cor:xy-window}).

\section{Proof Mechanisms}
\label{sec:techniques}
We explain the ideas behind our bounds here; Appendix~\ref{app:derivations}
shows how the constructions combine into the results of
Section~\ref{sec:results}.

\subsection{Hiding one direction at a time}
Why must a deterministic method pay $\Omega(d)$ calls for each hidden
direction, even though an exact real reply can carry infinitely many bits? We
separate two parts: an adversary argument, which shows that one unfinished
direction forces $\Omega(d)$ units of cost, and an analytic construction,
which packs as many directions as smoothness, the initial gap, and the trace
budget allow.

The hard function hides orthonormal directions $u_1,u_2,\ldots$, one per stage.
As in the chains of \citet{carmon2020lower}, later directions sit behind scalar
gates that vanish identically near zero, so at a point whose coordinate along
$u_j$ is zero and whose coordinates along later directions are at most $1/2$ in magnitude, the directions after $u_j$ contribute nothing to $f(x)$ or
$\nabla f(x)$. At a point where some stage is unfinished, meaning that the
point's coordinate along that stage's direction has magnitude below a fixed level, the
gradient has norm above $\eps$. A resisting oracle \citep{nemirovski1983problem}
(Lemma~\ref{lem:stage}) chooses each $u_j$ only after the earlier queries are
fixed. While the calls made so far in stage $j$ have total cost below
$\lfloor d/2\rfloor$, any $u_j$ orthogonal to their query points and projection
subspaces leaves every reply equal to that of the prefix function, which
involves only $u_1,\ldots,u_{j-1}$, as long as the later directions stay nearly
orthogonal to these points. On the call that closes stage $j$, the
oracle fixes $u_j$ before returning the reply. Keeping each later direction nearly orthogonal to the
queries of all earlier stages, so that the earlier replies remain valid, costs
the factor $\sqrt{\log(e+T)}$ (Lemma~\ref{a:lem:selection}). If the budget runs
out first, the oracle also chooses an unfinished direction orthogonal to the
output $\widehat x$, even if $\widehat x$ was never queried. One fixed function
in the class then produces every reply of the transcript and has
$\norm{\nabla f(\widehat x)}>\eps$.

\subsection{Fitting the directions into the class}
Hiding $m$ directions forces a cost of $\Omega(dm)$, and after rescaling, each
direction uses up a constant amount of the initial gap. To allow queries
anywhere in $\R^d$, we compose the chain with a radial compression of radius
$R$, which keeps the hidden coordinates bounded, and add a radial potential that
rules out small gradients far from the origin. For the lower bounds on
$\cB^d_{L,\Delta,E}$, the potential is weakly curved inside a ball that contains $\{f<f(0)\}$, and it
grows steeply only outside that ball, where $f(x)>f(0)$. Inside the ball, one fixed matrix,
a multiple of the projection onto the $m$ hidden directions plus a small
multiple of the identity, dominates the Hessian, so the trace budget pays only
for the compressed chain and the weak inner curvature. When
$T\log(e+T)\le E\le d$, every stage that the accuracy allows fits into this
budget. To obtain the lower bounds on $\cF^d_{L,\Delta,E}$, we multiply the chain,
instead of adding a potential, by the perspective weight $(1+\norm x^2/R^2)^{1/2}$
\citep{boyd2004convex}, which cancels the decay of the gradient far away; one
fixed matrix then dominates the Hessian on all of $\R^d$. Smoothing only the
scalar gates, while keeping their flat zero intervals, makes the hard functions
$C^\infty$, and the stage argument still applies (Appendix~\ref{gl:sec:global}).

\subsection{Both bounds count discovered directions}
The number of discovered directions is the bottleneck on both sides. Our
deterministic search, which adapts the scheme of \citet{cosson2022lowrank},
looks for a new unit direction $u_i$ only when the span of the earlier directions is
exhausted, decreases $f$ by more than $\eps^2/(10u_i^\top Hu_i)$ along $u_i$, and restarts with
an empty span after $r$ expansions (Algorithm~\ref{alg:subspace}). Orthonormal
directions within a block satisfy $\sum_iu_i^\top Hu_i\le LE$, so by
Cauchy--Schwarz a full block decreases $f$ by at least
$\Omega(r^2\eps^2/(LE))$, while the total decrease is at most $\Delta$. With
$r$ of order $\sqrt{ET}$, this allows only $O(\sqrt{ET})$ expansions when
$ET\le d$ (Appendix~\ref{a:sec:upper}). Conversely, the hidden chain forces
$\Omega(\min\{T,\sqrt{ET/\log(e+T)}\})$ directions, so the two bounds of
Theorem~\ref{thm:main-det} agree up to $\sqrt{\log(e+T)}$.

\subsection{The randomized lower bound}
Against a randomized algorithm, the hard function cannot be chosen after the
random seed is drawn. By Yao's principle \citep{yao1977probabilistic}, it
suffices to fix, before the run begins, a random quadratic whose unknown center
lies in a subspace known to the algorithm. Each call yields at most as many linear
observations of the center as it costs, and rotational symmetry bounds the
remaining error, even at an output that was never queried. The subspace
has dimension of order $ET$, which fits in $\R^d$ when $ET\le d$, and the
argument then gives the lower bound $\Omega(ET)$.

\subsection{Random directions need curvature control only below $f(0)$}
A Gaussian two-point step moves from $x$ to $x-\eta sa$, where $s$ is the
finite-difference slope of $f$ along $a\sim N(0,I_d)$. With $g=\nabla f(x)$,
the quadratic upper model and the Gaussian moment
$\E[(a^\top g)^2a^\top Ha]=\norm g^2\operatorname{tr}H+2g^\top Hg\le
L(E+2)\norm g^2$ involve the trace of $H$ rather than $d$. With the exact slope $a^\top g$ in place of
$s$, a step with step size $\eta=1/(L(E+2))$ would therefore decrease the quadratic upper
model on average by at least $\norm g^2/(2L(E+2))$; while $\norm g>\eps$, a small
finite-difference radius would keep most of this decrease. A step this
long may land outside the sublevel set, where only $L$-smoothness holds. Even so, the
capped value $\min\{f(x-\eta sa),f(0)\}$ satisfies the same quadratic bound
(Lemma~\ref{u:lem:cap}), so accepting only the steps that lower $f$
suffices (Theorem~\ref{b:thm:monotone}). To analyze the method without this
test, we compare its iterates with an auxiliary run that drops the rare
directions with large $a^\top Ha$; with a slightly smaller step and a small
finite-difference radius, the two runs stay close, at the price of a
$T\log(ET)$ term (Theorem~\ref{g2:thm:sublevel}).

\section{Related Work}
\label{sec:related}

\paragraph{Zeroth-order upper bounds under curvature conditions.}
Classical analyses of Gaussian finite differences reduce derivative-free optimization to stochastic first-order optimization at the cost of a factor of $d$ in the query complexity \citep{nesterov2017random,ghadimi2013stochastic,duchi2015optimal}.
To remove most of this dimension dependence for randomized methods, several works impose structural curvature conditions, which we compare with our setting in Section~\ref{sec:compare}.
These conditions measure curvature differently from our sublevel trace bound.
For example, one of our hard functions $F$, with $m$ hidden directions and compression radius $R$, satisfies $\nabla^2F(0)=\mu I$ for $\mu=11m/R^2$ (Theorem~\ref{w:thm:wrapper}), which gives an effective dimension of at least $d\sqrt{\mu}$ in the sense of \citet{yue2023beyond}, whereas its positive curvature trace is of order $m+\mu d$.
Under a global nuclear-norm bound $\|\nabla^2f\|_*\le L_*$, \citet{anonymous2026tracenorm} show that the Gaussian two-point method finds an $\eps$-stationary point using $O(L_*\Delta/\eps^2)$ exact function values, and they also analyze noisy values at the cost of an extra noise-dependent factor.
Their nuclear-norm class satisfies our pointwise trace bound with $L=L_*$ and $E=1$, and on this class the three-point method attains the same $O(L_*\Delta/\eps^2)$ rate.
Because fixed or neighborhood matrix bounds can require a trace of order $d$ on their Twisted Valley example while the pointwise bound remains constant, we prove our randomized upper bound under that pointwise condition (Appendix~\ref{app:classes}).
Unlike our work, neither \citet{anonymous2026tracenorm} nor \citet{anonymous2026symmetry} proves lower bounds.
For stochastic convex problems on a compact domain, when the subgradients lie in the range of a fixed structural preconditioner, \citet{sholokhov2026poem} use ellipsoidal sampling and obtain bounds that depend on the subspace rank, conditioning, and normalized trace.
Random subspace methods \citep{kozak2021stochastic,omiya2026subspace} exploit related low-dimensional structure under different assumptions, and local gradient smoothness has been studied for both first-order and derivative-free methods \citep{zhang2024firstorder,khanh2023general}.
On the deterministic side, we adapt the subspace expansion scheme of \citet{cosson2022lowrank} from low-dimensional gradient subspaces to a fixed trace budget.
Under higher-order smoothness, \citet{zhou2026finding} develop deterministic value-based methods that find stationary points in any dimension through a trapping-or-descent scheme with query complexity polynomial in $d$, whereas our function classes require only a Lipschitz gradient together with a trace budget.

\begin{figure*}[t]
\centering
\includegraphics[width=\textwidth]{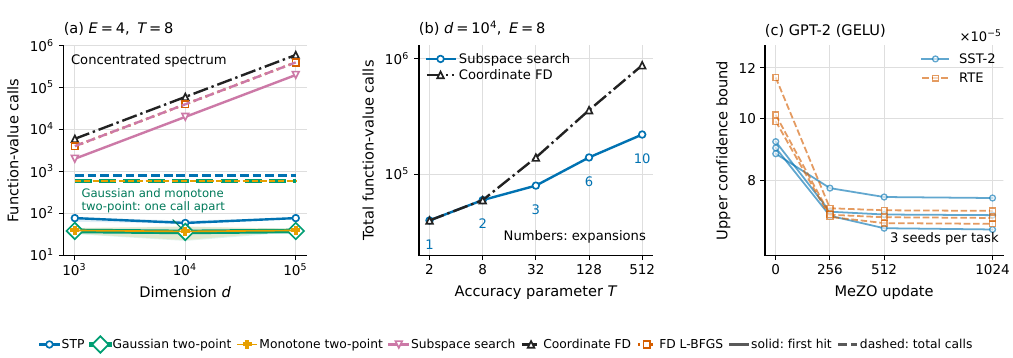}
\caption{Query costs on synthetic functions and local curvature of GPT-2
(GELU). (a) Concentrated-spectrum family with $E=4$ and $T=8$: offline
first-hit calls of the three-point method (STP), the Gaussian two-point
method, the monotone two-point method, the subspace search, coordinate finite differences, and
finite-difference L-BFGS, and actual total calls (dashed).
(b) Actual total calls of the subspace search and coordinate finite differences on a
tridiagonal quadratic chain with $d=10^4$ and $E=8$; labels give
the number of subspace expansions. In (a) and (b), points and bands are medians and
interquartile ranges over five seeds. (c) Upper confidence bounds, at simultaneous level $0.95$, for
$\operatorname{tr}(G_+)/(d\norm G_{\rm op})$, where $G$ is the Hessian on a
fixed batch, during MeZO fine-tuning with three seeds per task. Protocols and further
results are in Appendix~\ref{app:empirical}.}
\label{fig:empirical-mechanism}
\end{figure*}

\paragraph{Lower bounds for stationarity and for exact values.}
Query lower bounds for finding stationary points in a fixed dimension on a box date back to \citet{vavasis1993black}, and \citet{bubeck2020trap} extended them to randomized algorithms in $d=2$.
On $\R^2$, \citet{hollender2024complexity} show that deterministic algorithms require at least $\Omega(\sqrt{T})$ function-value or gradient calls for stationarity, and that $O(\sqrt{T})$ function-value calls suffice.
In high dimension, first-order $\Omega(T)$ lower bounds build on the hidden chains of \citet{carmon2020lower}, which also yield stochastic lower bounds \citep{arjevani2023lower}.
When function values are noisy, the query complexity of convex optimization grows with the dimension $d$ \citep{jamieson2012query,shamir2013complexity,duchi2015optimal}.
Recent work studies exact-value oracles for convex optimization on a ball, where a single real-valued reply can encode many coordinates.
For deterministic algorithms, \citet{kerger2026gap} proves a nearly quadratic lower bound for Lipschitz convex minimization, and \citet{wu2026deterministic,wu2026strongly} use shielded chains and delayed rotations to prove lower bounds for smooth and strongly convex optimization that match upper bounds on specified accuracy ranges.
For randomized algorithms, \citet{zhang2026randomized} use a posterior mean energy method to prove a lower bound with an extra factor of $d$, up to logarithmic factors, in Lipschitz convex optimization.
Under noisy evaluations on a ball, \citet{zhang2026joint} establish a joint dimension--accuracy lower bound for Goldstein stationarity.
Those exact-value convex bounds concern the objective error on a ball, whereas ours concern the gradient norm on all of $\R^d$ under a curvature trace budget and hold even when the oracle returns adaptive gradient projections charged by their rank.
On the unrestricted smooth class with exact function values, concurrent and independent work of \citet{xia2026global} proves a randomized lower bound of $\Omega(dT)$ when $T\ge c$ and $d\ge CT\log(2+T)$ for absolute constants $c,C>0$.
We prove a deterministic lower bound with the explicit constant $1/500$ that also covers these gradient projections (Theorem~\ref{thm:unrestricted}); for that oracle, a randomized $\Omega(dT)$ lower bound remains open.

\paragraph{The price of determinism.}
For Lipschitz nonsmooth functions and Goldstein stationarity, \citet{jordan2023deterministic} prove that deterministic methods require at least $\Omega(d)$ calls in a setting where randomized methods attain dimension-free rates, and \citet{kornowski2026exponential} shows that this deterministic overhead can be exponential.
In one dimension, \citet{chewi2023complexity} prove that randomness or function values help when finding stationary points of smooth functions.
We establish the high-dimensional smooth counterpart under a low curvature trace (Theorems~\ref{thm:main-random} and~\ref{thm:main-det}; Corollary~\ref{cor:main-window}).
Specifically, when $ET\le d$, determinism multiplies the query complexity by a factor of order $d/(E+\sqrt{ET})$ up to $\sqrt{\log(e+T)}$, and across the window $T\log(e+T)\le E\le d/T$, the deterministic-to-randomized cost ratio is of order $d/E$ up to constant factors.

\section{Empirical Illustration}
\label{sec:empirical}

We test the algorithms from our upper bounds on fixed instances to illustrate the mechanisms behind Theorems~\ref{thm:main-random} and~\ref{thm:main-det}.
The runs highlight four behaviors: (i) deterministic query counts scale linearly with $d$, whereas randomized query counts stay nearly constant in $d$ and grow with $E$ and $T$; (ii) the Gaussian two-point method and its monotone variant, which accepts only steps that lower $f$ (Theorems~\ref{g2:thm:global} and~\ref{b:thm:monotone}), achieve similar first-hit counts, roughly half those of the three-point method, even when curvature is controlled only below the starting value $f(0)$; (iii) fixed-seed runs can be defeated; and (iv) subspace search saves calls by reusing directions.
We run every method without access to the true gradient, count every function evaluation (including rejected proposals), and compute first-hit call counts offline from the logged iterates; full protocols and parameter grids appear in Appendix~\ref{app:empirical}.

\paragraph{Randomized versus deterministic.}
We compare the Gaussian two-point method \citep{nesterov2017random}, its monotone variant, and the three-point method (Theorem~\ref{x:thm:stp}) against coordinate central differences, finite-difference L-BFGS \citep{nocedal1980updating,liu1989limited}, and restarted subspace search on rank-one quadratics and rotated separable nonconvex functions.
Because deterministic methods spend about $d$ calls per gradient estimate, their cost grows linearly with $d$, whereas that of the randomized methods stays nearly constant in $d$ (Figure~\ref{fig:empirical-mechanism}(a)).
At $d=10^5$ in panel~(a), subspace search reaches the target after $200{,}009$ calls and coordinate finite differences after $600{,}001$ calls, whereas the three-point method uses $77$ calls, the two-point method uses $38$, and the monotone two-point method uses $39$.
Across the $215$ instances with $E\in\{1,4,16\}$ (five seeds per cell; Appendix~\ref{app:empirical}), the monotone two-point method has a median first-hit count of $31$ calls against $30$ for the plain two-point method.
On rank-one quadratics whose direction is orthogonal to every query of a fixed-seed run, that run never reaches the target, whereas fresh seeds succeed in $17$--$20$ of $20$ runs.

\paragraph{Curvature controlled only below $f(0)$.}
On wall functions whose Hessian is small where $f(x)<f(0)$ but of order $L$ along almost every direction at points where $f(x)>f(0)$ (Appendix~\ref{app:empirical}), the monotone two-point method again matches the plain two-point method, with median first-hit counts of $69$ and $72$ calls against $173$ for the three-point method.

\paragraph{Subspace search.}
On a tridiagonal quadratic chain, gradients reveal new directions one at a time, so subspace search uses one expansion at $T=2$ and ten at $T=512$ (Figure~\ref{fig:empirical-mechanism}(b)).
By reusing discovered directions instead of re-estimating full gradients, at $d=10^4$ and $T=512$ it uses about a quarter of the calls of coordinate finite differences.

\paragraph{Local curvature of a smooth model.}
Because GPT-2 uses a smooth activation, its Hessian is defined everywhere.
During MeZO fine-tuning, a simultaneous upper confidence bound for $\operatorname{tr}(G_+)/(d\norm{G}_{\rm op})$, where $G$ is the fixed-batch Hessian, stays between $6.7\times10^{-5}$ and $1.2\times10^{-4}$ (Figure~\ref{fig:empirical-mechanism}(c)).
For $1.24\times10^8$ parameters, this places the upper confidence bound on the local trace budget $\operatorname{tr}(G_+)/\norm{G}_{\rm op}$ below $1.5\times10^4$, comparable to $\sqrt{d}$ and consistent with the few large Hessian eigenvalues reported for neural networks \citep{ghorbani2019investigation}.

\FloatBarrier
\section{Discussion and Open Problems}
\label{sec:discussion}

\paragraph{Conclusion.}
We quantified the price of determinism for smooth nonconvex zeroth-order
optimization under a low curvature trace. Even when the curvature is controlled
only below $f(0)$, randomized methods pay only for the trace, and the Gaussian two-point method with an acceptance test
matches our $\Omega(ET)$ lower bound for $ET\le d$. Deterministic methods also
exploit a low trace, but they pay about $d$ calls for each discovered
direction, so for $ET\le d$ their cost exceeds the randomized cost by a factor of order $d/(E+\sqrt{ET})$, up to
$\sqrt{\log(e+T)}$ (Theorems~\ref{thm:main-random} and~\ref{thm:main-det}).

\paragraph{Limitations and open problems.}
We prove the separation for exact function values, whereas \citet{anonymous2026tracenorm} also allow noisy evaluations; because independent noise can supply random bits, our deterministic lower bounds do not transfer to that model.
For the Gaussian two-point method on $\cB^d_{L,\Delta,E}$ without an acceptance test, our bound carries an extra $T\log(ET)$ term and requires a very small finite-difference radius (Theorem~\ref{g2:thm:sublevel}), so it is open whether the plain method with a step size of order $1/(LE)$ achieves $O(ET)$ calls when $E<\log(1000ET)$.
The methods behind our randomized upper bounds also take $E$ as an input, leaving open whether a randomized algorithm can attain an $O(ET)$ query cap without knowing $E$ in advance.
Other open problems are the remaining factor $\sqrt{\log(e+T)}$, the deterministic complexity on the local pointwise class of Appendix~\ref{app:classes}, and the complexities for $ET>d$.

\bibliographystyle{plainnat}
\bibliography{references}

\section*{AI Usage Statement}
We used AI assistants, including OpenAI's GPT models and Anthropic's Claude,
to help develop and check proofs, formalize them in Lean, implement the
experiments, and edit the text. The authors take full responsibility for the content of
this paper.

\clearpage
\appendix
\section{Main Theorem and Proof Overview}\label{a:sec:main}
Write $T=L\Delta/\varepsilon^2\ge1$ and $1\le E\le d$. The class
$\mathcal B$ contains globally $L$-smooth $C^{1,1}$ functions with
$f(0)-\inf f\le\Delta$ and one fixed unknown matrix $0\preceq H\preceq LI$,
$\operatorname{tr}H\le LE$, dominating the Hessian almost everywhere on
$\{f<f(0)\}$. The subclass $\mathcal F$ requires the matrix bound globally.
An exact value costs one; feedback $(f(x),V^\top\nabla f(x))$ costs
$1+\operatorname{rank}V$. The output need not be queried; exact computation
is uncharged, and the superscripts $\mathrm{cap}$ and $\mathrm{exp}$ mark the cost cap of Section~\ref{sec:model} and the worst-case expected cost.
\begin{theorem}[Main bounds with explicit constants]\label{a:thm:main}
For either oracle on $\mathcal B$,
\begin{align}
 N_{\rm rand}^{\rm cap}&\le23ET,&
 N_{\rm rand}^{\rm cap}&\ge\min\{d,ET\}/12,&
 N_{\rm rand}^{\rm exp}&\ge\min\{d,ET\}/144,\nonumber\\
 N_{\rm det}&\le\min\{10dT,100\sqrt{dET(d+T)}\},&
 N_{\rm det}&\ge d.\label{a:eq:main-bounds}
\end{align}
If $ET\le d$, then
\[
 N_{\rm det}\ge\frac d{672}
 \min\bigl\{T,\sqrt{ET/\log(e+T)}\bigr\}.
\]
Thus for $T\log(e+T)\le E\le d/T$ the deterministic bounds are
$\Omega(dT)$ and $O(dT)$, and the randomized bounds are $\Omega(ET)$
and $O(ET)$. On the unrestricted globally smooth class,
$N_{\rm det}\ge\frac d{500}\min\{T,d/\log(ed)\}$.
Every bound above for $\mathcal B$ also holds on $\mathcal F$: the upper
bounds because $\mathcal F\subseteq\mathcal B$, the randomized lower bounds
and $N_{\rm det}\ge d$ because their quadratic witnesses lie in $\mathcal F$,
and the bound for $ET\le d$ by Corollary~\ref{gl:cor:global-rate}.
\end{theorem}
\ifphaseinappendix\begin{figure}[t]
\centering
\includegraphics[width=\phasefigwidth]{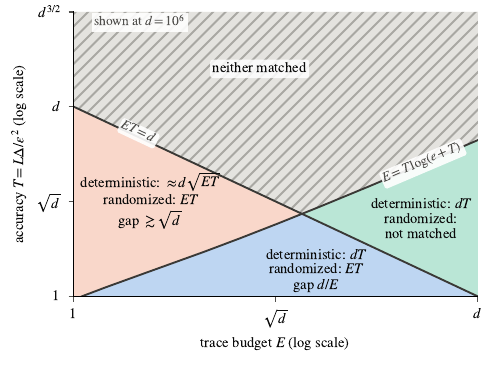}
\vskip -0.08in
\caption{Which of our bounds match? The curves $E=T\log(e+T)$ and $ET=d$ divide
the parameter plane, drawn for $d=10^6$. In each region, the figure gives the
deterministic and randomized costs where our lower and upper bounds match up to
constant factors, or up to $\sqrt{\log(e+T)}$ where marked with $\approx$;
``not matched'' marks regions where our lower and upper bounds differ. Theorem~\ref{a:thm:main} and Corollary~\ref{gl:cor:global-window} give
the conditions.}
\label{fig:phase}
\vskip -0.1in
\end{figure}
\fi
\paragraph{Mechanism and roadmap.}
For the upper bounds (Appendix~\ref{a:sec:upper}), stopping a segment at its
first exit from the initial sublevel set gives a capped quadratic upper model,
so comparison and acceptance steps pay only for the trace, and the
deterministic search buys a full gradient estimate only when its current
subspace is exhausted, with its orthogonal expansions limited by the trace
budget. The deterministic lower bounds combine one scalar chain with a radial
compression and two choices of its parameters; a transcript lemma handles
pending and closing calls and unqueried outputs, finite partial coloring
selects the hidden directions, and scaling places the witnesses in the class.
A quadratic family with an unknown center gives the randomized lower bound.
\paragraph{Scope.}
The multistage witnesses of this proof chain are $C^{1,1}$, and the Lean
formalization of the lower bounds (Appendix~\ref{app:lean}) follows this chain.
Appendix~\ref{app:further} gives the three-point bound on the larger class
$\cA^d_{L,\Delta,E}$ (the local pointwise class of Appendix~\ref{app:classes}) and, reusing the steps of this chain, $C^\infty$ witnesses
with a Lipschitz Hessian for the same deterministic lower bound on $\mathcal F$,
where a perspective weight replaces the radial potential. Appendix~\ref{rb:sec:main} shows that, for $500<T\le d/\log(ed)$, the hard functions of the unrestricted bound are minimized exactly, with probability at least $5/6$, by a randomized method using $O(d\sqrt{T\log(e+T)})$ values, and combines our bounds with the lower bound of \citet{xia2026global} into a wider window with matching bounds for exact values. In the unrestricted bound the witnesses are
coercive and nonconvex when $\min\{T,d/\log(ed)\}>500$; otherwise the
rank-one quadratic of Lemma~\ref{lem:dimension-floor} is used.

\subsection{From Theorem~\ref{a:thm:main} to Section~\ref{sec:results}}
\label{app:derivations}
Theorem~\ref{thm:main-random} consists of the bounds $23ET$ and
$\min\{d,ET\}/12$; the upper bound also holds on $\cA^d_{L,\Delta,E}$, with
$43ET$ calls (Theorem~\ref{x:thm:stp}). The Gaussian two-point method attains
it with $12ET+26T+3$ calls when it accepts only steps that lower $f$
(Theorem~\ref{b:thm:monotone}); without this test it uses $34(E+2)T+2$
calls on $\mathcal F$ and $O(ET+T\log(ET))$ calls on $\mathcal B$
(Theorems~\ref{g2:thm:global} and~\ref{g2:thm:sublevel}). In Theorem~\ref{thm:main-det},
$ET\le d$ and $E\ge1$ give $T\le d$, so
$100\sqrt{dET(d+T)}\le100\sqrt2\,d\sqrt{ET}$; the lower bound is
Corollary~\ref{a:cor:smooth-rate}, and on $\mathcal F$ it is
Corollary~\ref{gl:cor:global-rate}. If $T\log(e+T)\le E\le d/T$, then
$ET\le d$ and $\sqrt{ET/\log(e+T)}\ge T$, which gives
Corollary~\ref{cor:main-window}, with the explicit ratio bounds of
Corollary~\ref{b:cor:aligned}. Theorem~\ref{thm:unrestricted}
combines Theorem~\ref{thm:linear-lower} with the coordinate method in the
proof of Theorem~\ref{u:thm:main}, which uses only global
$L$-smoothness. The remaining statements of Section~\ref{sec:results} are
Lemma~\ref{lem:dimension-floor}, Theorem~\ref{u:thm:main},
Corollaries~\ref{a:cor:smooth-rate}, \ref{gl:cor:global-rate}
and~\ref{gl:cor:global-window}, and, for exact values beyond $ET\le d$,
Corollary~\ref{rb:cor:xy-window}.

\section{Upper Bounds}
\label{a:sec:upper}
\begin{theorem}[Restarted deterministic subspace search]
\label{u:thm:main}
For all the preceding parameters, including $T>d$, one has
\begin{equation}
 N_{\rm det}^{\rm lin}(\mathcal B)\le N_{\rm det}^{\rm val}(\mathcal B)
 \le \min\{10dT,\;100\sqrt{dET(d+T)}\}.
 \label{u:eq:main}
\end{equation}
The bounds hold for every function in the class with a single deterministic,
finite algorithm that uses exact values, chosen from the public parameters. No
convexity, Hessian rank bound, or Hessian Lipschitz assumption is added.
The same result holds on the global fixed-matrix subclass $\mathcal F$.
A single algorithm not given $E$ or $H$ has the hard bound
\begin{equation}
 Q\le\min\{20dT,\;560\sqrt{dET(d+T)}\}.
 \label{u:eq:unknown}
\end{equation}
It is still given $d,L,\Delta,\varepsilon$.
\end{theorem}

\begin{lemma}[Hessians almost everywhere and on every segment]
\label{a:lem:hessian-segment}
Let $f\in C^{1,1}(\mathbb R^d)$, $U\subseteq\mathbb R^d$ be open,
and $H$ be symmetric. Then $\nabla^2f\preceq H$ almost everywhere on $U$
is equivalent to
\begin{equation}
 f(x+v)\le f(x)+\langle\nabla f(x),v\rangle+\tfrac12 v^\top Hv
 \quad\text{for each }[x,x+v]\subset U.
 \label{a:eq:segment}
\end{equation}
\end{lemma}
\begin{proof}
Choose a measurable null set $N\subset U$ outside which the Lipschitz map
$\nabla f$ is differentiable, and its derivative is symmetric and bounded
above by $H$. Fix $v\ne0$ and $[x,x+v]\subset U$. By compactness a closed
tube of sufficiently small positive radius about this segment lies in $U$.
Use orthonormal coordinates with first direction $v/\|v\|$. Fubini applied
to $\mathbf1_N$ on the tube implies that for almost every sufficiently
small transverse displacement $z\in v^\perp$ the set
$\{t\in[0,1]:x+z+tv\in N\}$ has one-dimensional measure zero.
For each such $z$, $G_z(t)=\nabla f(x+z+tv)$ is Lipschitz and hence
absolutely continuous. For almost every $t$,
$G'_z(t)=\nabla^2f(x+z+tv)v$, whence
\[
 \langle G_z(t)-G_z(0),v\rangle
 =\int_0^t v^\top\nabla^2 f(x+z+sv)v\,ds
 \le t v^\top Hv.
\]
Integration over $t\in[0,1]$ proves~\eqref{a:eq:segment} on the translated
segment. The permitted transverse displacements have full measure in a
ball, so choose $z_n\to0$ among them. Continuity of $f$ and $\nabla f$
proves the inequality for the given segment. When $d=1$, the inverse image
of a one-dimensional null set under an injective affine map is null; the
same argument works without translation. The case $v=0$ is immediate.
Conversely, at each point where $\nabla f$ is differentiable with derivative
$B$, continuity and integration of the gradient give
\[
 f(x+tv)=f(x)+t\langle\nabla f(x),v\rangle+
             \tfrac{t^2}{2}v^\top Bv+o(t^2).
\]
Apply~\eqref{a:eq:segment} for small positive $t$ and divide by $t^2/2$;
then $v^\top Bv\le v^\top Hv$. Because the differentiability point was fixed,
this holds for every $v$ at once. Rademacher's theorem for $\nabla f$
provides such points almost everywhere.
\end{proof}
\subsection{The capped upper model and finite differences}
\label{u:analytic}
\begin{lemma}[First exit and the capped upper model]
\label{u:lem:cap}
Suppose $f\in C^1$, $H\succeq0$, and every segment $[x,y]$ contained in
$\{f<c\}$ satisfies
\[
 f(y)\le f(x)+\langle\nabla f(x),y-x\rangle+\tfrac12(y-x)^\top H(y-x).
\]
For every $f(x)\le c$ and every $v$,
\begin{equation}
 \min\{f(x+v),c\}\le
 f(x)+\langle\nabla f(x),v\rangle+\tfrac12v^\top Hv.
 \label{u:eq:cap}
\end{equation}
The hypothesis holds for $\mathcal B$ with $c=f(0)$.
\end{lemma}
\begin{proof}
For $f(x)<c$, let $q(t)=f(x)+t\langle\nabla f(x),v\rangle+
 t^2v^\top Hv/2$. If the segment stays strictly below $c$, use the hypothesis.
Otherwise let $s\in(0,1]$ be its first exit time. The hypothesis at times
$t<s$ and continuity imply $q(s)\ge c$. Convexity gives
$q(s)\le(1-s)q(0)+s q(1)$, hence $q(1)\ge c$. This proves the claim.
At $f(x)=c$, a nonnegative directional derivative makes the assertion
immediate. With a negative directional derivative, $x+sv$ lies strictly
below $c$ for all sufficiently small positive $s$. Apply the proved claim
there, with displacement $(1-s)v$, and take $s\downarrow0$.

The full equivalence is in Lemma~\ref{a:lem:hessian-segment}.
In particular, the hypothesis for $\mathcal B$ is supplied by that lemma.
\end{proof}

For an orthonormal $d\times k$ matrix $V$, $k\ge1$, define
\begin{equation}
 b_k=\frac{\varepsilon}{8L\sqrt k},\qquad
 [D_Vf(x)]_i=\frac{f(x+b_kv_i)-f(x-b_kv_i)}{2b_k}.
 \label{u:eq:diff}
\end{equation}
It costs $2k$ values and has vector error at most $\varepsilon/16$.
Indeed, the scalar error is bounded by
$(2b_k)^{-1}\int_{-b_k}^{b_k}L|t|\,dt=Lb_k/2$;
summing its squares proves the assertion. For $k=0$ use the empty vector
at zero cost. Global $L$-smoothness, not the sublevel matrix condition,
controls these arbitrary finite probes.

\begin{lemma}[Restricted descent and a certified residual]
\label{u:lem:residual}
Repeatedly compute $z=D_Vf(x)$, stop when $\|z\|\le7\varepsilon/16$, and
otherwise set $x\leftarrow x-Vz/L$. Each nonterminal step decreases $f$ by
more than $35\varepsilon^2/(512L)$. On return,
$\|V^\top\nabla f(x)\|\le\varepsilon/2$.
Next compute $g=D_{I_d}f(x)$. If $\|g\|\le15\varepsilon/16$, then
$\|\nabla f(x)\|\le\varepsilon$. Otherwise
\[
 u=\frac{g-VV^\top g}{\|g-VV^\top g\|}
\]
is well defined, orthogonal to $V$, and
$\langle\nabla f(x),u\rangle>11\varepsilon/16$.
\end{lemma}
\begin{proof}
Writing $w=V^\top\nabla f(x)$ and $e=z-w$ gives
\[
 f(x-Vz/L)\le f(x)-\|w\|^2/(2L)+\|e\|^2/(2L).
\]
At a nonterminal step $\|w\|>3\varepsilon/8$ and
$\|e\|\le\varepsilon/16$, giving $35/512$. The terminal assertion follows
by the triangle inequality. At a failed full test,
$\|V^\top g\|\le9\varepsilon/16$, whereas $\|g\|>15\varepsilon/16$.
Pythagoras gives $\|g-VV^\top g\|>3\varepsilon/4$. Subtract the full
estimation error $\varepsilon/16$ to obtain the asserted slope.
The successful full test also follows from that error bound.
\end{proof}

\subsection{A doubling line search}
\label{u:line}
The search uses an Armijo-type sufficient decrease test \citep{armijo1966minimization}
with doubling trial steps.
\begin{lemma}[Decrease along a new direction and amortized query cost]
\label{u:lem:line}
Suppose $f(x)\le f(0)$ and $u$ is a unit vector satisfying
$\langle\nabla f(x),u\rangle>11\varepsilon/16$. Query $f(x)$, and test
$t_j=2^j\varepsilon/(2L)$ for $j=0,1,\ldots$. A test passes precisely when
\begin{equation}
 f(x-t_ju)\le f(x)-\varepsilon t_j/4.
 \label{u:eq:armijo}
\end{equation}
Stop at the first failure and return the preceding accepted point $x^+$.
For $\lambda=u^\top Hu$ one has $\lambda>0$ and
\begin{equation}
 D_x:=f(x)-f(x^+)>
 \frac{7\varepsilon^2}{64\lambda}>
 \frac{\varepsilon^2}{10\lambda}.
 \label{u:eq:inverse}
\end{equation}
The search terminates. Its number of proposal evaluations, including the
failed proposal but excluding its one center evaluation, is at most
$8LD_x/\varepsilon^2+1$.
\end{lemma}
\begin{proof}
Apply \eqref{u:eq:cap} to $-tu$. If $\lambda=0$, the capped upper bound
tends to $-\infty$, contradicting boundedness below. If
$0<t\le7\varepsilon/(8\lambda)$, the bound is strictly below
$f(x)-\varepsilon t/4<f(0)$, so the actual value equals its cap and passes
the test. Since $\lambda\le L$, the initial test passes.
An accepted test must have $t\le4(f(x)-\inf f)/\varepsilon$, so a failure
occurs at a finite index. If $s$ was the preceding accepted step, the failed
step $2s$ exceeds $7\varepsilon/(8\lambda)$. Therefore
$D_x\ge\varepsilon s/4>7\varepsilon^2/(64\lambda)$.
No monotonicity of test outcomes after the guaranteed initial interval is
assumed.

Write $s=2^j\varepsilon/(2L)$. The search used $j+2$ proposals, and
$D_x\ge2^j\varepsilon^2/(8L)$. The elementary induction
$j+2\le2^j+1$ gives the last assertion.
\end{proof}

Consequently, for $n$ expansion moves along \emph{one} monotone trajectory,
all their proposal evaluations together number at most $8T+n$.
The $n$ center evaluations are charged separately. There is no
$n\log T$ term.

\subsection{The restarted algorithm and its cost}
\label{u:algorithm}
Algorithm~\ref{alg:subspace} collects the steps of
Lemmas~\ref{u:lem:residual} and~\ref{u:lem:line} for a fixed
capacity $r\in\{1,\ldots,d\}$. Within the current span $V$ it descends by
central differences; when no progress is certified there, it estimates the
full gradient, returns if the estimate is small, and otherwise adds the
normalized residual direction, moves along it by a doubling line search, and
restarts with an empty span after $r$ additions.

\begin{algorithm}[t]
\caption{Restarted subspace search with capacity $r$. For an orthonormal
$d\times k$ matrix $V$ with columns $v_1,\ldots,v_k$, the vector $D_Vf(x)$ has
entries $[f(x+b_kv_i)-f(x-b_kv_i)]/(2b_k)$ with $b_k=\eps/(8L\sqrt k)$, as
in~\eqref{u:eq:diff}, and is empty when $k=0$; $D_{I_d}$ uses the
coordinate vectors. Steps are $t_j=2^j\eps/(2L)$.}
\label{alg:subspace}
\begin{algorithmic}[1]
\Require $d$, $L$, $\eps$, a capacity $r\in\{1,\ldots,d\}$, and exact values of $f$
\State $x\gets0$, and let $V$ be the empty $d\times0$ matrix
\Loop
  \State $z\gets D_Vf(x)$
  \While{$\norm z>7\eps/16$} \hfill{\itshape descent within the span}
    \State $x\gets x-Vz/L$, \ $z\gets D_Vf(x)$
  \EndWhile
  \State $g\gets D_{I_d}f(x)$ \hfill{\itshape full estimate, $2d$ values}
  \If{$\norm g\le15\eps/16$}
    \State \Return $x$
  \EndIf
  \State $u\gets(g-VV^\top g)/\norm{g-VV^\top g}$ \hfill{\itshape new direction}
  \State $j\gets0$
  \While{$f(x-t_ju)\le f(x)-\eps t_j/4$} \hfill{\itshape doubling line search}
    \State $j\gets j+1$
  \EndWhile
  \State $x\gets x-t_{j-1}u$, \ $V\gets[V\ u]$
  \If{$V$ has $r$ columns}
    \State let $V$ be empty \hfill{\itshape restart}
  \EndIf
\EndLoop
\end{algorithmic}
\end{algorithm}

The test at $t_0$ always passes (Lemma~\ref{u:lem:line}), so $j\ge1$ when
the line search stops. Theorem~\ref{u:thm:main} uses
$r=\lceil\sqrt{dET/(d+T)}\rceil$; without $E$ and $H$,
Section~\ref{u:unknown} runs copies with $r=1,2,4,\ldots$ under fixed
budgets.
Each branch uses exact function values and deterministic operations. A
single fixed-$r$ run does not need $E,H$, or $\Delta$. The basis dimension
at the beginning of any restricted search is at most $r-1$.

\begin{lemma}[A square budget across restarts]
\label{u:lem:blocks}
Suppose there are $n=qr+s$ expansions in total, where $q\ge0$ and
$0\le s<r$. Then
\begin{equation}
 qr^2+s^2<10ET,\qquad
 n<\frac{10ET}{r}+\frac r4.
 \label{u:eq:blocks}
\end{equation}
All restricted nonterminal updates together number at most
$J=\lfloor512T/35\rfloor$. The algorithm stops with a certified output.
\end{lemma}
\begin{proof}
In a restart block containing $k$ expansions, the appended vectors are
orthonormal. For their positive curvatures $\lambda_i=u_i^\top Hu_i$,
\[
 \sum_{i=1}^k\lambda_i\le LE,\qquad
 \sum_{i=1}^kD_i>\frac{\varepsilon^2}{10}
                           \sum_{i=1}^k\lambda_i^{-1}
 \ge\frac{\varepsilon^2k^2}{10LE}.
\]
The last inequality is Cauchy--Schwarz. Crucially, this trace sum is used
\emph{within each block only}; vectors from different blocks need not be
orthogonal. Their actual decreases are disjoint parts of one monotone
trajectory, so summing the block inequalities gives $qr^2+s^2<10ET$.
Now
\[
 n=\frac{qr^2+s^2}{r}+s(1-s/r)
 <\frac{10ET}{r}+\frac r4,
\]
since $(s-r/2)^2\ge0$.

Lemma~\ref{u:lem:residual} and the same monotone gap budget bound the
restricted updates by $J$. Every line search is finite. An infinite run
would therefore require infinitely many expansions, hence infinitely many
complete blocks, contradicting their positive uniform gap charge.
The only return is the certified return after a full gradient estimate.
\end{proof}

\begin{theorem}[Cost at a fixed capacity]
\label{u:thm:capacity}
For any $r\in\{1,\ldots,d\}$ the preceding deterministic algorithm has the
worst-case query count
\begin{align}
 Q_r&\le2(d+r-1)(n+1)+2(r-1)J+8T+2n,\label{u:eq:ledger}\\
 Q_r&<\frac{40dET}{r}+30(d+T)r.\label{u:eq:capacity}
\end{align}
Its stored basis uses at most $dr$ real entries.
\end{theorem}
\begin{proof}
There are $n+1$ full estimates at cost $2d$ each and at most $n+1$ terminal
restricted estimates at cost $2(r-1)$ each. The $J$ nonterminal restricted
estimates also cost at most $2(r-1)$ each. The preceding line search
amortization charges at most $8T+n$ proposals and $n$ centers. This proves
\eqref{u:eq:ledger}; in particular it charges the final tests and every rejected proposal.

Because $r\le d$, \eqref{u:eq:ledger} and \eqref{u:eq:blocks} yield
\[
 Q_r<\frac{40dET}{r}+dr+4d+\frac{1024}{35}(r-1)T+8T.
\]
Use $dr+4d\le5dr$, and
$(1024/35)(r-1)T+8T\le(1024/35)rT<30rT$.
The resulting expression is smaller than \eqref{u:eq:capacity}.
\end{proof}

\begin{proof}[Proof of \eqref{u:eq:main}]
Set $A=dET/(d+T)$ and $Z=\sqrt{dET(d+T)}$. One has $A\ge1/2$ and
$A<d^2$, so $r=\lceil\sqrt A\rceil$ lies in $\{1,\ldots,d\}$ and
\[
 \sqrt A\le r\le2\sqrt A.
\]
For $\sqrt A\le1$ the latter follows from $A\ge1/2$; for $\sqrt A>1$
it follows from $\lceil\sqrt A\rceil\le\sqrt A+1$.
Equation~\eqref{u:eq:capacity} then gives $Q_r<40Z+60Z=100Z$.

For completeness, the ordinary coordinate algorithm
(Algorithm~\ref{alg:coordinate}) uses difference
radius $\varepsilon/(4L\sqrt d)$, vector error $\varepsilon/8$, stopping
threshold $7\varepsilon/8$, and update $x-g/L$ otherwise. Every
nonterminal step decreases $f$ by more than $35\varepsilon^2/(128L)$.
Including the final estimate costs at most
$2d(1+\lfloor128T/35\rfloor)\le10dT$ for $T\ge1$.
Selecting the algorithm with the smaller public bound proves the minimum.
If an exact comparison of bounds is inconvenient, interleave the two
correct procedures one charged query at a time; this loses at most a factor
two, not correctness.
\end{proof}

\begin{algorithm}[t]
\caption{Coordinate finite differences. The estimate uses central differences
along the coordinate vectors $e_1,\ldots,e_d$ with radius
$b=\varepsilon/(4L\sqrt d)$.}
\label{alg:coordinate}
\begin{algorithmic}[1]
\Require $d$, $L$, $\varepsilon$, and exact values of $f$
\State $x\gets0$
\Loop
  \State $g\gets\bigl([f(x+be_a)-f(x-be_a)]/(2b)\bigr)_{a=1}^d$ \hfill{\itshape $2d$ values}
  \If{$\norm g\le7\varepsilon/8$}
    \State \Return $x$
  \EndIf
  \State $x\gets x-g/L$
\EndLoop
\end{algorithmic}
\end{algorithm}

\subsection{Removing knowledge of the trace}
\label{u:unknown}
Algorithm~\ref{alg:unknown} combines restarted trials of
Algorithm~\ref{alg:subspace} with growing capacity and the coordinate
algorithm; it reads neither $E$ nor $H$.

\begin{algorithm}[t]
\caption{Subspace search without knowledge of $E$ or $H$. Each trial is a fresh
run of Algorithm~\ref{alg:subspace} from the origin, stopped after its budget;
an unfinished trial is discarded.}
\label{alg:unknown}
\begin{algorithmic}[1]
\Require $d$, $L$, $\Delta$, $\varepsilon$, and exact values of $f$; $T=L\Delta/\varepsilon^2$
\Procedure{Trials}{}
  \For{$j=0,1,\ldots,\lceil\log_2d\rceil$}
    \State $r_j\gets\min\{2^j,d\}$, \ $B_j\gets\lfloor70(d+T)r_j\rfloor$
    \State run Algorithm~\ref{alg:subspace} with capacity $r_j$ for at most $B_j$ value calls
    \If{that run returned a point $x$}
      \State \Return $x$
    \EndIf
  \EndFor
\EndProcedure
\State Run \textsc{Trials} and Algorithm~\ref{alg:coordinate} in alternation, one value call
from each at a time with separate states, and return the first output.
\end{algorithmic}
\end{algorithm}

\begin{proof}[Proof of \eqref{u:eq:unknown}]
Use capacities $r_j=\min\{2^j,d\}$ for $j=0,\ldots,\lceil\log_2d\rceil$.
Run a fresh copy of the fixed-$r_j$ procedure from the origin, capped at
\[
 B_j=\lfloor70(d+T)r_j\rfloor
\]
actual value evaluations. A completed certification at the budget is valid;
otherwise discard an unfinished computation, increase the capacity, and
restart. No trial reads $E$ or $H$.

Define $A,Z$ as above only for the analysis, and let $j_*$ be the first
index with $r_{j_*}\ge\sqrt A$. It exists because $\sqrt A<d$.
For that trial \eqref{u:eq:capacity} gives the strict inequality
$Q_{r_{j_*}}<70(d+T)r_{j_*}$, and since $Q_{r_{j_*}}$ is an integer we have
$Q_{r_{j_*}}\le B_{j_*}$. It therefore completes. If $j_*=0$, then
$\sum_{j=0}^{j_*}r_j=1\le\sqrt{2A}<4\sqrt A$; if $j_*\ge1$, then
$2^{j_*-1}=r_{j_*-1}<\sqrt A$, so $\sum_{j=0}^{j_*-1}r_j=2^{j_*}-1<2\sqrt A$
and $r_{j_*}\le2^{j_*}<2\sqrt A$, giving $\sum_{j=0}^{j_*}r_j<4\sqrt A$ even
when the final capacity is shortened to $d$. The total is less than
$280(d+T)\sqrt A=280Z$.

Finally interleave this entire procedure, which needs no knowledge of the trace, and the coordinate
algorithm, keeping their states separate, one actual value call from each
in alternation. Stop when either returns. Both returns are certified, and
the cost is at most twice the smaller individual cost. This gives
\eqref{u:eq:unknown}. The restarting trials use separate gap budgets;
only their already bounded oracle costs are added.
\end{proof}

\subsection{The three-point method on the main class}
\label{b:sec:stp}
Algorithm~\ref{alg:stp} states the three-point method with step $h$ and $N$
iterations; Theorem~\ref{b:thm:stp} below and Theorem~\ref{x:thm:stp} in
Appendix~\ref{app:further} fix these parameters.

\begin{algorithm}[t]
\caption{Three-point method with step $h$ and $N$ iterations. The comparison
step is the update of \citet{bergou2020stochastic}. On $\mathcal B$,
$h=4\varepsilon/(5LE)$ and $N=\lceil10ET\rceil$ (Theorem~\ref{b:thm:stp});
on $\mathcal A$, $h=4\varepsilon/(5L(E+1))$ and
$N=\lceil10L(E+1)\Delta/\varepsilon^2\rceil$ (Theorem~\ref{x:thm:stp}).}
\label{alg:stp}
\begin{algorithmic}[1]
\Require $h$, $N$, and exact values of $f$
\State $x_0\gets0$; query $f(x_0)$
\For{$t=0,\ldots,N-1$}
  \State draw $a_t\sim N(0,I_d)$ independently; query $f(x_t+ha_t)$ and $f(x_t-ha_t)$
  \State $x_{t+1}\gets$ a point of lowest value among $x_t$, $x_t+ha_t$, $x_t-ha_t$, keeping $x_t$ on a tie
\EndFor
\State draw $J$ uniformly from $\{0,\ldots,N-1\}$, independently, and \Return $x_J$
\end{algorithmic}
\end{algorithm}

\begin{theorem}[An elementary three-point bound on $\mathcal B$]
\label{b:thm:stp}
For $f\in\mathcal B$, take $h=4\varepsilon/(5LE)$ and
$N=\lceil10ET\rceil$. Query $f(0)$ once. For $t=0,\ldots,N-1$, draw an
independent $a_t\sim N(0,I_d)$ and keep a point of lowest value among
$x_t,x_t+ha_t,x_t-ha_t$, caching its value and retaining the center on a tie.
Return an independently uniform point from $x_0,\ldots,x_{N-1}$.
Then
\[
 \mathbb P(\|\nabla f(\widehat x)\|>\varepsilon)<25/78<1/3,
 \qquad Q=1+2\lceil10ET\rceil\le23ET.
\]
This is the update of \citet{bergou2020stochastic}; on $\mathcal B$ its
analysis needs only Lemma~\ref{u:lem:cap} and two Gaussian moments.
\end{theorem}
\begin{proof}
At a base point $x$ with $f(x)\le f(0)$, write $g=\nabla f(x)$.
Of the two proposals, the sign $-\operatorname{sign}(a^\top g)$ has capped
quadratic bound
$f(x)-h|a^\top g|+(h^2/2)a^\top Ha$.
The true three-point minimum is no larger than this expression: if the
expression is below $f(x)$, Lemma~\ref{u:lem:cap} forces an actual proposal
below $f(x)$; otherwise the retained center already suffices.
Consequently
\[
 \mathbb E[f(x)-f(x^+)\mid x]
 \ge h\sqrt{2/\pi}\,\|g\|-(h^2/2)\operatorname{tr}H.
\]
It is also nonnegative because each realized update is monotone.
For $\|g\|>\varepsilon$, use $\sqrt{2/\pi}>79/100$ (for example from
$\pi<22/7$) to lower bound the decrease by
$39\varepsilon^2/(125LE)$.
Sum over all steps: the expected number of indices with gradient above
$\varepsilon$ is less than $125ET/39$.
Uniform output and $N\ge10ET$ give $25/78$. Every proposal, accepted or
rejected, costs one query, while center values are cached. Only Gaussian
first and second moments have been used.
\end{proof}

\subsection{A monotone two-point method on the main class}
\label{b:sec:monotone}
The three-point method moves by a fixed length proportional to $\varepsilon$.
The Gaussian two-point method of \citet{nesterov2017random} instead scales each
step by an estimated directional derivative, but a step may leave the initial
sublevel set, where $\mathcal B$ controls only the gradient Lipschitz constant.
Accepting a proposal only when it lowers $f$ keeps every iterate below $f(0)$,
so Lemma~\ref{u:lem:cap} again controls every step.
Algorithm~\ref{alg:monotone} states the method with the parameters of
Theorem~\ref{b:thm:monotone}.

\begin{algorithm}[t]
\caption{Monotone two-point method, with $\eta=1/(L(E+2))$,
$N=\lceil(6E+13)T\rceil$, and
$\mu=\varepsilon/(4L(6E+13)\sqrt{d(d+2)(d+4)})$ as in
Theorem~\ref{b:thm:monotone}.}
\label{alg:monotone}
\begin{algorithmic}[1]
\Require $\eta$, $N$, $\mu$, and exact values of $f$
\State $x_0\gets0$; query $f(x_0)$
\For{$t=0,\ldots,N-1$}
  \State draw $a_t\sim N(0,I_d)$ independently; query $f(x_t+\mu a_t)$
  \State $s_t\gets[f(x_t+\mu a_t)-f(x_t)]/\mu$ \hfill{\itshape $f(x_t)$ is cached}
  \State $y_t\gets x_t-\eta s_ta_t$; query $f(y_t)$
  \If{$f(y_t)<f(x_t)$}
    \State $x_{t+1}\gets y_t$
  \Else
    \State $x_{t+1}\gets x_t$
  \EndIf
\EndFor
\State draw $J$ uniformly from $\{0,\ldots,N-1\}$, independently, and \Return $x_J$
\end{algorithmic}
\end{algorithm}

\begin{lemma}[Gaussian moments]
\label{b:lem:gauss-fourth}
Let $a\sim N(0,I_d)$, $g\in\mathbb R^d$, and let $H$ be symmetric. Then
$\mathbb E[(g^\top a)^2(a^\top Ha)]=\|g\|^2\operatorname{tr}H+2g^\top Hg$, and
$\mathbb E\|a\|^{2k}=d(d+2)\cdots(d+2k-2)$ for $k=1,2,3$.
\end{lemma}
\begin{proof}
For $g\ne0$, take an orthonormal basis $v_1,\ldots,v_d$ with $v_1=g/\|g\|$ and
let $z_i=v_i^\top a$, which are independent standard normal variables. Then
$(g^\top a)^2=\|g\|^2z_1^2$, and $\mathbb E[z_1^4]=3$, $\mathbb E[z_1^2z_i^2]=1$
for $i\ge2$, and $\mathbb E[z_1^2z_iz_j]=0$ for $i\ne j$ give
$\|g\|^2(3v_1^\top Hv_1+\sum_{i\ge2}v_i^\top Hv_i)$. The moments of
$\|a\|^2\sim\chi^2_d$ are standard.
\end{proof}

\begin{theorem}[A monotone two-point bound on $\mathcal B$]
\label{b:thm:monotone}
For $f\in\mathcal B$, set
\[
 \eta=\frac1{L(E+2)},\qquad N=\lceil(6E+13)T\rceil,\qquad
 \mu=\frac{\varepsilon}{4L(6E+13)\sqrt{d(d+2)(d+4)}}.
\]
Query $f(0)$ once. For $t=0,\ldots,N-1$, draw an independent
$a_t\sim N(0,I_d)$, query $f(x_t+\mu a_t)$, and with the cached value $f(x_t)$
set $s_t=[f(x_t+\mu a_t)-f(x_t)]/\mu$ and $y_t=x_t-\eta s_ta_t$. Query $f(y_t)$,
and set $x_{t+1}=y_t$ if $f(y_t)<f(x_t)$ and $x_{t+1}=x_t$ otherwise. Return an
independently uniform point from $x_0,\ldots,x_{N-1}$. Then
\[
 \mathbb P(\|\nabla f(\widehat x)\|>\varepsilon)<1/3,
 \qquad Q=1+2\lceil(6E+13)T\rceil\le12ET+26T+3.
\]
\end{theorem}
\begin{proof}
The acceptance rule keeps $f(x_t)\le f(0)$. At a base point $x$ with
$f(x)\le f(0)$, write $g=\nabla f(x)$, $a=a_t$, $s=s_t$, $M=L(E+2)=1/\eta$, and
$x^+$ for the next iterate. Since
$f(x^+)=\min\{f(x),f(y_t)\}\le\min\{f(y_t),f(0)\}$,
Lemma~\ref{u:lem:cap} with displacement $-\eta sa$ gives
\[
 f(x)-f(x^+)\ge\eta s(g^\top a)-\tfrac{\eta^2}2s^2(a^\top Ha).
\]
Global $L$-smoothness bounds the difference error $\delta=s-g^\top a$ by
$|\delta|\le L\mu\|a\|^2/2$. With $\gamma=1/(4(6E+13))$, Young's inequality
gives $s(g^\top a)\ge(1-\gamma/2)(g^\top a)^2-\delta^2/(2\gamma)$ and
$s^2\le(1+\gamma)(g^\top a)^2+(1+1/\gamma)\delta^2$. Taking expectations and
using Lemma~\ref{b:lem:gauss-fourth} with $\operatorname{tr}H\le LE$ and
$g^\top Hg\le L\|g\|^2$,
\[
 \mathbb E[f(x)-f(x^+)\mid x]\ge\frac{1-2\gamma}{2M}\|g\|^2
 -\frac1{2\gamma M}\Bigl(\mathbb E[\delta^2]
 +\frac{1+\gamma}M\,\mathbb E[\delta^2(a^\top Ha)]\Bigr).
\]
By Lemma~\ref{b:lem:gauss-fourth} and $a^\top Ha\le L\|a\|^2$,
$\mathbb E[\delta^2]\le L^2\mu^2d(d+2)/4<\varepsilon^2/(64(6E+13)^2)$ and
$\mathbb E[\delta^2(a^\top Ha)]\le L^3\mu^2d(d+2)(d+4)/4
=L\varepsilon^2/(64(6E+13)^2)$. Since $L/M\le1/3$ and $\gamma\le1/76$, the term
in parentheses is below $2\varepsilon^2/(64(6E+13)^2)$, so the subtracted term is
below $\varepsilon^2/(16M(6E+13))$. For $\|g\|>\varepsilon$, as
$2\gamma=1/(2(6E+13))$, this gives
\[
 \mathbb E[f(x)-f(x^+)\mid x]
 >\Bigl(1-\frac1{6E+13}\Bigr)\frac{\varepsilon^2}{2M}
 =\frac{3\varepsilon^2}{L(6E+13)},
\]
and every realized update is nonnegative. Summing over all steps and using
$f(0)-\inf f\le\Delta$, the expected number of indices with gradient norm above
$\varepsilon$ is less than $(6E+13)T/3\le N/3$, which gives the failure bound
under uniform output. Each step queries $f(x_t+\mu a_t)$ and $f(y_t)$ and
reuses the cached value $f(x_t)$.
\end{proof}

\subsection{Gaussian two-point methods without an acceptance test}
\label{g2:sec:gaussian}
We now analyze the Gaussian two-point method of \citet{nesterov2017random},
the estimator behind zeroth-order fine-tuning \citep{malladi2023mezo}, with
$q$ directions per step and no acceptance test. For $q\ge1$, let $a_{t,1},\ldots,a_{t,q}$ be independent standard
Gaussian vectors in $\R^d$, fresh at each iteration. We consider either the
forward differences
\begin{equation}
 s_{t,i}^{\rm f}=
 \frac{f(x_t+\mu a_{t,i})-f(x_t)}{\mu},
 \label{g2:eq:forward}
\end{equation}
or the central differences
\begin{equation}
 s_{t,i}^{\rm c}=
 \frac{f(x_t+\mu a_{t,i})-f(x_t-\mu a_{t,i})}{2\mu}.
 \label{g2:eq:central}
\end{equation}
In either case the update is
\begin{equation}
 x_{t+1}=x_t-\frac\eta q\sum_{i=1}^q s_{t,i}a_{t,i},
 \qquad x_0=0.
 \label{g2:eq:update}
\end{equation}
There is no test or rejection of $x_{t+1}$. The output is $x_J$ for an
independent $J$ uniform on $\{0,\ldots,N-1\}$ (Algorithm~\ref{alg:twopoint}). For forward differences,
$f(x_t)$ is shared among the directions, so each iteration costs $q+1$
values; for central differences it costs $2q$ values. Both costs are at
most $2q$. In particular, no evaluation of $f(x_{t+1})$ is required in the
current iteration.

\begin{algorithm}[t]
\caption{Gaussian two-point method with $q$ directions per step and no
acceptance test. Theorem~\ref{g2:thm:global} gives $\eta,\mu,N$ on
$\mathcal F$, and Theorem~\ref{g2:thm:sublevel} gives them on
$\mathcal B$.}
\label{alg:twopoint}
\begin{algorithmic}[1]
\Require $q$, $\eta$, $\mu$, $N$, the choice of forward or central differences, and exact values of $f$
\State $x_0\gets0$
\For{$t=0,\ldots,N-1$}
  \State draw $a_{t,1},\ldots,a_{t,q}\sim N(0,I_d)$ independently
  \For{$i=1,\ldots,q$}
    \State $s_{t,i}\gets[f(x_t+\mu a_{t,i})-f(x_t)]/\mu$ (forward) or $[f(x_t+\mu a_{t,i})-f(x_t-\mu a_{t,i})]/(2\mu)$ (central)
  \EndFor
  \State $x_{t+1}\gets x_t-\frac\eta q\sum_{i=1}^qs_{t,i}a_{t,i}$
\EndFor
\State draw $J$ uniformly from $\{0,\ldots,N-1\}$, independently, and \Return $x_J$
\end{algorithmic}
\end{algorithm}

\begin{lemma}[Two-point errors and batch Gaussian curvature]
\label{g2:lem:batch}
Suppose $f$ is globally $L$-smooth. For either difference in
\eqref{g2:eq:forward}--\eqref{g2:eq:central}, writing
$g=\nabla f(x)$, $\sigma_i=a_i^\top g$, and $s_i=\sigma_i+\beta_i$,
\begin{equation}
 |\beta_i|\le\frac{L\mu}{2}\norm{a_i}^2.
 \label{g2:eq:error}
\end{equation}
Define $v=q^{-1}\sum_i\sigma_i a_i$ and
$e=q^{-1}\sum_i\beta_i a_i$, and write
$D_6=d(d+2)(d+4)$. Then
\begin{equation}
 \E v=g,\qquad \E\norm e^2\le\frac{L^2\mu^2D_6}{4}.
 \label{g2:eq:bias}
\end{equation}
If also $H\succeq0$, $\norm H_{\rm op}\le L$, and $\tr H\le LE$,
then
\begin{equation}
 \E[v^\top Hv]\le\frac{L(E+q+1)}q\norm g^2.
 \label{g2:eq:batch-moment}
\end{equation}
\end{lemma}
\begin{proof}
For forward differences integrate
$(\nabla f(x+ta_i)-g)^\top a_i$ over $t\in[0,\mu]$; for central
differences integrate over $[-\mu,\mu]$ and divide by $2\mu$.
Global smoothness gives \eqref{g2:eq:error} in either case.
Jensen's inequality applied to the average defining $e$, followed by
$\E\norm a^6=D_6$, proves \eqref{g2:eq:bias}. The identity
$\E[(a^\top g)a]=g$ proves the other part of \eqref{g2:eq:bias}.
For the last claim expand the quadratic form of $v$. The $q$
diagonal contributions are bounded, using
Lemma~\ref{b:lem:gauss-fourth}, by
$\norm g^2\tr H+2g^\top Hg\le L(E+2)\norm g^2$.
By independence, each of the $q(q-1)$ off-diagonal contributions is
$g^\top Hg\le L\norm g^2$. Division by $q^2$ gives
\eqref{g2:eq:batch-moment}.
\end{proof}

\begin{theorem}[Gaussian two-point method on $\mathcal F$]
\label{g2:thm:global}
Let $f\in\cF^d_{L,\Delta,E}$ and $1\le E\le d$, $T=L\Delta/\eps^2\ge1$.
For any integer $q\ge1$, run \eqref{g2:eq:update} with either
\eqref{g2:eq:forward} or \eqref{g2:eq:central}, using
\begin{equation}
 \eta=\frac{q}{L(E+q+1)},\quad
 \mu=\frac{\eps}{8L\sqrt{d(d+2)(d+4)}},\quad
 N=\left\lceil\frac{17(E+q+1)T}{q}\right\rceil.
 \label{g2:eq:global-parameters}
\end{equation}
Then on each function
\begin{equation}
 \Prob(\norm{\nabla f(x_J)}>\eps)\le\frac{11}{34}<\frac13,
 \qquad Q\le2qN\le34(E+q+1)T+2q.
 \label{g2:eq:global-guarantee}
\end{equation}
In particular, with one direction per step ($q=1$) the method uses at most
$34(E+2)T+2$ value calls.
\end{theorem}
\begin{proof}
Let $C=L(E+q+1)/q=1/\eta$. The global quadratic upper model
\eqref{eq:trace-class} applied with the random displacement
$-\eta(v+e)$ implies, conditionally on $x$,
\[
 \E[f(x)-f(x-\eta(v+e))]\ge
 \eta\norm g^2-\eta\norm g\,\E\norm e
 -\frac{\eta^2}{2}\E[(v+e)^\top H(v+e)].
\]
Since $H\succeq0$, the matrix Young inequality with parameter $1/8$ gives
\[
 (v+e)^\top H(v+e)
 \le\frac98 v^\top Hv+9e^\top He.
\]
Put $\kappa=L\mu\sqrt{D_6}/2=\eps/16$. By
Lemma~\ref{g2:lem:batch}, Jensen and $\norm H_{\rm op}\le L$,
\[
 \E\norm e\le\kappa,\quad
 \E[e^\top He]\le L\kappa^2,\quad
 \E[v^\top Hv]\le C\norm g^2.
\]
Since $\kappa\norm g\le\norm g^2/16+4\kappa^2$ and
$\eta L=q/(E+q+1)\le1$,
\begin{align*}
 \E[f(x)-f(x_{t+1})\mid x_t=x]
 &\ge\eta\bigl(\tfrac7{16}\norm g^2
       -\kappa\norm g-\tfrac92\eta L\kappa^2\bigr)\\
 &\ge\eta\bigl(\tfrac38\norm g^2-\tfrac{17}2\kappa^2\bigr)\\
 &\ge\eta\bigl(\tfrac38\norm g^2-\tfrac1{16}\eps^2\bigr),
\end{align*}
where $17/512\le1/16$ is the last numerical inequality.
Summing the conditional expectations and using
$\E f(x_N)\ge\inf f\ge f(0)-\Delta$ yields
\[
 \frac1N\sum_{t=0}^{N-1}\E\norm{\nabla f(x_t)}^2
 \le\frac{8\Delta}{3\eta N}+\frac{\eps^2}{6}.
\]
By the independence and uniform distribution of $J$, Markov's
inequality and the choice of $N$ imply
\[
 \Prob(\norm{\nabla f(x_J)}>\eps)
 \le\frac8{3}\frac{(E+q+1)T}{qN}+\frac16
 \le\frac8{51}+\frac16=\frac{11}{34}.
\]
Every iteration makes the stated number of value calls whatever the
directions, so the cap holds on every trajectory.
\end{proof}

The first bound follows from Lemma~1 of \citet{laurent2000adaptive}.
\begin{lemma}[Gaussian quadratic tail]
\label{g2:lem:gaussian-tail}
Let $a\sim N(0,I_d)$ and $M\succeq0$, $\norm M_{\rm op}\le L$,
$\tr M\le Lr$, with $r>0$. For $u>0$,
\begin{equation}
 \Prob\!\left(a^\top Ma>L(r+2\sqrt{ru}+2u)\right)\le e^{-u}.
 \label{g2:eq:gaussian-tail}
\end{equation}
Consequently, if $\tr H\le LE$, then
\begin{equation}
 \Prob(a^\top Ha>L(2E+3u))\le e^{-u},\qquad
 \Prob(\norm a^2>2d+3u)\le e^{-u}.
 \label{g2:eq:two-tails}
\end{equation}
\end{lemma}
\begin{proof}
Diagonalize $M/L$, with eigenvalues $w_j\in[0,1]$ and
$\sum_jw_j\le r$. For independent $Z_j\sim N(0,1)$ and
$0<\lambda<1/2$,
\begin{align*}
 \log\E\exp\bigl(\lambda\textstyle\sum_jw_j(Z_j^2-1)\bigr)
 &=\sum_j\left[-\lambda w_j-\tfrac12\log(1-2\lambda w_j)\right]\\
 &\le\frac{\lambda^2\sum_jw_j^2}{1-2\lambda}
 \le\frac{r\lambda^2}{1-2\lambda}.
\end{align*}
The first inequality follows by expanding $-\log(1-z)-z$
into its nonnegative Taylor series and bounding it by
$z^2/[2(1-z)]$, for $0\le z<1$.
Apply the exponential Markov inequality at
$\lambda=\sqrt u/(\sqrt r+2\sqrt u)$; the resulting exponent
at $2\sqrt{ru}+2u$ is $-u$. Adding the mean, at most $r$,
proves \eqref{g2:eq:gaussian-tail}. For the first claim in
\eqref{g2:eq:two-tails} use $2\sqrt{Eu}\le E+u$;
for the second set $M=I_d$, $L=1$, $r=d$ and apply the
same inequality.
\end{proof}

\begin{lemma}[Capped descent for a truncated Gaussian batch]
\label{g2:lem:truncated}
Let $f\in\cB^d_{L,\Delta,E}$, let $H$ be its fixed upper matrix,
and let $x$ satisfy $f(x)\le f(0)$. Fix $\Lambda>0$ and draw
$a_1,\ldots,a_q$ independently from $N(0,I_d)$. Set
\begin{equation}
 v=\frac1q\sum_{i=1}^q
   (a_i^\top\nabla f(x))a_i\,\mathbf 1_{\{a_i^\top Ha_i\le\Lambda\}},
 \qquad \eta=\Lambda^{-1},
 \label{g2:eq:truncate}
\end{equation}

Then $f(x-\eta v)\le f(x)$ and, pathwise,
\begin{equation}
 f(x)-f(x-\eta v)\ge\frac\eta{2q}
 \sum_{i=1}^q(a_i^\top\nabla f(x))^2
          \mathbf 1_{\{a_i^\top Ha_i\le\Lambda\}}.
 \label{g2:eq:truncated-descent}
\end{equation}
If $\Prob(a^\top Ha>\Lambda)\le1/48$, then
\begin{equation}
 \E[f(x)-f(x-\eta v)\mid x]
 \ge\frac{3\eta}8\norm{\nabla f(x)}^2.
 \label{g2:eq:truncated-drift}
\end{equation}
\end{lemma}
\begin{proof}
Write $g=\nabla f(x)$, $\sigma_i=a_i^\top g$ and
$I=\{i:a_i^\top Ha_i\le\Lambda\}$. Cauchy--Schwarz in the
$H$ seminorm, followed by scalar Cauchy--Schwarz, yields
\[
 v^\top Hv\le\frac1{q^2}
   \left(\sum_{i\in I}|\sigma_i|\sqrt{a_i^\top Ha_i}\right)^2
 \le\frac\Lambda q\sum_{i\in I}\sigma_i^2
 =\Lambda g^\top v.
\]
Therefore, by the capped upper model in
Lemma~\ref{u:lem:cap},
\[
 \min\{f(x-\eta v),f(0)\}
 \le f(x)-\eta g^\top v+\tfrac{\eta^2}{2}v^\top Hv
 \le f(x)-\tfrac\eta2g^\top v.
\]
If $g^\top v>0$ the right side is strictly below $f(0)$,
so the minimum must be the actual value. If $g^\top v=0$,
then every retained $\sigma_i$ vanishes, $v=0$, and the
assertion is immediate. This proves \eqref{g2:eq:truncated-descent}.
For $\sigma=a^\top g$ and $p=\Prob(a^\top Ha>\Lambda)$,
Cauchy--Schwarz and the normal fourth moment give
\[
 \E[\sigma^2\mathbf 1_{\{a^\top Ha>\Lambda\}}]
 \le\sqrt{\E\sigma^4\,p}
 =\sqrt{3p}\norm g^2\le\tfrac14\norm g^2.
\]
Independence and the preceding pathwise inequality yield
\eqref{g2:eq:truncated-drift}.
\end{proof}

\begin{theorem}[Gaussian two-point method on $\mathcal B$]
\label{g2:thm:sublevel}
Let $f\in\cB^d_{L,\Delta,E}$, $1\le E\le d$ and
$T=L\Delta/\eps^2\ge1$. For any integer $q\ge1$ define the public
parameters
\begin{align}
 W&=E+\log(1000qET),&N&=\lceil256TW\rceil,\nonumber\\
 u&=\log(48qN),&m&=2E+3u,\nonumber\\
 R&=2d+3u,&\eta&=(Lm)^{-1},\label{g2:eq:sublevel-parameters}\\
 \mu&=\frac{\eps}{4L\sqrt R(1+\eta LR)^N}.& &\nonumber
\end{align}
Run the update \eqref{g2:eq:update}, without an acceptance test, with
this fixed $\eta,\mu,N$ and either forward or central differences.
Then, for every such $f$,
\begin{equation}
 \Prob(\norm{\nabla f(x_J)}>\eps)\le\frac{13}{56}<\frac13,
 \qquad Q\le2qN\le512qT\bigl(E+\log(1000qET)\bigr)+2q.
 \label{g2:eq:sublevel-guarantee}
\end{equation}
In particular, for $q=1$ this is an
$O(ET+T\log(ET))$ value bound.
For $q=1$ and $E\ge\log(1000ET)$, the chosen step size has order
$1/(LE)$ and the call bound is $O(ET)$, with absolute constants.
\end{theorem}
\begin{proof}
Let $H$ be the unknown matrix witnessing membership in $\cB$.
Its use below is confined to the analysis; the algorithm's
parameters depend only on $d,L,\Delta,\eps,E,q$.
Since $TW\ge1$, $N\le257TW$. By
$\log(48\cdot257)<10$, $\log(qT)\le\log(qET)\le W$,
$\log W\le W$, and $10\le2W$, we have
\begin{equation}
 u\le10+\log(qT)+\log W\le4W,
 \qquad m=2E+3u\le14W.
 \label{g2:eq:parameter-check}
\end{equation}
Lemma~\ref{g2:lem:gaussian-tail} applied to $H$ gives
$\Prob(a^\top Ha>Lm)\le e^{-u}=1/(48qN)$.

Construct an auxiliary process $z_0=0$, and, for
$t=0,\ldots,N-1$, define $z_{t+1}$ by the truncated
\emph{directional derivative} step \eqref{g2:eq:truncate} with
$\Lambda=Lm$ and the same Gaussian directions as the actual
algorithm. This process is only a proof device; the actual
algorithm evaluates neither directional derivatives nor $H$.
By Lemma~\ref{g2:lem:truncated}, its iterates satisfy
$f(z_t)\le f(0)$ pathwise and
\[
 \E[f(z_t)-f(z_{t+1})\mid z_t]
 \ge\frac{3\eta}8\norm{\nabla f(z_t)}^2.
\]
Telescoping and using the initial gap yield
\begin{equation}
 \Prob\!\left(\norm{\nabla f(z_J)}>\frac{7\eps}{8}\right)
 \le\frac{512}{147}\frac{mT}{N}
 \le\frac4{21}.
 \label{g2:eq:aux-failure}
\end{equation}
In the first inequality we used the second-moment bound
$N^{-1}\sum_t\E\norm{\nabla f(z_t)}^2
 \le8\Delta/(3\eta N)$ and Markov's inequality;
the last inequality follows from \eqref{g2:eq:parameter-check}
and $N\ge256TW$.

Let $\mathcal G$ be the event that, for every one of the
$qN$ generated directions,
\begin{equation}
 a_{t,i}^\top Ha_{t,i}\le Lm,
 \qquad\norm{a_{t,i}}^2\le R.
 \label{g2:eq:good-event}
\end{equation}
Both Gaussian bounds in \eqref{g2:eq:two-tails} and a union
bound imply $\Prob(\mathcal G^c)\le1/24$.
On $\mathcal G$, none of the auxiliary directions is removed.
We compare its entire trajectory to that of the actual finite
radius algorithm. Write $D_t=\norm{x_t-z_t}$ and
$A_t=q^{-1}\sum_i a_{t,i}a_{t,i}^\top$. On $\mathcal G$,
$\norm{A_t}_{\rm op}\le R$ and
\[
 \norm{q^{-1}\sum_i(s_{t,i}-a_{t,i}^\top\nabla f(x_t))a_{t,i}}
 \le\frac{L\mu R^{3/2}}2
\]
by \eqref{g2:eq:error}. Global $L$-smoothness of $f$ gives
\[
 D_{t+1}\le(1+\eta LR)D_t+\frac{\eta L\mu R^{3/2}}2,
 \qquad D_0=0.
\]
Summing the geometric recursion,
\begin{equation}
 \max_{0\le t\le N}D_t
 \le\frac{\mu\sqrt R}{2}\bigl((1+\eta LR)^N-1\bigr)
 \le\frac\eps{8L}.
 \label{g2:eq:path-coupling}
\end{equation}
Thus, on $\mathcal G$,
$\norm{\nabla f(x_J)}\le\norm{\nabla f(z_J)}+\eps/8$.
Combining this deterministic implication with
\eqref{g2:eq:aux-failure} and the union bound,
\[
 \Prob(\norm{\nabla f(x_J)}>\eps)
 \le\frac4{21}+\frac1{24}=\frac{13}{56}<\frac13.
\]
Exactly $N$ iterations are run, and each consumes at most $2q$ calls. Finally,
$N\le256TW+1$ gives the asserted call cap.
\end{proof}

The radius $\mu$ in \eqref{g2:eq:sublevel-parameters} is exponentially small
in $N$. Exact values allow it, and the proof needs it: with a nonzero radius a
step can raise $f$ slightly even when $a^\top Ha$ is small, so the iterates are
compared with the auxiliary process along the whole run. For
$E<\log(1000ET)$, whether a step of order $1/(LE)$ without an acceptance test
gives $O(ET)$ calls on $\cB$ remains open. With the acceptance test,
Theorem~\ref{b:thm:monotone} gives $O(ET)$; its comparison uses the value
$f(y_t)$, which the forward method without the test also queries as the next
base value $f(x_{t+1})$.

\section{Direction Selection and the Stage Argument}
\label{a:sec:geometry}
\subsection{Direction selection by finite partial coloring}
The selection lemma uses the classical finite partial coloring argument \citep{spencer1985six}: quantize linear forms of sign vectors, find a large common fiber, and subtract two distant members.

\begin{lemma}[Finite entropy facts; see \citealp{cover2006elements}]\label{a:lem:entropy}
For a finitely supported probability mass function $p$, write
$H(p)=\sum_{j:p_j>0}p_j\log(1/p_j)$. If reference masses $q_j>0$ on the support of $p$ have total mass at most one, then
$H(p)\le\sum_jp_j\log(1/q_j)$. Joint entropy is at most the sum of marginal entropies, and some atom has mass $\ge\exp[-H(p)]$.
\end{lemma}
\begin{proof}
Use $\log t\le t-1$ in
$\sum_{p_j>0}p_j\log(q_j/p_j)\le\sum_{p_j>0}q_j-1\le0$.
For subadditivity choose the product of the marginal laws as reference; it is positive on the joint support. Finally $H(p)\ge\log(1/\max_jp_j)$.
\end{proof}

\begin{lemma}[Strip selection]
\label{a:lem:selection}
If $S\subset\mathbb R^d$ has dimension $p\ge1$ and $\|a_i\|\le R$ for $1\le i\le M$, there is a unit $u\in S$ such that
\begin{equation}\label{a:eq:selection}
 \max_{i\le M}|\langle a_i,u\rangle|
 \le\frac{13}{3}R\sqrt{\frac{\log(e+M/p)}p}.
\end{equation}
\end{lemma}
\begin{proof}
If $M<p$ or $R=0$, choose a unit vector $u\in S$ orthogonal to the projections of $a_1,\ldots,a_M$ onto $S$, giving $\max_{i\le M}|\langle a_i,u\rangle|=0$. Assume now that $s:=M/p\ge1$ and $R>0$. Use orthonormal coordinates on $S$ and project the $a_i$ to $S$. Let $X$ be uniform on $\{-1,1\}^p$; set $a=\frac72\log(e+s)$, $t=\sqrt{2a}$,
$Y_i=\langle a_i,X\rangle/R$, and
$K_i=\lfloor Y_i/(2t)+1/2\rfloor$. Since $(e+1)^7>e^9$, we have $a\ge\frac72\log(e+1)>\frac92$. The inequality
$\log\cosh z\le z^2/2$ follows by integrating $\tanh z\le z$ for $z\ge0$ and using evenness. Independence gives
$\mathbb E e^{\lambda Y_i}\le e^{\lambda^2/2}$, hence
$\mathbb P(|Y_i|\ge b)\le2e^{-b^2/2}$, as in Hoeffding's inequality \citep{hoeffding1963probability}. Choose an integer $N$ bounding every value $|K_i(X)|$. Since $(2k-1)^2\ge k$ for every integer $k\ge1$,
\[
 \mathbb P(|K_i|\ge k)\le2e^{-(2k-1)^2a}\le2e^{-ka},\qquad
 \mathbb E|K_i|=\sum_{k=1}^N\mathbb P(|K_i|\ge k)<\frac{2e^{-a}}{1-e^{-a}}.
\]
Take reference masses $q_k=Z^{-1}e^{-a|k|}$ on $\{-N,\ldots,N\}$. Finite geometric summation gives $\log Z\le Z-1<2e^{-a}/(1-e^{-a})$. Lemma~\ref{a:lem:entropy} implies
\begin{equation}\label{eq:selection-entropy}
 H(K_i)\le\log Z+a\mathbb E|K_i|
 <\frac{2(a+1)e^{-a}}{1-e^{-a}}.
\end{equation}
For $s\ge1$ and $a=\frac72\log(e+s)>\frac92$, the derivative of $\log s+\log(a+1)-a$ with respect to $s$ is at most $\frac1s-\frac{63}{22(e+s)}$, which is positive only for $s<\frac{22}{41}e<\frac32$, and across all $s\ge1$ one has $sH(K_i)<\frac3{25}$. By subadditivity, the joint entropy of $K=(K_1,\ldots,K_M)$ satisfies $H(K)\le\sum_{i=1}^MH(K_i)<\frac{3M}{25s}=\frac3{25}p$. By Lemma~\ref{a:lem:entropy}, some common fiber $F\subseteq\{-1,1\}^p$ of $K$ has cardinality $|F|\ge2^pe^{-H(K)}$.

Let $W$ be uniform on $F$ with mean $\mu=\mathbb EW\in[-1,1]^p$. Applying Lemma~\ref{a:lem:entropy} to each coordinate $W_j$ with reference masses $e^{\pm\mu_j}/(2\cosh\mu_j)$ gives
\[
 H(W_j)\le\log(2\cosh\mu_j)-\mu_j^2\le\log2-\tfrac12\mu_j^2.
\]
Subadditivity of entropy across the $p$ coordinates gives $p\log2-H(K)\le\log|F|=H(W)\le p\log2-\frac12\|\mu\|^2$, whence $\|\mu\|^2\le2H(K)<\frac6{25}p$. Two independent uniform draws $X,X'$ from $F$ satisfy
\[
 \mathbb E\|X-X'\|^2=2\bigl(p-\|\mu\|^2\bigr)>\frac{38}{25}p,
\]
so some pair $X,X'\in F$ satisfies $\|X-X'\|>\sqrt{(38/25)p}$, while equal bins give $|\langle a_i,X-X'\rangle|<2tR=\sqrt{28\log(e+s)}\,R$. Normalizing $X-X'$ yields a unit vector $u\in S$ with squared projection bound at most $\frac{350}{19}R^2\log(e+s)/p<\bigl(\frac{13}{3}R\bigr)^2\log(e+s)/p$.
\end{proof}

\begin{lemma}[Avoiding spherical caps by a union bound]\label{a:lem:cap-union}
Let $R>0$ and $\tau\ge0$. Let $S$ have dimension $p\ge1$, and let $\|a_i\|\le R$. If
$2M\exp(-p\tau^2/(2R^2))<1$, there is a unit $u\in S$ with
$|\langle u,a_i\rangle|\le\tau$ for all $i$.
\end{lemma}
\begin{proof}
A uniform sphere direction $u$ satisfies
$\mathbb P(|\langle u,a\rangle|>\tau)\le
2e^{-p\tau^2/(2R^2)}$.
Indeed its odd moments vanish and its $2k$-th projection moment is
$\|P_Sa\|^{2k}(2k-1)!!/[p(p+2)\cdots(p+2k-2)]$, where $P_S$ is the orthogonal projection onto $S$.
The moment-generating series is dominated by
$e^{\lambda^2R^2/(2p)}$ and exponential Markov proves the tail. For the moment identity, polar coordinates for an isotropic Gaussian give an independent uniform direction and radius; scalar Gaussian moments are $(2k-1)!!$ and the radius moments are $p(p+2)\cdots(p+2k-2)$ by integration by parts, including the $p=1$ case. The union bound now leaves positive probability of avoiding all the caps.
\end{proof}

\subsection{One transcript argument}
The following lemma handles the deterministic lower bounds. It handles values, rank-priced linear feedback, closing calls,
unqueried outputs, and the identification of the simulated run with the run
on one fixed function.

\begin{lemma}[Stage argument]
\label{lem:stage}\label{a:lem:stage}
Let $F_U$ range over orthonormal $m$-frames $U=(u_1,\ldots,u_m)$ in $\R^d$,
and let $z:\R^d\to\R^d$ satisfy $\norm{z(x)}\le R$. Suppose that for each
partial frame there are a prefix function $r_j$ and a coefficient $c_j$,
depending only on $u_1,\ldots,u_{j-1}$, such that
\[
 F_U(x)=r_j(x),\qquad \nabla F_U(x)=\nabla r_j(x)-c_j(x)u_j
\]
whenever $\ip{u_j}{z(x)}=0$ and $|\ip{u_k}{z(x)}|\le\tau$ for all $k>j$.
Assume also that $\ip{u_j}{z(x)}=0$ for some $j$ implies
$\norm{\nabla F_U(x)}>b$.

Let $m,h\ge1$ and $Q\ge0$ be integers and put $p=d-m-h+1\ge1$. Assume that
for every subspace $S$ of dimension at least $p$ and every list of at most
$Q$ points $a_i$ of norm at most $R$ there is a unit $u\in S$ with
$|\ip u{a_i}|\le\tau$ for all $i$. Then every deterministic algorithm whose
calls $(x,V)$ return $(F_U(x),V^\top\nabla F_U(x))$ at cost $1+\operatorname{rank}V$,
and which on every transcript makes at most $Q$ calls of total cost below
$hm$, outputs a point with gradient norm greater than $b$ for at least one
fixed $F_U$. Queries, matrices, and the output are unrestricted. The
geometric assumption holds, for $\tau\ge0$, under any one of the conditions
\[
 Q<p;\qquad
 R>0\ \text{ and }\ 2Q\exp\Bigl(-\frac{p\tau^2}{2R^2}\Bigr)<1;\qquad
 \frac{13}{3}R\sqrt{\frac{\log(e+Q/p)}p}\le\tau.
\]
\end{lemma}

\begin{proof}
Maintain the selected directions, the points $z(x)$ of closed stages, and
the pending calls of the current stage. If the pending calls have total cost
$\ell<h$, their points $z(x)$ and matrix column spaces together span at most
$\ell$ dimensions. For a new call $(x,V)$ of cost $w=1+\operatorname{rank}V$:
if $\ell+w<h$, return the prefix reply $(r_j(x),V^\top\nabla r_j(x))$ and
record the call as pending. If $\ell+w\ge h$, select $u_j$ before answering:
perpendicular to the preceding frame, to all pending constraints, and to
$z(x)$, and with inner products at most $\tau$ in magnitude with all points
of closed stages. These exact constraints have dimension at most
$(m-1)+(h-1)+1$, which leaves dimension at least $p$, so the geometric
assumption supplies $u_j$. The current column space is not constrained.
Return $(r_j(x),V^\top[\nabla r_j(x)-c_j(x)u_j])$, close the stage, move its
points to the list of past points, and reset $\ell=0$. Each closed stage consumes
at least $h$ units of cost, so at most $m-1$ stages close.

At the output $\widehat x$, select the next direction in the same way, with
$z(\widehat x)$ in place of a closing query point. Move the pending points to
the list of past points and complete the frame with directions perpendicular to
their predecessors and with inner products at most $\tau$ at every past
point; there are at least $p$ residual dimensions and at most $Q$ past
points. For the completed frame, every call of stage $j$ has zero $j$th
coordinate and later coordinates bounded by $\tau$, so its reply is the
prefix reply together with the term of the current direction. That term vanishes for
pending calls, whose column spaces were among the orthogonality constraints,
and closing calls received it explicitly. Induction over the deterministic
transcript shows that the simulated run is the run on this single function.
Its output has a zero hidden coordinate, so its gradient norm exceeds $b$.

Exact orthogonality to fewer than $p$ past points is possible inside a
$p$-dimensional subspace; the other two conditions are
Lemmas~\ref{a:lem:cap-union} and~\ref{a:lem:selection}, whose bounds only improve
in larger subspaces.
\end{proof}

A budget asserted only on functions in the class reduces to this form for whole transcripts by truncating before any call that would exceed the budget: every
finite simulated prefix extends to a fixed function in the class, on which the
algorithm terminates within its budget, so the truncation never changes that
function's run. This applies to call caps and cost caps even when their
worst cases occur on different functions.

\section{The Chain and the Radial Compression}
\label{a:sec:family}
The deterministic witnesses share the form $F_{U,R,Q}(x)=g_m(U^\top\rho_R(x))+Q(\|x\|)$. Only the public radial parameters differ. All scalar gates are piecewise polynomial; no convolution is needed.
\subsection{One scalar chain}
\label{app:chain}

Define
\[
 \varphi(t)=
 \begin{cases}
 0,&t\le-5/4,\\
 2(t+5/4)^2,&-5/4<t<-1,\\
 t+9/8,&-1\le t\le1,\\
 9/4-2(5/4-t)^2,&1<t<5/4,\\
 9/4,&t\ge5/4,
 \end{cases}
 \qquad
 \psi(t)=
 \begin{cases}
 0,&t\le1/2,\\
 8(t-1/2)^2,&1/2<t\le3/4,\\
 1-8(1-t)^2,&3/4<t<1,\\
 1,&t\ge1,
 \end{cases}
\]
and, for $m\ge1$,
\begin{equation}
 g_m(y)=-\varphi(y_1)+\sum_{i=2}^m
 \bigl[\psi(-y_{i-1})\varphi(-y_i)-\psi(y_{i-1})\varphi(y_i)\bigr].
 \label{eq:chain}
\end{equation}
The same chain is used for the lower bounds with and without a trace budget; only radial parameters change.

\begin{lemma}[The scalar chain]
\label{lem:chain}
Let $p=\nabla g_m(y)$ and $I(y)=\{i:|y_i|\le5/4\}$. The function $g_m$ is
$C^{1,1}$ and bounded, and
\[
 \norm p\le10\sqrt m,\qquad
 \Lip(\nabla g_m)\le42,\qquad
 g_m(0)-\inf g_m\le\frac94\Bigl(m-\frac12\Bigr).
\]
Every $p_i\le0$, and $|y_i|\ge5/4$ implies $p_i=0$. If $j$ is the first
index with $|y_j|<1$, then $p_j\le-1$. On each smooth piece, the Hessian
$B=\nabla^2g_m(y)$ is supported on $I(y)\times I(y)$ and satisfies
$\norm B_{\rm op}\le42$. Consequently,
\begin{equation}
 \norm{By}\le44\sqrt m,\qquad
 |y^\top By|\le44m,\qquad
 |y^\top p|\le\frac{31}{4}m.
 \label{eq:chain-support}
\end{equation}
\end{lemma}

\begin{proof}
The formulas have matching values and first derivatives at every junction.
They give
\[
 0\le\varphi\le9/4,\quad 0\le\varphi'\le1,\quad
 |\varphi''|\le4,\quad
 0\le\psi\le1,\quad 0\le\psi'\le4,\quad |\psi''|\le16
\]
on their smooth pieces. Moreover, $\varphi'$ is even and equals one on
$[-1,1]$, so $\varphi''(t)$ vanishes outside $1\le|t|\le5/4$, whereas
$\psi'(\pm t)$ and $\psi''(\pm t)$ are supported on $1/2\le|t|\le1$. At
most one of $\psi(t),\psi(-t)$ is nonzero. With $\gamma_0=1$ and
$\gamma_{i-1}=\psi(-y_{i-1})+\psi(y_{i-1})\in[0,1]$ for $i>1$,
\begin{equation}
 p_i=-\gamma_{i-1}\varphi'(y_i)
 -\mathbf 1_{\{i<m\}}
 \bigl[\psi'(-y_i)\varphi(-y_{i+1})+\psi'(y_i)\varphi(y_{i+1})\bigr].
 \label{eq:chain-gradient}
\end{equation}
This proves the sign, the support, the coordinate bound $|p_i|\le10$, and
the barrier: at the first index with $|y_j|<1$ the preceding gate is
saturated. Because $|s\varphi'(s)|\le1$ on $\R$ (as $4s(5/4-s)=1-4(s-1)(s-1/4)\le1$ on $[1,5/4]$) and $t\psi'(t)\le3$ on $[1/2,1]$ (as $16t(1-t)=3-16(t-3/4)(t-1/4)\le3$ on $[3/4,1]$), each coordinate satisfies $|y_ip_i|\le1+3(9/4)=31/4$, which sums to $|y^\top p|\le(31/4)m$.

The Hessian $B$ is symmetric tridiagonal with $|B_{i,i-1}|\le4$, and $B_{i,i-1}\ne0$ requires $|y_{i-1}|\le1$. Since the supports of $\varphi''(y_i)$ and $\psi''(\pm y_i)$ meet only at $|y_i|=1$, there are two cases for row $i$. If $|y_i|\le1$, then $\varphi''(y_i)=0$, and with $s=\pm y_{i+1}$ one has
\begin{equation}\label{eq:row-sum-identity}
 |B_{i,i-1}|+|B_{ii}|+|B_{i,i+1}|\le4+16\varphi(s)+4\varphi'(s)\le42,
\end{equation}
because $4+16\varphi(s)+4\varphi'(s)$ is at most $4+16(17/8)+4=42$ for $s\le1$, equals $42-32(s-1)^2\le42$ on $[1,5/4]$, and equals $40$ for $s\ge5/4$. In the same case $|y_{i-1}|\le1$, $|y_i|\le1$, and $|y_{i+1}|\varphi'(\pm y_{i+1})\le1$, so $|(By)_i|\le4\cdot1+36\cdot1+4\cdot1=44$ and $|y_i(By)_i|\le44$. If $1<|y_i|\le5/4$, then $\psi'(\pm y_i)=\psi''(\pm y_i)=0$, so $B_{i,i+1}=0$, $|B_{ii}|\le4$, the absolute row sum is at most $8$, $|(By)_i|\le4(5/4)+4\cdot1=9$, and $|y_i(By)_i|\le45/4<44$. Thus $\|B\|_{\rm op}\le42$, and \eqref{eq:chain-support} holds. Matching derivatives extend $\|B\|_{\rm op}\le42$ to $\Lip(\nabla g_m)\le42$ by integration over the finitely many polynomial pieces of a line segment. Finally $g_m(0)=-9/8$, and each of the $m$ summands of $g_m$ is at least $-9/4$.
\end{proof}

\subsection{Radial compression and shifted Huber potentials}
\label{app:radial}

For $R>0$ write
\begin{equation}
 a(x)=(1+\norm x^2/R^2)^{-1/2},\qquad\rho_R(x)=a(x)x.
 \label{eq:rho}
\end{equation}
For $\lambda,s>0$ and $r_0\ge0$, define the shifted
radial Huber potential \citep{huber1964robust} through its derivative,
\begin{equation}
 Q'(r)=\lambda\min\{(r-r_0)_+,s\},\qquad Q(0)=0,\qquad q(x)=Q(\norm x).
 \label{eq:shifted-huber}
\end{equation}
It vanishes on the ball of radius $r_0$, is quadratic across an annulus of
width $s$, and is linear afterwards. The construction with a trace budget below uses a related sum of two such potentials; the unrestricted choice uses one flat on an inner ball.

\begin{lemma}[Radial derivatives]
\label{lem:radial}
The potential \eqref{eq:shifted-huber} is nonnegative, convex, coercive, and
$C^{1,1}$, with $\norm{\nabla q}\le\lambda s$ and $\Lip(\nabla q)\le\lambda$.
For $z=\rho_R(x)$,
\begin{equation}
 D\rho_R(x)=a(x)\Bigl(I-\frac{zz^\top}{R^2}\Bigr),\qquad
 \norm{D\rho_R(x)}_{\rm op}\le1,\qquad
 \norm{D^2\rho_R(x)[v,w]}\le\frac{2}{\sqrt3R}\norm v\norm w.
 \label{eq:radial-derivatives}
\end{equation}
\end{lemma}

\begin{proof}
For $r>0$, $\nabla q=Q'(r)x/r$. On every smooth piece its radial and
tangential Hessian eigenvalues are $Q''(r)$ and $Q'(r)/r$, both in
$[0,\lambda]$, because $Q'$ is nondecreasing, $Q''\in\{0,\lambda\}$, and
$Q'(r)\le\lambda r$. The gradient is continuous across the junction spheres,
and integrating along line segments gives the Lipschitz bound. The other
properties follow from $Q'\ge0$ and $Q'(r)=\lambda s$ for large $r$.

Differentiation gives the formula for $D\rho_R$, whose eigenvalues are $a$
tangentially and $a^3$ radially. A second differentiation gives
\[
 D^2\rho_R(x)[v,w]
 =-\frac{a^3}{R^2}\bigl[(x^\top w)v+(x^\top v)w+(v^\top w)x\bigr]
 +\frac{3a^5}{R^4}(x^\top v)(x^\top w)x.
\]
For a unit $v$ and $x\ne0$, put $t=\norm x/R$, $C=(x^\top v/\norm x)^2$, and
$z=t^2/(1+t^2)$. Then
\[
 \norm{D^2\rho_R(x)[v,v]}^2
 =\frac{t^2}{R^2(1+t^2)^3}\bigl[1+(8-6z)C+(9z^2-12z)C^2\bigr].
\]
For fixed $C\in[0,1]$ the bracket is convex in $z\in[0,1]$; at the endpoints
it equals $1+8C\le9$ and $1+2C-3C^2\le4/3$. Hence the diagonal norm is at
most $3t/[R(1+t^2)^{3/2}]\le2/(\sqrt3R)$, and it vanishes at $x=0$.
Polarization, $D^2\rho_R[v,w]=\frac14(D^2\rho_R[v+w,v+w]-D^2\rho_R[v-w,v-w])$,
and the parallelogram law give the same bound for unit $v,w$.
\end{proof}

\subsection{Integration on finitely many smooth pieces}
\label{app:finite-piece}

\begin{lemma}[Integration over finitely many pieces]
\label{lem:finite-piece}
Let $F:\R^d\to\R$ be continuously differentiable. Suppose that along every
segment $x+tv$, $0\le t\le1$, its gradient is continuously differentiable
except at finitely many points, and that on each remaining interval
\[
 \frac{d}{dt}\nabla F(x+tv)=B(t)v,\qquad
 \norm{B(t)}_{\rm op}\le L_0,\qquad B(t)\preceq H,
\]
where $B(t)$ is symmetric and $H$ is a fixed symmetric matrix. Then
$\Lip(\nabla F)\le L_0$ and
$F(x+v)\le F(x)+\nabla F(x)^\top v+\frac12v^\top Hv$.
\end{lemma}

\begin{proof}
Apply the fundamental theorem of calculus on each interval and add; the
endpoint terms cancel by continuity of the gradient. This gives
$\norm{\nabla F(x+tv)-\nabla F(x)}\le L_0t\norm v$ and
$v^\top[\nabla F(x+tv)-\nabla F(x)]\le t\,v^\top Hv$. Integrating the second
inequality over $t\in[0,1]$ proves the upper model.
\end{proof}

The lemma applies to every function built from \eqref{eq:chain},
$\rho_R$, and \eqref{eq:shifted-huber}. On a segment, each compressed
coordinate has the form $(\alpha+\beta t)/(1+\norm{x+tv}^2/R^2)^{1/2}$.
Squaring shows that it equals a nonzero scalar junction value at most twice,
unless it does so identically; in that case its derivative along the segment
vanishes, and either adjacent one-sided second derivative gives the correct expression from the chain rule. Crossings of a Huber junction sphere are roots of a
quadratic polynomial as well. The resulting finite partition makes the
gradient smooth on each open interval.

\subsection{Sublevel parameter choice}
\begin{theorem}[Wrapper with a sublevel trace bound]
\label{w:thm:wrapper}
Let $m\ge1$ and $R^2\ge64m$. Set $r_0=\frac23R$, $\mu=11m/R^2$, and define a public radial potential $q(x)=Q(\norm x)$ by
\begin{equation}
 Q(0)=0,\qquad Q'(r)=\mu\min\{r,r_0\}+\min\{(r-r_0)_+,1\}.
 \label{w:eq:potential}
\end{equation}
For an orthonormal frame $U$ set
\[
 F_U(x)=g_m(U^\top\rho_R(x))+q(x).
\]
Then $F_U$ is coercive, nonconvex, and $C^{1,1}$, and
\begin{align}
 \Lip(\nabla F_U)&<45,&\quad \norm{\nabla F_U}&\le12\sqrt m,\label{w:eq:regularity}\\
 F_U(0)-\inf F_U&\le\tfrac94(m-\tfrac12),&
 \{F_U\le F_U(0)\}&\subset\{\norm x<\tfrac23R\}.\label{w:eq:sublevel}
\end{align}
On the initial sublevel, almost everywhere, one fixed matrix satisfies
\begin{equation}
 \nabla^2 F_U\preceq K_U:=54UU^\top+\frac{80m}{R^2}I,
 \qquad \|K_U\|_{\rm op}<56.
 \label{a:eq:wrapper-matrix}
\end{equation}
Consequently,
\begin{equation}
 \tr[(\nabla^2F_U)_+]
 \le54m+\frac{80dm}{R^2}.
 \label{w:eq:wrapper-trace}
\end{equation}
If any compressed coordinate has magnitude below one, then at any $x\in\R^d$,
\begin{equation}
 \norm{\nabla F_U(x)}>3/4.
 \label{w:eq:barrier}
\end{equation}
Finally its exact prefix replies have the following form: if $y=U^\top\rho_R(x)$, $y_j=0$, and $|y_k|\le1/2$ for $k>j$, then, with $r_j(x)=g_m(y_1,\ldots,y_{j-1},0,\ldots,0)+q(x)$,
\begin{equation}
 F_U(x)=r_j(x),\qquad
 \nabla F_U(x)=\nabla r_j(x)-a(x)\gamma_{j-1}(x)u_j.
 \label{w:eq:prefix}
\end{equation}
\end{theorem}
\begin{proof}
We have $R\ge8\sqrt m\ge8$, $\mu\le11/64$, and $\mu r_0=22m/(3R)\le(11/12)\sqrt m<\sqrt m$.
The derivative $Q'$ is continuous, nonnegative and nondecreasing, with slopes $\mu$, $1$, and $0$ on its successive intervals. Also $Q'(r)\le r$, and its final value is $\mu r_0+1<2\sqrt m$. The radial and tangential Hessian eigenvalues of $q$ are $Q''(r)$ and $Q'(r)/r$, so its gradient is globally $1$-Lipschitz and bounded by $2\sqrt m$. Thus
\[
 \Lip(\nabla F_U)\le42+\frac{20\sqrt m}{\sqrt3R}+1
 \le43+\frac5{2\sqrt3}<45.
\]
The gradient bound, gap, and coercivity follow from the chain bounds, $q\ge0$, and its positive linear growth. Moreover $Q(r_0)=\mu r_0^2/2=22m/9$, so for $r\ge r_0$,
$F_U(x)\ge-9m/4+22m/9=7m/36>0>F_U(0)=-9/8$. This proves the sublevel inclusion. Consequently throughout the indicated sublevel the penalty is $\mu\norm x^2/2$.

For the sublevel Hessian bound write $p=\nabla g_m(y)$, $B=\nabla^2g_m(y)$, $\alpha=U^\top v$, and $\beta=a^2x^\top v/R^2$. Then $U^\top D\rho_Rv=a\alpha-\beta y$, and on $\{\norm x<\frac23R\}$ one has $|\beta|\le\frac{2/3}{1+4/9}\frac{\norm v}{R}=\frac6{13}\frac{\norm v}{R}$ and $a^2\ge\frac9{13}$. The second term of the chain rule for the Hessian equals
\[
 \ip{Up}{D^2\rho_R[v,v]}
 =-2a\beta p^\top\alpha+
       (3\beta^2-a^2\norm v^2/R^2)y^\top p.
\]
Since $0\le3\beta^2\le\frac{108}{169}\frac{\norm v^2}{R^2}<\frac9{13}\frac{\norm v^2}{R^2}\le\frac{a^2\norm v^2}{R^2}\le\frac{\norm v^2}{R^2}$, we have $|3\beta^2-a^2\norm v^2/R^2|\le\norm v^2/R^2$. Combining the cross terms $-2a\beta\alpha^\top(By+p)$ via Young's inequality with parameter $\gamma=2/9$ and using $\norm B_{\rm op}\le42$, $\norm{By}+\norm p\le54\sqrt m$, $|y^\top By|\le44m$, and $|y^\top p|\le31m/4$ gives
\[
 (a\alpha-\beta y)^\top B(a\alpha-\beta y)+\ip{Up}{D^2\rho_R[v,v]}
 \le\bigl(42+54\gamma\bigr)\norm\alpha^2+\Bigl[\Bigl(\frac{54}{\gamma}+44\Bigr)\frac{36}{169}+\frac{31}{4}\Bigr]\frac{m}{R^2}\norm v^2.
\]
Adding $\mu I=(11m/R^2)I$ and noting that $42+54(2/9)=54$ and $11+287(36/169)+31/4=54003/676<80$ gives $\nabla^2F_U\preceq54UU^\top+(80m/R^2)I$ on the initial sublevel, with operator norm at most $54+80/64<56$. The same calculation bounds the absolute value of the quadratic form. For symmetric $A\preceq H$, $H\succeq0$, one has $\tr A_+\le\tr H$: trace against the positive spectral projection of $A$. This proves \eqref{w:eq:wrapper-trace}. For global smoothness one may directly use
$\nabla(g_m\circ U^\top\rho_R)=D\rho_R^\top U\nabla g_m(U^\top\rho_R)$:
$D\rho_R$ has norm at most one and Lipschitz constant $2/(\sqrt3 R)$, while $\nabla g_m$ is bounded and $42$-Lipschitz. The product rule for differences gives the displayed global bound. For the trace estimate, the finitely many scalar junction hyperplanes have null preimages under $U^\top\rho_R$, a smooth submersion; the radial junction spheres are null as well.

For the barrier on the whole space, let $j$ be the first unfinished coordinate, so $p_j\le-1$ and $|y_j|<1$. Write $r=\norm x$, $w=\norm{\rho_R(x)}=ar$, and $g=\nabla F_U(x)=D\rho_R(x)Up+\frac{Q'(r)}{r}x$. If $w^2\le\frac{13}{4}m$, then since $w(r_0)^2=\frac4{13}R^2\ge\frac{256}{13}m>\frac{13}{4}m$, we have $r<r_0$, $Q'(r)/r=\mu\le11/64$, and $a^2=1-w^2/R^2\ge243/256$. Since $D\rho_R(x)^{-1}u_j=a^{-1}(u_j+a^{-1}y_jx/R^2)$ satisfies $\ip{D\rho_R(x)^{-1}u_j}{x}=a^{-4}y_j$,
\begin{equation}\label{eq:inner-inv-barrier}
 |\ip{D\rho_R(x)^{-1}u_j}{g}|=|p_j+a^{-4}\mu y_j|\ge1-\frac{11}{64}\Bigl(\frac{256}{243}\Bigr)^2,
\end{equation}
while $\norm{D\rho_R(x)^{-1}u_j}^2=a^{-2}[1+(1+a^2)a^{-4}y_j^2/R^2]\le\frac{256}{243}[1+\frac{2}{64}(\frac{256}{243})^2]$ because $1+a^2\le2$ and $y_j^2/R^2<1/64$, giving $\norm g>3/4$. Next suppose $w^2\ge\frac{13}{4}m$ and $r\le r_0+1$. Let $I=\{i:|y_i|\le5/4\}$ and $W=\operatorname{span}\{u_i:i\in I\}$. Since $g=aUp+cx$ for a scalar $c$ and $D^2:=\norm{P_{W^\perp}\rho_R(x)}^2\ge w^2-\frac{25}{16}m$, projecting $g$ onto the orthogonal directions $u_j$ and $P_{W^\perp}x$ gives
\begin{equation}\label{eq:annulus-proj}
 \norm g^2\ge\frac{a^2p_j^2D^2}{D^2+y_j^2}
 \ge\Bigl(1-\frac{w^2}{R^2}\Bigr)\frac{w^2/m-25/16}{w^2/m-9/16}.
\end{equation}
The right-hand side is concave in $w^2$ on $[\frac{13}{4}m,w(r_0+1)^2]$. At $w^2=\frac{13}{4}m$ it is at least $\frac{243}{256}\cdot\frac{27}{43}=\frac{6561}{11008}>\frac9{16}$; at $r=r_0+1$, since $r/R\le2/3+1/8=19/24$ and $w(r_0+1)^2\ge w(r_0)^2\ge\frac{256}{13}m$, it is at least $\frac{576}{937}\cdot\frac{3771}{3979}>\frac9{16}$. Finally, if $r\ge r_0+1$, then $Q'(r)\ge1$, $w\ge\frac{16}{\sqrt{13}}\sqrt m$, and $v:=D\rho_R(x)Up$ satisfies $|\ip{x/r}{v}|\le\frac{(5/4)\sqrt m}{w}\norm v$ because $D\rho_R$ shrinks the radial component relative to the tangential component; projecting $g=v+Q'(r)x/r$ onto $v^\perp$ gives $\norm g^2\ge1-\frac{25m}{16w^2}\ge\frac{3771}{4096}>\frac9{16}$.

To verify nonconvexity, choose the point on the negative $u_1$ axis with $y_1=-9/8$. Since $R\ge8$, it lies in $\{\norm x<r_0\}$ with $a^2\ge1-81/4096>31/32$, where the second axis derivative is at most $-4(31/32)^3+11/64<0$. Finally, after a zero coordinate every later gate and its derivative vanishes. The chain derivative at coordinate $j$ is $-\gamma_{j-1}$. Since $y_j=0$ implies $u_j^\top x=0$, $D\rho_Ru_j=au_j$. The radial term is public and cancels from prefix differences, proving \eqref{w:eq:prefix}.
\end{proof}

\section{Scaling, Dimension Floor, and Deterministic Lower Bounds}
\label{a:sec:lower}
\begin{lemma}[One scaling rule]\label{a:lem:scaling}\label{lem:transfer}
Let $\Lambda,b>0$, and let $F$ have gradient Lipschitz constant at most $\Lambda$, initial gap at most $D$, and one fixed positive semidefinite Hessian upper matrix $K$ of operator norm at most $\Lambda$, either globally or on its initial sublevel. Assume $\|\nabla F\|>b$ at the points used in a lower bound. For $0<L'\le L$ set
\begin{equation}\label{a:eq:scaling}
 \sigma=\frac{\Lambda\varepsilon}{L'b},\qquad
 \alpha=\frac{\Lambda\varepsilon^2}{L'b^2},\qquad
 f(x)=\alpha F(x/\sigma).
\end{equation}
If $\Lambda D/b^2\le L'\Delta/\varepsilon^2$ and
$L'\tr K/\Lambda\le LE$, then $f\in\mathcal B$, or $f\in\mathcal F$ when the matrix bound is global. The barrier becomes strictly greater than $\varepsilon$ and all oracle costs are preserved.
\end{lemma}
\begin{proof}
The gradient multiplier is $\alpha/\sigma=\varepsilon/b$, the Hessian multiplier is $\alpha/\sigma^2=L'/\Lambda$, and the gap is at most $\alpha D\le\Delta$. The fixed upper matrix becomes $(L'/\Lambda)K$, whose norm is at most $L'\le L$ and whose trace satisfies the hypothesis. Positive scaling maps initial sublevels exactly. A call $(x,V)$ is simulated by $(x/\sigma,V)$ on $F$, multiplying values by $\alpha$ and derivative projections by $\alpha/\sigma$; its rank and cost are unchanged.
\end{proof}

\begin{lemma}[Dimension floor]\label{lem:dimension-floor}\label{a:lem:dimension-floor}
For $T>1/2$ and $1\le E\le d$, $N^{\rm lin}_{\rm det}\ge d$, already on the global fixed-matrix class $\mathcal F$.
\end{lemma}
\begin{proof}
Let $r=\sqrt{2\Delta/L}$ and $f_u(x)=\frac L2(\langle u,x\rangle-r)^2$ for a unit vector $u$. Simulate the reply $(\Delta,0)$ against a deterministic algorithm truncated below cost $d$. The span of all query points and requested column spaces has dimension strictly less than $d$. Choose a perpendicular unit $u$, with its sign arranged so that $\langle u,\widehat x\rangle\le0$. The function $f_u$ reproduces all replies, has upper matrix $Luu^\top$, and its output gradient norm is at least $Lr=\sqrt{2T}\,\varepsilon>\varepsilon$. Truncation is justified as in Lemma~\ref{lem:stage}.
\end{proof}

\begin{theorem}[Multistage lower bound on the main class]\label{a:thm:stages}
If an integer $m\ge2$ satisfies
\begin{equation}\label{a:eq:stage-conditions}
 224m\le T,\qquad 224^2m^2\log(e+m)\le ET,\qquad
 10000m\log(e+m)\le d,
\end{equation}
then both deterministic oracle costs on $\mathcal B$ are at least
$\lfloor d/2\rfloor m$. The witnesses are $C^{1,1}$, globally $L$-smooth, coercive, nonconvex, and globally $2\sqrt{L\Delta}$-Lipschitz.
\end{theorem}
\begin{proof}
Set $\ell=\log(e+m)$, $R^2=d/(156\ell)$, $h=\lfloor d/2\rfloor$,
$p=d-m-h+1$, and use Theorem~\ref{w:thm:wrapper}. Since $156\cdot64=9984<10000$, the last inequality of~\eqref{a:eq:stage-conditions} implies $R^2\ge64m$, $m\le d/128$, and $p\ge63d/128$. For a list of $M\le hm$ past points, $M/p\le64m/63$. Since $\log(e+cm)\le c\log(e+m)$ when $c\ge1$, Lemma~\ref{a:lem:selection} gives squared projection bound at most
\[
 \frac{(13/3)^2R^2}{p}\log(e+M/p)
 \le\frac{169\cdot128\cdot64}{9\cdot63^2\cdot156}=\frac{26624}{107163}<\frac14.
\]
Thus the transcript lemma applies with $\tau=1/2$ and $b=3/4$.
For the fixed matrix~\eqref{a:eq:wrapper-matrix}, since $\ell\ge\log(e+2)>3/2$,
\[
 \frac{\tr K_U}{56}
 =\frac{54m+80\cdot156m\ell}{56}<\frac{3129}{14}m\ell<224m\ell.
\]
Take $L'=224mL/T$, $\Lambda=56$, and apply Lemma~\ref{a:lem:scaling}. We have $L'\le L$, and
$56(9/4)/(3/4)^2=224$, so the condition on the initial gap holds. The trace becomes at most
$L'\,224m\ell\le LE$ by the second inequality of~\eqref{a:eq:stage-conditions}. The normalized matrix also has operator norm at most $L'$, so the instance lies in $\mathcal B$. The global gradient bound is
$\|\nabla f\|\le16\sqrt m\,\varepsilon\le2\sqrt{L\Delta}$ because $T\ge224m$. Positive scaling preserves coercivity and nonconvexity.
\end{proof}

\begin{corollary}[Rounded main lower bound]\label{a:cor:smooth-rate}
If $ET\le d$, then
\begin{equation}\label{a:eq:det-lower}
 N_{\rm det}\ge\max\left\{d,\frac d{672}
 \min\left\{T,\sqrt{\frac{ET}{\log(e+T)}}\right\}\right\}.
\end{equation}
Without the assumption $ET\le d$, $E\ge T\log(e+T)$ implies
$N_{\rm det}\ge dT/30000$.
\end{corollary}
\begin{proof}
Put $M=\min\{T,\sqrt{ET/\log(e+T)}\}$, $z=M/224$, and $m=\lfloor z\rfloor$. If $z<3$, the dimension floor dominates $dz/3$. If $z\ge3$, the first two stage conditions follow from $224m\le M$ and $\log(e+m)\le\log(e+T)$. The third follows from $10000<2\cdot224^2\le m\,224^2$ and $ET\le d$. Moreover $m/z\ge3/4$ and $h/d\ge15/32$, so
$hm\ge(45/128)dz>dz/3=dM/672$.
For the final claim take $m=\lfloor T/10000\rfloor$ and apply the same rounding argument. Its stage conditions follow from $T\log(e+T)\le E\le d$ and $224^2<10000^2$; the dimension floor covers the case $T/10000<3$.
\end{proof}

\begin{corollary}[Matched window and cost ratio]\label{b:cor:aligned}\label{a:cor:ratio}
For $ET\le d$, write $\ell=\log(e+T)$. Then
\[
 \frac{d}{15456(E+\sqrt{ET})\sqrt\ell}
 \le\frac{N_{\rm det}}{N_{\rm rand}^{\rm cap}}
 \le\frac{1824d}{E+\sqrt{ET}}.
\]
For $T\ell\le E\le d/T$, the ratio lies between $d/(15456E)$ and $120d/E$.
\end{corollary}
\begin{proof}
As $ET\le d$ implies $T\le d$, the deterministic upper bound is at most
$d\min\{10T,100\sqrt2\,\sqrt{ET}\}$, and
\[
 \min\{10T,100\sqrt2\,\sqrt{ET}\}(E+\sqrt{ET})
 \le10TE+100\sqrt2\,\sqrt{ET}\sqrt{ET}
 =(10+100\sqrt2)ET<152ET.
\]
For the lower bound use $ab/(a+b)\le\min(a,b)$ with $a=T$, $b=\sqrt{ET}$ and
$\min\{T,\sqrt{ET/\ell}\}\ge\min\{T,\sqrt{ET}\}/\sqrt\ell$.
Divide deterministic lower and upper bounds by $23ET$ and $ET/12$, respectively, noting that $672\cdot23=15456$ and $12\cdot152=1824$. For the matched window use $10dT$, $dT/672$ and the random bounds directly.
\end{proof}
\subsection{Unrestricted parameter choice}
Assume $m\ge2$ and put
\[
 R^2=5m,\qquad r_0^2=4m,\qquad r_1^2=\frac92m.
\]
Let $q$ be the shifted Huber potential \eqref{eq:shifted-huber} with these
$r_0$, with $s=r_1-r_0$, and with $\lambda=1/(r_1-r_0)$; thus $q$ vanishes for
$\norm x\le r_0$, its radial derivative increases linearly from zero to one on
$[r_0,r_1]$, and it equals one afterwards. For an orthonormal frame
$U=(u_1,\ldots,u_m)$ define
\[
 \tilde F_{m,U}(x)=g_m(U^\top\rho_R(x))+q(x).
\]

\begin{lemma}[Smoothness, gap, and nonconvexity]
\label{lem:xu-smooth}
For every $m\ge2$ and every orthonormal $U$, the function $\tilde F_{m,U}$ is
coercive and nonconvex, and
\[
 \Lip(\nabla\tilde F_{m,U})<44,\qquad
 \tilde F_{m,U}(0)-\inf\tilde F_{m,U}\le\frac94\Bigl(m-\frac12\Bigr).
\]
\end{lemma}

\begin{proof}
At $m\ge2$ one has $r_1-r_0=(\sqrt{9/2}-2)\sqrt m\ge3-2\sqrt2>1/6$, so
$\lambda<6$. Write $t=\norm x/R$ and $a=(1+t^2)^{-1/2}$. On every smooth piece
of $\tilde F_{m,U}$, the chain Hessian has operator norm at most
$42a^2+\norm{Up}\norm{D^2\rho_R(x)}\le42/(1+t^2)+(30/\sqrt5)ta^3$ by
Lemmas~\ref{lem:chain} and~\ref{lem:radial}. For $r<r_0$, $\nabla^2q(x)=0$ and
\[
 \norm{\nabla^2\tilde F_{m,U}(x)}_{\rm op}
 \le\frac{42+(30/\sqrt5)t}{1+t^2}
 \le42+\frac{225}{42\cdot5}=\frac{603}{14}<44.
\]
For $r>r_0$, $t^2\ge4/5$, $a^2\le5/9$, $\norm{D^2\rho_R}\le2/\sqrt{15m}$, and $\norm{\nabla^2q}_{\rm op}\le\lambda<6$, so
\[
 \norm{\nabla^2\tilde F_{m,U}(x)}_{\rm op}
 \le42\cdot\frac59+\frac{20}{\sqrt{15}}+6<44.
\]
Lemma~\ref{lem:finite-piece} integrates this piecewise Hessian bound to
$\Lip(\nabla\tilde F_{m,U})<44$. The bounded chain and the eventually linear
growth of $q$ give coercivity. Since $q\ge0$ and $q(0)=0$, Lemma~\ref{lem:chain}
gives the gap.

For nonconvexity, take $x$ on the negative $u_1$ axis with $y_1=-9/8$ and all
other compressed coordinates zero. This point lies inside $r_0$, where $q=0$,
and the gate $\psi(-y_1)$ is saturated near it. Along the axis the only varying
chain term is $-\varphi(y_1)$. There $\varphi''=4$ and $\varphi'=1/2$, while the
scalar compression $s\mapsto-s/\sqrt{1+s^2/R^2}$ has positive second derivative
for $s>0$. The second derivative of the composite along the axis is therefore
negative.
\end{proof}

\begin{lemma}[Gradient barrier on the whole space]
\label{lem:xu-barrier}
If some coordinate of $U^\top\rho_R(x)$ has magnitude less than one, then
$\norm{\nabla\tilde F_{m,U}(x)}>3/5$.
\end{lemma}

\begin{proof}
Write $r=\norm x$, $w=\norm{\rho_R(x)}=ar$, $y=U^\top\rho_R(x)$,
$p=\nabla g_m(y)$, and $g=\nabla\tilde F_{m,U}(x)$. Let $j$ be the first index
with $|y_j|<1$, so that $p_j\le-1$ by Lemma~\ref{lem:chain}.

\emph{Inner ball, $r\le r_0$.} At $x=0$ one has $g=U\nabla g_m(0)=-u_1$, of
norm one, so assume $x\ne0$. Here $g=D\rho_R(x)Up$. With $t=r/R$, since
$t^2\le4/5$ and $R^2\ge10$,
\[
 \norm{D\rho_R(x)^{-1}u_j}^2
 =a^{-2}\Bigl[1+(a^{-4}-1)\frac{y_j^2}{a^2r^2}\Bigr]
 \le(1+t^2)\Bigl[1+\frac{(2+t^2)(1+t^2)}{R^2}\Bigr]
 \le\frac{1692}{625}<\frac{25}{9}.
\]
Hence $1\le|p_j|=|\ip{D\rho_R(x)^{-1}u_j}g|<(5/3)\norm g$, so $\norm g>3/5$.

\emph{Transition annulus, $r_0\le r\le r_1$.} Here $w^2/m\in[20/9,45/19]$ and
$a^2=1-w^2/(5m)$. Let $I=\{i:|y_i|<5/4\}$ and
$W=\operatorname{span}\{u_i:i\in I\}$. By Lemma~\ref{lem:chain}, $Up\in W$ and
$j\in I$. Since $|y_j|<1$ and $|y_i|<5/4$ for the other indices in $I$,
$\sum_{i\in I\setminus\{j\}}y_i^2\le\frac{25}{16}(m-1)$. With
$D=\norm{P_{W^\perp}\rho_R(x)}$, since $m\ge2$ and $w^2/m\ge20/9>25/16$,
\[
 D^2+y_j^2=w^2-\sum_{i\in I\setminus\{j\}}y_i^2\ge w^2-\frac{25}{16}(m-1)
 =2\Bigl(\frac{w^2}m-\frac{25}{32}\Bigr)+(m-2)\Bigl(\frac{w^2}m-\frac{25}{16}\Bigr)
 \ge2\Bigl(\frac{w^2}m-\frac{25}{32}\Bigr).
\]
As $y_j^2<1$, projecting $g=aUp+cx$ onto $u_j$ and $P_{W^\perp}x$ gives
\begin{equation}\label{eq:xu-annulus-conc}
 \norm g^2\ge\frac{a^2p_j^2D^2}{D^2+y_j^2}
 \ge\Bigl(1-\frac{w^2}{5m}\Bigr)\Bigl(1-\frac{1/2}{w^2/m-25/32}\Bigr).
\end{equation}
The right-hand side of~\eqref{eq:xu-annulus-conc} is concave in $w^2/m$ on
$[20/9,45/19]$, with endpoint values $271/747>9/25$ and $1322/3667>9/25$, so
$\norm g>3/5$.

\emph{Exterior, $r\ge r_1$.} Now $\nabla q=n:=x/r$, $w^2\ge\frac{45}{19}m$,
$a^2=1-w^2/(5m)\le\frac{10}{19}$, and
$\norm{P_W\rho_R(x)}^2=\sum_{i\in I}y_i^2\le\frac{25}{16}m$. If $p\ne0$, put
$c=|\ip n{Up}|/\norm{Up}\le\norm{P_Wn}\le\sqrt{(25/16)m}/w$, so
$c^2\le\frac{25/16}{45/19}=\frac{95}{144}$. Since $D\rho_R(x)$ has radial
eigenvalue $a^3$ and tangential eigenvalue $a$, for $v=D\rho_R(x)Up$ we have
$|\ip nv|^2/\norm v^2=a^4c^2/(1-c^2+a^4c^2)\le\frac{500}{1431}$, and
projecting $g=v+n$ onto $v^\perp$ gives
$\norm g^2\ge1-\frac{500}{1431}=\frac{931}{1431}>\frac9{25}$. If $p=0$, then
$\norm g=1$.
\end{proof}

When $y_j=0$ and $|y_i|\le1/2$ for all $i>j$, let
\[
 \tilde r_j(x)=g_m\bigl(u_1^\top\rho_R(x),\ldots,u_{j-1}^\top\rho_R(x),
 0,\ldots,0\bigr)+q(x).
\]
Every later gate and its derivative vanish, $p_j=-\gamma_{j-1}$, and
$u_j^\top x=0$ gives $D\rho_R(x)u_j=a(x)u_j$. Hence
\begin{equation}
 \tilde F_{m,U}(x)=\tilde r_j(x),\qquad
 \nabla\tilde F_{m,U}(x)=\nabla\tilde r_j(x)-a(x)\gamma_{j-1}(x)u_j.
 \label{eq:xu-prefix}
\end{equation}

\begin{theorem}[Lower bound on the unrestricted smooth class]\label{thm:linear-lower}\label{a:thm:linear-lower}
For all $d,T\ge1$ on the unrestricted globally $L$-smooth class,
$N_{\rm det}^{\rm lin}\ge\frac d{500}\min\{T,d/\log(ed)\}$;
the same bound holds for exact values.
\end{theorem}
\subsubsection{Proof of Theorem~\ref{thm:linear-lower}}

Put $b=3/5$, $A=44\cdot(9/4)/b^2=275$, and $S=\min\{T,d/\log(ed)\}$.

\emph{Case $S>500$.} Let $m=\lfloor S/A+1/2\rfloor$ and $h=\lfloor7d/10\rfloor$.
Since $500>3A/2=825/2$, we have $m\ge2$, and $m-1/2\le S/A\le d/(A\log(ed))$ gives
$d/m\ge(A/2)(2-1/m)\log(ed)$ and $m<d/1000$. Hence $p_0=d-m-h\ge3d/10-m>0$ and, with $\tau=1/2$ and $R^2=5m$,
\[
 \frac{p_0\tau^2}{2R^2}=\frac{p_0}{40m}
 \ge\frac{3d}{400m}-\frac1{40}
 \ge\frac{33}{32}\Bigl(2-\frac1m\Bigr)\log(ed)-\frac1{40}
 >\Bigl(2-\frac1m\Bigr)\log(ed).
\]
Since $d>500$ and $m<d/1000$, we have $2m(ed)^{1/m}\le\max\{4\sqrt{ed},2em\}<e^2d$, so $2hm<2dm<(ed)^{2-1/m}$ and
\[
 2hm\exp\Bigl(-\frac{p_0\tau^2}{2R^2}\Bigr)
 <(ed)^{2-1/m}\exp\Bigl(-\Bigl(2-\frac1m\Bigr)\log(ed)\Bigr)=1,
\]
which is the cap condition of Lemma~\ref{lem:stage}, whose residual dimension
$p_0+1$ only improves it. Apply Lemma~\ref{a:lem:scaling} with $L'=L$, $\Lambda=44$, $K=44I$, and $D=\frac94(m-\frac12)$; the trace condition
holds because $E=d$, and the accuracy condition is
$T\ge44D/b^2=A(m-\frac12)$, which follows from $A(m-\frac12)\le S\le T$.
Lemmas~\ref{lem:xu-smooth} and~\ref{lem:xu-barrier} supply the smoothness,
the gap, and a barrier larger than $\eps$ after scaling, and
\eqref{eq:xu-prefix} verifies the prefix hypotheses of Lemma~\ref{lem:stage}
with $z=\rho_R$ and $c_j=a\gamma_{j-1}$. Every deterministic algorithm with
rank-priced cost below $hm$ therefore fails on some fixed scaled
$\tilde F_{m,U}$. Finally, $S<A(m+1/2)$ and $m\ge2$ give $m>4S/(5A)$; since
$d>500$,
\[
 hm>\frac{dS}{A}\Bigl(\frac{14}{25}-\frac4{5d}\Bigr)>\frac{dS}{500},
\]
using $14/25-A/500=1/100>4/(5d)$.

The witnesses in the chain regime are coercive and nonconvex by Lemma~\ref{lem:xu-smooth}. They
also have gradient norm at most $2\sqrt{L\Delta}$: $\norm{\nabla\tilde
F_{m,U}}\le10\sqrt m+1\le11\sqrt m$, so the scaled gradient is at most
$(55/3)\sqrt m\,\eps$, while $L\Delta\ge A(m-1/2)\eps^2\ge(A/2)m\eps^2$ and
$(55/3)^2<2A$.

\emph{Case $S\le500$.} Lemma~\ref{lem:dimension-floor} with $E=d$ gives
$N^{\rm lin}_{\rm det}\ge d\ge dS/500$. Algorithms that use only values are a special case of rank-priced ones, so
$N^{\rm val}_{\rm det}\ge N^{\rm lin}_{\rm det}$. \qed

\section{Quadratic Randomized Lower Bound}
\label{a:sec:random}
Algorithms here are measurable functions of their independent seed and finite transcript; success is at least $2/3$ on each fixed function in the class, not merely on average over the distribution used in the lower bound. A subscript $E$, as in $N_{{\rm rand},E}$, indicates the class $\cF^d_{L,\Delta,E}$.

\begin{lemma}[Explicit spherical coordinate splitting]
\label{a:lem:sphere-split}
For $q\ge2$, a uniform $U\in S^{q-1}$ admits the product representation
\[
 U=(C,\sqrt{1-C^2}\,W),\qquad
 W\sim\operatorname{Unif}(S^{q-2}),\quad W\text{ independent of }C,
 \quad\mathbb EC^2=1/q.
\]
It is valid after any specified orthogonal change of coordinates. For $q=1$ the sphere is a fair sign.
\end{lemma}
\begin{proof}
Let $(G_1,G')$ be standard Gaussian in $\mathbb R\times\mathbb R^{q-1}$. Its normalized direction is uniform by rotational invariance of Gaussian density. Polar coordinates for $G'$ show that $W=G'/\|G'\|$ is uniform and independent of $(G_1,\|G'\|)$; for $q=2$ this is independence of the sign and magnitude of a scalar Gaussian. Set
$C=G_1/\sqrt{G_1^2+\|G'\|^2}$. Coordinate exchangeability and $\sum_iU_i^2=1$ give $\mathbb EC^2=1/q$.
For a specified unit direction $v$, a measurable alignment is the Householder matrix $I-2(e_1-v)(e_1-v)^\top/\|e_1-v\|^2$ if $v\ne e_1$ and $I$ otherwise; it preserves the uniform law.
\end{proof}

\begin{lemma}[Adaptive projections: energy and antipodes]
\label{a:lem:adaptive-projections}\label{lem:projections}
Let $u$ be uniform on the unit sphere in an $e$-dimensional Euclidean space and independent of an algorithm's seed. After at most $w$ adaptively chosen exact linear observations, where $0\le w\le e-1$, any measurable estimate $z$ satisfies
\begin{equation}\label{a:eq:adaptive-projections}
 \mathbb P(\|z-u\|\le r)\le\frac12+\frac{w}{2e(1-r^2)}
 \quad (0\le r<1).
\end{equation}
\end{lemma}
\begin{proof}
Fix the seed. Orthogonalize independent queried directions, omitting dependent directions whose replies are already determined. After termination pad the span using Gram--Schmidt on a fixed basis until its dimension is exactly $w$; the extra observations can be revealed to the analysis even if the estimator ignores them. The successive unit directions $v_j$ are measurable functions of previous observations, and each is orthogonal to those preceding it.

We construct the joint law explicitly on a product space. Begin with residual squared norm $1$ and an orthonormal basis. At step $j$, whose residual dimension is $q=e-j$, align the next chosen direction to the first residual axis by Lemma~\ref{a:lem:sphere-split}. Draw an independent scalar $C_j$ with the law of the first coordinate in dimension $q$. Reveal
$\alpha_j=\sqrt{1-Z_j}\,C_j$, define
$Z_{j+1}=Z_j+(1-Z_j)C_j^2$, and use the remaining aligned axes as a basis of the new residual subspace. Finally draw an independent uniform $W$ in the residual dimension $e-w$ and form
\begin{equation}\label{a:eq:product-residual}
 u=\sum_{j=0}^{w-1}\alpha_jv_j+\sqrt{1-Z_w}\,B_wW,
\end{equation}
where $B_w$ is the residual isometry determined by the revealed history.
This produces the same joint integrals as a uniform $u$ together with the observations selected by the adaptive rule: at each step apply the product representation of Lemma~\ref{a:lem:sphere-split} to the innermost integral over the residual sphere, hold the outer history fixed, and then use Fubini. Induction over the finite $w$ steps establishes this for every bounded measurable test function of $(u,\text{history})$. When the residual radius is zero, any residual basis can be used without changing the vector. No regular conditional probability is invoked.

Because $C_j$ is independent of the preceding history and has second moment $1/(e-j)$,
\[
 \mathbb E Z_{j+1}=\mathbb E Z_j+
 \frac{1-\mathbb E Z_j}{e-j},\qquad \mathbb E Z_w=w/e.
\]
Fix a history on the product space and hence the estimator $z$. If $1-Z_w>r^2$, the two vectors in~\eqref{a:eq:product-residual} generated by $W$ and $-W$ have separation $2\sqrt{1-Z_w}>2r$, so at most one lies within distance $r$ of $z$. Paired integration over the final symmetric sphere bounds success by $1/2$ on this event, and by one otherwise. Markov's inequality gives
$\mathbb P(Z_w\ge1-r^2)\le w/[e(1-r^2)]$. This proves~\eqref{a:eq:adaptive-projections}; average over the independent seed.
\end{proof}

\begin{proposition}[Quadratic randomized lower bound]
\label{a:prop:quadratic-baseline}\label{prop:quadratic-baseline}
Let $W\subseteq\mathbb R^d$ be a public subspace of dimension $e\ge1$, let $\ell,\Delta,\varepsilon>0$, and suppose $\theta=\ell\Delta/\varepsilon^2>1/2$. Define
\[
 a_\theta=1-\frac{1}{2\theta},\qquad
 R_0=\sqrt{\frac{2\Delta}{\ell}},\qquad
 f_u(x)=\frac\ell2\|P_Wx-R_0u\|^2,
 \quad u\sim\operatorname{Unif}(S(W)).
\]
Every $f_u$ is convex and $\ell$-smooth, has initial gap exactly $\Delta$, and admits the global fixed upper matrix $\ell P_W$. Any randomized algorithm successful with probability at least $2/3$ on every $f_u$ has rank-priced cost cap at least $ea_\theta/3$ and worst-case expected rank-priced cost at least $ea_\theta/36$. These also hold for exact values.
\end{proposition}
\begin{proof}
We have
\[
 f_u(x)=\frac\ell2\|P_Wx\|^2+\Delta-\ell R_0\langle u,P_Wx\rangle,
 \qquad\nabla f_u(x)=\ell(P_Wx-R_0u).
\]
A response to call $(x,V)$ is determined by the projections of $u$ on $\operatorname{span}\{P_Wx\}+\operatorname{col}(P_WV)$, whose dimension is at most $1+\operatorname{rank}V$ and is chosen before the reply. Revealing a basis of those projections simulates the response within its cost. Success means
$\|P_W\widehat x/R_0-u\|\le(2\theta)^{-1/2}$.
Costs are integers. If a cost cap $w$ is below $e$, Lemma~\ref{a:lem:adaptive-projections} bounds average success over $u$ by $1/2+w/(2ea_\theta)$, so success probability $2/3$ forces $w\ge ea_\theta/3$. Caps $w\ge e$ satisfy this since $a_\theta<1$.

For expected cost let $\mathsf W$ be the random cost. For every integer $0\le k<e$, truncate before any call that would raise cost above $k$, returning a fixed point. It agrees with the original whenever $\mathsf W\le k$; therefore, with probability averaged over $u$ and the seed,
$\mathbb P(\mathsf W>k)\ge(1/6-k/(2ea_\theta))_+$.
Since $ea_\theta/3<e$, tail summation gives
\[
 \mathbb E\mathsf W\ge\sum_{k\ge0}\left(\frac16-\frac{k}{2ea_\theta}\right)_+
 \ge\int_0^{ea_\theta/3}\left(\frac16-\frac{t}{2ea_\theta}\right)dt
 =\frac{ea_\theta}{36},
\]
as the integrand is decreasing and nonnegative. The supremum over $u$ of conditional expected cost is at least this average. Exact value calls have rank zero.
\end{proof}

\begin{corollary}[Randomized baseline on the trace class]
\label{cor:quadratic-baseline}\label{a:cor:quadratic-baseline}
For all $d\ge1$, $1\le E\le d$, and $T\ge1$,
\[
 N_{\rm rand,E}^{\rm val}\ge N_{\rm rand,E}^{\rm lin}\ge
 \frac{\min\{d,ET\}}{12},
\]
and both worst-case expected costs are at least $\min\{d,ET\}/144$.
\end{corollary}
\begin{proof}
Take $e=\min\{d,\lfloor ET\rfloor\}$, a coordinate subspace $W$ of dimension $e$, and $\ell=\min\{L,LE/e\}$. Since $ET\ge1$, $e\ge\min\{d,ET\}/2$. Further, $\ell\le L$, $\ell e\le LE$, and $\theta=\min\{T,ET/e\}\ge1$. Thus these quadratics lie in $\mathcal F^d_{L,\Delta,E}$, $a_\theta\ge1/2$, and the proposition yields cap at least $e/6$ and expected cost at least $e/72$.
\end{proof}

\section{Further Results}
\label{app:further}
Theorem~\ref{a:thm:main} on $\mathcal B$ does not use this appendix; its
extension to $\mathcal F$ uses Section~\ref{gl:sec:global}. The appendix defines the pointwise classes, records the
inclusions between the classes, gives the randomized
upper bound on the larger class $\cA^d_{L,\Delta,E}$, and $C^\infty$ hard
functions with a Lipschitz Hessian in the global fixed-matrix class
$\cF^d_{L,\Delta,E}$, which give the deterministic lower bounds on every class
that contains it.

\subsection{Pointwise classes and inclusions}
\label{app:classes}
Our randomized upper bound holds on a larger class, the local pointwise class
$\cA^d_{L,\Delta,E}$, that requires neither a common matrix $H$ nor smoothness
above $f(0)$: it consists of the continuously differentiable $f$ with
$f(0)-\inf f\le\Delta$ whose gradient is locally $L$-Lipschitz on $U$ and which
satisfy $\operatorname{tr}[(\nabla^2f(x))_+]\le LE$ for almost every $x\in U$,
where $A_+$ keeps the positive eigenvalues of a symmetric matrix $A$ ($U$ need
not be bounded or connected). Since $A\preceq H$ and $H\succeq0$ imply
$\operatorname{tr}A_+\le\operatorname{tr}H$, we have
$\cB^d_{L,\Delta,E}\subseteq\cA^d_{L,\Delta,E}$.
The global pointwise class
$\cP^d_{L,\Delta,E}$ requires $f\in\cF^d_{L,\Delta}$ and
$\operatorname{tr}[(\nabla^2f(x))_+]\le LE$ for almost every $x\in\R^d$.

\begin{lemma}[Class inclusions]\label{p:lem:inclusions}
At each fixed choice of the public parameters,
\[
 \cF^d_{L,\Delta,E}\subseteq\cP^d_{L,\Delta,E}\subseteq\cA^d_{L,\Delta,E},
 \qquad
 \cF^d_{L,\Delta,E}\subseteq\cB^d_{L,\Delta,E}\subseteq\cA^d_{L,\Delta,E}.
\]
\end{lemma}
\begin{proof}
Let $f\in\cF^d_{L,\Delta,E}$ have upper model $H$. At a point where
$\nabla f$ is differentiable, the quadratic inequality \eqref{eq:trace-class},
applied to $tv$ and letting $t\to0$, gives $v^\top\nabla^2f(x)v\le v^\top Hv$
for every $v$. By Rademacher's theorem this holds almost everywhere on $\R^d$,
so $f\in\cB^d_{L,\Delta,E}$. For symmetric $A\preceq H$ with $H\succeq0$, let
$P_+$ be the spectral projection onto the positive eigenspaces of $A$. Then
\[
 \operatorname{tr}A_+=\operatorname{tr}(P_+A)
 \le\operatorname{tr}(P_+H)\le\operatorname{tr}H\le LE .
\]
Applied almost everywhere on $\R^d$, this gives $f\in\cP^d_{L,\Delta,E}$.
Applied almost everywhere on the initial sublevel set, it gives the trace bound
of $\cA^d_{L,\Delta,E}$ for every $f\in\cB^d_{L,\Delta,E}$. Finally, a globally
$L$-Lipschitz gradient is locally $L$-Lipschitz on the initial sublevel set,
and a trace bound on $\R^d$ holds in particular there; this gives both
inclusions in $\cA^d_{L,\Delta,E}$.
\end{proof}
Consequently lower bounds proved on $\cF^d_{L,\Delta,E}$ and upper bounds
proved on $\cA^d_{L,\Delta,E}$ hold on all four classes.

\subsection{The randomized upper bound on the local pointwise class}
\label{app:local-class}
Recall that $f\in\cA^d_{L,\Delta,E}$ means $f\in C^1(\R^d)$,
$f(0)-\inf f\le\Delta$, and, on $U=\{f<f(0)\}$, a gradient that is locally
$L$-Lipschitz with one constant throughout $U$ and
$\operatorname{tr}((\nabla^2f)_+)\le LE$ almost everywhere. No common matrix is
assumed, so a capped function replaces the first exit
Lemma~\ref{u:lem:cap}.

\begin{lemma}[Capped function]
\label{x:lem:cap}
Let $f\in\mathcal A^d_{L,\Delta,E}$, $c=f(0)$,
$G=\sqrt{2L\Delta}$, and $F_c=\min\{f,c\}$. Then $F_c$ is globally
$G$-Lipschitz and bounded. For $f(x)\le c$,
\begin{equation}
 F_c(x+v)\le f(x)+\langle\nabla f(x),v\rangle+\frac L2\|v\|^2.
 \label{x:eq:cap-model}
\end{equation}
For any fixed orthogonal projection $P$, $Z\sim N(0,P)$, $t\ge0$, and
$y\in\mathbb R^d$,
\begin{equation}
 \mathbb E F_c(y+\sqrt t Z)\le F_c(y)+\frac{LE}{2}t.
 \label{x:eq:heat}
\end{equation}
\end{lemma}
\begin{proof}
For $x\in U$, put $g=\nabla f(x)$. Follow $x-tg$ up to $t=1/L$.
Before its first possible exit from $U$, local smoothness and integration
along the segment give
$f(x-tg)\le f(x)-t(1-Lt/2)\|g\|^2$. A first exit before $1/L$ would
contradict this strict decrease and continuity. Consequently
$\|g\|^2\le2L(f(x)-\inf f)\le G^2$.
The restriction of $F_c$ to any compact line segment is absolutely
continuous: $f$ is $C^1$ there and truncation is Lipschitz. Its derivative is
bounded in magnitude by $G$ almost everywhere, being either a derivative
of $f$ below $c$ or zero. This proves the global Lipschitz bound.
Boundedness follows from $\inf f\le F_c\le c$.

We argue by smoothing. For $\delta>0$, choose
a smooth concave nondecreasing function $\phi_\delta$ with
$0\le\phi_\delta'\le1$, derivative one below $c-2\delta$, derivative zero
above $c-\delta$, and
$\sup_s|\phi_\delta(s)-\min(s,c)|\le2\delta$.
It is obtained by integrating a smooth decreasing transition on
$[c-2\delta,c-\delta]$ and fixing the additive constant. The function
$u_\delta=\phi_\delta\circ f$ is globally $C^{1,1}$: where its derivatives
are nonzero one is in $U$, its Hessian has norm bounded by
$L+\|\phi_\delta''\|_\infty G^2$, and near the remaining points the
composition is constant. Local bounds with this common constant integrate
to a global bound. Almost everywhere,
\[
 \nabla^2u_\delta
 =\phi_\delta'(f)\nabla^2f+
   \phi_\delta''(f)\nabla f\nabla f^\top\preceq LI,
 \qquad \operatorname{tr}(P\nabla^2u_\delta)\le LE.
\]
The second inequality uses
$\operatorname{tr}(P B)\le\operatorname{tr}(B_+)$ for every symmetric $B$
and $0\preceq P\preceq I$; the cutoff's second term is nonpositive.
Convolve in \emph{all $d$ dimensions} with a nonnegative, symmetric,
compactly supported smooth approximate identity. Both inequalities survive.
For the resulting smooth bounded function $u$, the quadratic upper model
follows by integration. Gaussian integration by parts gives
\[
 \frac{d}{dt}\mathbb E u(y+\sqrt t Z)
 =\frac12\mathbb E\operatorname{tr}(P\nabla^2u(y+\sqrt t Z))\le LE/2.
\]
This is also valid for degenerate $P$ by using coordinates on its range;
it is applied only \emph{after} full-dimensional smoothing. Bounded
Hessians and Gaussian moments justify differentiation and integration by
parts. Integrate from $0$ to $t$, send the convolution radius to zero, and
then send $\delta$ to zero. The convergence of the clipped values is uniform,
which proves \eqref{x:eq:heat}.

For $x\in U$, the same argument gives \eqref{x:eq:cap-model}, because
$u_\delta(x)\to f(x)$ and $\nabla u_\delta(x)\to\nabla f(x)$.
If $f(x)=c$ and $\nabla f(x)\ne0$, approach $x$ from the local downhill
side in $U$ and use continuity. If $\nabla f(x)=0$, use $F_c\le c$.
\end{proof}

In Lemma~\ref{x:lem:cap} the concave scalar
cutoff handles the negative curvature created by truncation, and the Gaussian
identity is applied only after smoothing in all $d$ dimensions, because an almost everywhere bound
in $\mathbb R^d$ may fail on a hyperplane.

The update below is the stochastic three-point method of
Theorem~\ref{b:thm:stp} (Algorithm~\ref{alg:stp}), with the curvature parameter $L(E+1)$ in place of
$LE$.

\begin{theorem}[Three-point comparison on $\mathcal A$]
\label{x:thm:stp}
Let $M=L(E+1)$, $h=4\varepsilon/(5M)$, and
$N=\lceil10M\Delta/\varepsilon^2\rceil$. Query $f(0)$ once and cache the
current value. At each iteration draw $a\sim N(0,I_d)$, query
$f(x+ha)$ and $f(x-ha)$, and retain a point of lowest value among
$\{x,x+ha,x-ha\}$, keeping $x$ on a tie. Return an independent uniformly selected iterate from before the updates. For every fixed $f\in\mathcal A$, the
probability of failure is at most $25/78<1/3$, and the hard query cap is
\[
 Q=1+2\lceil10(E+1)T\rceil\le43ET\qquad(E,T\ge1).
\]
\end{theorem}
\begin{proof}
All centers satisfy $f(x)\le c=f(0)$. Fix such a center and write
$g=\nabla f(x)\ne0$, $u=g/\|g\|$, and $a=su+z$, where $s\sim N(0,1)$
and $z\sim N(0,I-uu^\top)$ are independent. Of the two proposals, one is
\[
 x-h|s|u-h\operatorname{sign}(s)z.
\]
Although the algorithm does not know this sign, it compares both proposals.
Conditioning on $s$, apply \eqref{x:eq:heat} in the orthogonal subspace,
then apply \eqref{x:eq:cap-model} at $x$ along $-h|s|u$.
Since the minimum of the three true values is at most the capped value of
this particular proposal, this gives
\begin{equation}
 \mathbb E[f(x^+)\mid x]
 \le f(x)-\alpha_0h\|g\|+\frac M2h^2,
 \qquad \alpha_0=\mathbb E|s|=\sqrt{2/\pi}>79/100.
 \label{x:eq:comparison-descent}
\end{equation}
Only capped functions and the accepted value are integrated, so the value at
a rejected proposal need not have a finite Gaussian expectation.

The accepted value never increases. Thus when $\|g\|>\varepsilon$ the
conditional expected decrease is at least
\[
 \left(\frac{79}{100}\frac45-\frac12\frac{16}{25}\right)
 \frac{\varepsilon^2}{M}
 =\frac{39}{125}\frac{\varepsilon^2}{M},
\]
and at every other center it is nonnegative. Summation and the initial
gap yield
\[
 \frac1N\sum_{t=0}^{N-1}\mathbb P(\|\nabla f(x_t)\|>\varepsilon)
 \le\frac{125M\Delta}{39N\varepsilon^2}\le\frac{25}{78}.
\]
This is exactly the failure probability of the independent uniform output.
Every iteration uses two new values, including unsuccessful proposals.
The final bound follows from $1+2\lceil10(E+1)T\rceil
\le3+20(E+1)T\le43ET$.
\end{proof}

\subsection{Smooth lower bounds on the global classes}
\label{gl:sec:global}

This subsection proves the deterministic lower bounds of
Theorem~\ref{thm:main-det} and Corollary~\ref{cor:main-window} with $C^\infty$
witnesses that have a Lipschitz Hessian and lie in the global fixed-matrix
class $\cF^d_{L,\Delta,E}$, and hence in $\cP^d_{L,\Delta,E}$ and
$\cB^d_{L,\Delta,E}$. It changes the witnesses, not the definition of the
optimization oracle. Smoothing an entire hard function need not keep the hidden
directions invisible to exact values, so we smooth the one-dimensional building
blocks of the chain of Appendix~\ref{app:chain} and retain an open interval on
which each gate is \emph{exactly} zero. The smoothed chain is multiplied by a
perspective weight instead of being added to a radial potential, so the Hessian
bound holds on all of $\R^d$.

The constants of this subsection are derived with the following selection
lemma. It uses the symmetric Gaussian rectangle inequality
\citep{sidak1967rectangular} instead of Lemma~\ref{a:lem:selection}, and its
constant $12/5$ is smaller than the constant $13/3$ there.

\begin{lemma}[Subspace strip avoidance]
\label{lem:strip-existence}
Let $S\subseteq\R^d$ have dimension $p\ge1$, and let $a_1,\ldots,a_M\in\R^d$
satisfy $\norm{a_i}\le R$. There is a unit $u\in S$ such that
\begin{equation}
 \max_{i\le M}|\ip u{a_i}|\le\frac{12}{5}R\sqrt{\frac{\log(e+M/p)}{p}}.
 \label{x:eq:strip}
\end{equation}
\end{lemma}

\begin{proof}
If $M<p$ or $R=0$, choose a unit vector $u\in S$ orthogonal to the projections
of $a_1,\ldots,a_M$ onto $S$. Assume now that $s:=M/p\ge1$ and $R>0$. Draw a
standard Gaussian $g$ in $S$ and put $t=\frac95\sqrt{\log(e+s)}>2$. Each
$X_i=\ip g{a_i}$ is centered Gaussian with variance at most $R^2$, so
\[
 \Prob\{|X_i|>Rt\}\le\sqrt{\frac2\pi}\frac{e^{-t^2/2}}t
 <\frac4{5t}(e+s)^{-81/50}<\frac1{16s}.
\]
The symmetric Gaussian rectangle inequality gives
\[
 \Prob\Bigl\{\max_i|X_i|\le Rt\Bigr\}
 \ge\prod_{i=1}^M\Prob\{|X_i|\le Rt\}
 \ge\Bigl(1-\frac1{16s}\Bigr)^M\ge e^{-p/15},
\]
using $\log(1-z)\ge-z/(1-z)$ and $M/s=p$. Since $\norm g^2$ is
chi-squared with $p$ degrees of freedom, exponential Markov with parameter
$7/18$ gives
$\Prob\{\norm g^2\le\frac9{16}p\}\le(\frac34e^{7/32})^p<e^{-p/15}$. Hence some $g$
satisfies both $\max_i|X_i|\le Rt$ and $\norm g>\frac34\sqrt p$, and $u=g/\norm g$
satisfies $\max_i|\ip u{a_i}|\le\frac43Rt/\sqrt p=\frac{12}{5}R\sqrt{\log(e+s)/p}$.
\end{proof}

\subsubsection{One-dimensional smoothing}
Let $\varphi,\psi$ be the scalar functions of Appendix~\ref{app:chain}.
An equivalent definition, useful for checking their bounds, is
\[
 \varphi(t)=\int_{-\infty}^t
   \bigl(\min\{1,4(s+5/4),4(5/4-s)\}\bigr)_+\,ds,
 \quad
 \psi(t)=\int_{-\infty}^t16\min\{(s-1/2)_+,(1-s)_+\}\,ds.
\]
Thus $0\le\varphi\le9/4$, $\varphi'(t)=1$ on $[-1,1]$,
$|\varphi''|\le4$ almost everywhere, $0\le\psi'\le4$, and
$|\psi''|\le16$ almost everywhere. Put $\eta=1/24$ and
\[
 w(t)=Z^{-1}\exp[-1/(1-t^2)]\mathbf1_{\{|t|<1\}},
 \quad Z=\int_{-1}^1\exp[-1/(1-t^2)]\,dt,
 \quad w_\eta(t)=\eta^{-1}w(t/\eta).
\]
Define
\begin{equation}
 \widetilde\varphi=\varphi*w_\eta,
 \qquad \widetilde\psi(t)=(\psi*w_\eta)(t+2\eta).
 \label{x:eq:gates}
\end{equation}
The shift by $2\eta$ moves the transition of $\widetilde\psi$ into
$[3/8,23/24]$, so $\widetilde\psi=1$ on $[1,\infty)$, as is $\psi$.

\begin{lemma}[Smooth gates and a dimension-free chain estimate]
\label{x:lem:smooth-chain}
The gates in \eqref{x:eq:gates} are $C^\infty$. Write
$a_0=3/8$ and $a_1=23/24$. Then $\widetilde\psi=0$ on
$(-\infty,a_0]$, $\widetilde\psi=1$ on $[a_1,\infty)$,
$0\le\widetilde\psi'\le4$, $|\widetilde\psi''|\le16$, and
$|\widetilde\psi'''|<788$.
The function $\widetilde\varphi$ takes values in $[0,9/4]$,
$\widetilde\varphi(0)=9/8$, its derivative is even, is one on
$[-a_1,a_1]$, and satisfies
$0\le\widetilde\varphi'\le1$, $|\widetilde\varphi''|\le4$,
$|\widetilde\varphi'''|<100$.

For the chain \eqref{eq:chain} with these gates, denoted $\widetilde g_m$,
put $p=\nabla\widetilde g_m(y)$ and $B=\nabla^2\widetilde g_m(y)$. Then
\begin{align}
 &p_i\le0,\quad \|p\|\le10\sqrt m,\quad \|B\|_{\rm op}\le42,
 \quad \|D^3\widetilde g_m\|_{\rm op}\le1872,\label{x:eq:chain-norms}\\
 &\|By\|\le44\sqrt m,\quad |y^\top By|\le44m,
 \quad |y^\top p|\le8m,\label{x:eq:chain-weighted}\\
 &\widetilde g_m(0)-\inf\widetilde g_m
       \le\tfrac94(m-\tfrac12).\label{x:eq:chain-gap}
\end{align}
If some $|y_i|<a_1$, the first such index $j$ satisfies $p_j\le-1$.
If $y_j=0$ and $|y_i|\le a_0$ for $i>j$, the exact prefix identities
of \eqref{eq:chain-gradient} hold; here $\widetilde\varphi'(0)=1$, and the
current derivative is $-\gamma_{j-1}$, where $\gamma_0=1$ and
$\gamma_{j-1}=\widetilde\psi(-y_{j-1})+\widetilde\psi(y_{j-1})\in[0,1]$
for $j>1$. Values at such a point may still depend on several earlier
coordinates.
\end{lemma}
\begin{proof}
Convolution with a nonnegative probability kernel preserves all the stated
bounds on the values and the first and second derivatives. The support and plateau
claims follow by enlarging endpoints by $\eta$ and applying the shift
$2\eta$. Evenness gives $\widetilde\varphi(0)=9/8$ and an even derivative.
On $[-3/5,3/5]$, $1/(1-t^2)\le1+(25/16)t^2$ gives
$\exp[-1/(1-t^2)]\ge e^{-1}(1-(25/16)t^2)$, so integrating over $[-3/5,3/5]$
yields $Z>(39/40)e^{-1}$ and $w(0)=e^{-1}/Z<40/39$. Because $\varphi''$ is a
step function with jumps of magnitude $4$ at $\{-5/4,-1,1,5/4\}$ and $\psi''$
has jumps of magnitude at most $32$ at $\{1/2,3/4,1\}$, and the jumps in each
set are separated by $1/4>2\eta$, the bumps in $\widetilde\varphi'''=\varphi'''*w_\eta$
and $\widetilde\psi'''(t)=(\psi'''*w_\eta)(t+2\eta)$ have disjoint supports.
Thus $|\widetilde\varphi'''|\le4w(0)/\eta<100$ and
$|\widetilde\psi'''|\le32w(0)/\eta<788$.

The coordinate gradient is the sum of a nonpositive predecessor term of
magnitude at most $1$ and a nonpositive successor term of magnitude at most
$4(9/4)=9$. Since $\widetilde\varphi''$ and $\widetilde\varphi'''$ are
supported on $\{|t|\in[a_1,5/4+\eta]\}$ whereas $\widetilde\psi'$,
$\widetilde\psi''$, and $\widetilde\psi'''$ are supported on
$\{|t|\in[a_0,a_1]\}$, their supports meet only at $|t|=a_1$, where both
higher derivatives vanish. Convolving \eqref{eq:row-sum-identity} with
$w_\eta\ge0$ gives $4+16\widetilde\varphi(s)+4\widetilde\varphi'(s)\le42$ for
every $s\in\R$, so the Hessian row sum is at most $42$ for $|y_i|\le a_1$ and
at most $8$ for $|y_i|>a_1$, proving $\|B\|_{\rm op}\le42$. Similarly, a diagonal third
derivative has magnitude at most $\max\{100,(9/4)788\}=1773$, and adding the
six types of adjacent mixed trilinear terms (each bounded by $16$) gives
$1773+6\cdot16<1872$.

Every active coordinate has $|y_i|\le5/4+\eta=31/24<4/3$. For $|y_i|\le a_1$,
$|(By)_i|\le36a_1+4a_1+4(31/24)=87/2<44$ and $|y_i(By)_i|<44$; for
$a_1<|y_i|\le31/24$, $|(By)_i|\le4(31/24)+4a_1<10$. Also $s\psi'(s)\le3$ and
$|s\varphi'(s)|\le1$ imply $t\widetilde\psi'(t)\le3$ and
$|t\widetilde\varphi'(t)|\le1+\eta$, so $|y_ip_i|\le1+\eta+27/4<8$.
The barrier and prefix identities follow as in Lemma~\ref{lem:chain}.
\end{proof}

\subsubsection{Perspective scaling and exact derivatives}
\label{gl:subsec:identities}

For $R>0$, define the perspective weight, compression factor, and bounded
radial compression map on $\R^d$ by
\begin{equation}
 w(x)=\sqrt{1+\frac{\norm x^2}{R^2}},\qquad
 a(x)=\frac1{w(x)},\qquad
 z(x)=\rho_R(x)=\frac{x}{w(x)}=a(x)x.
 \label{gl:eq:perspective-defs}
\end{equation}
Then $w\in C^\infty(\R^d)$, $w(x)\ge1$, $0<a(x)\le1$, and $\norm{z(x)}<R$ for
all $x\in\R^d$. Multiplying a scalar chain by $w(x)$ while evaluating it at the
compressed coordinates $U^\top z(x)=U^\top x/w(x)$ forms the perspective function \citep[Section~3.2.6]{boyd2004convex} of $G$ evaluated at $(U^\top x,w(x))$, namely $w(x)G(U^\top x/w(x))$. Because $w(x)a(x)=1$ identically, the weight
$w(x)$ cancels the radial attenuation $a(x)$ in the leading chain gradient
while keeping the second derivative bounded across all of $\R^d$.

\begin{lemma}[Exact derivatives under perspective scaling]
\label{gl:lem:perspective-derivatives}
Let $U=(u_1,\ldots,u_m)\in\R^{d\times m}$ have orthonormal columns, let
$G:\R^m\to\R$ be continuously differentiable, and set
\[
 \Phi_U(x)=w(x)\,G\bigl(U^\top\rho_R(x)\bigr),\qquad
 y=U^\top z(x).
\]
At every point $x\in\R^d$, with $p=\nabla G(y)$,
\begin{equation}
 \nabla\Phi_U(x)=Up+\bigl[G(y)-y^\top p\bigr]\frac{z(x)}{R^2}.
 \label{gl:eq:perspective-grad}
\end{equation}
At every point $x\in\R^d$ where $G$ is twice differentiable at $y$ with Hessian
$B=\nabla^2G(y)$, for every $v\in\R^d$, writing $\alpha=U^\top v$ and
$\beta=z(x)^\top v/R^2$,
\begin{equation}
 v^\top\nabla^2\Phi_U(x)v
 =a(x)(\alpha-\beta y)^\top B(\alpha-\beta y)
  +a(x)\bigl[G(y)-y^\top p\bigr]\Bigl(\frac{\norm v^2}{R^2}-\beta^2\Bigr).
 \label{gl:eq:perspective-hess}
\end{equation}
If $G$ is three times differentiable at $y$, then for every $v\in\R^d$, with
$u=\alpha-\beta y$,
\begin{equation}
\begin{aligned}
 D^3\Phi_U(x)[v,v,v]
 &=a(x)^2D^3G(y)[u,u,u]
  -3a(x)^2\beta\,u^\top Bu\\
 &\quad-3a(x)^2\Bigl(\frac{\norm v^2}{R^2}-\beta^2\Bigr)u^\top By
  -3a(x)^2\beta\bigl[G(y)-y^\top p\bigr]\Bigl(\frac{\norm v^2}{R^2}-\beta^2\Bigr).
\end{aligned}
\label{gl:eq:perspective-d3}
\end{equation}
\end{lemma}

\begin{proof}
Fix $v\in\R^d$ and write $\alpha=U^\top v$ and $\beta=z(x)^\top v/R^2$. Direct
differentiation of $w(x)=(1+\norm x^2/R^2)^{1/2}$ and $z(x)=x/w(x)$ gives
\[
 Dw(x)v=\frac{x^\top v}{R^2w(x)}=\beta,\qquad
 Dz(x)v=\frac{v}{w(x)}-\frac{x(Dw(x)v)}{w(x)^2}=a(x)\bigl(v-\beta z(x)\bigr),
\]
and hence $Dy(x)v=U^\top Dz(x)v=a(x)(\alpha-\beta y)$ with $u:=\alpha-\beta y$.
By the product and chain rules,
\[
 D\Phi_U(x)v=(Dw(x)v)G(y)+w(x)p^\top Dy(x)v
 =\beta G(y)+w(x)a(x)p^\top(\alpha-\beta y)
 =p^\top\alpha+\bigl[G(y)-y^\top p\bigr]\beta,
\]
since $w(x)a(x)=1$. Because $\alpha=U^\top v$ and $\beta=\ip{z(x)/R^2}{v}$,
this is \eqref{gl:eq:perspective-grad}.

Next differentiate $D\Phi_U(x)v=p(y)^\top\alpha+[G(y)-y^\top p(y)]\beta$ along
the constant vector $v$. Since $\nabla_y[G(y)-y^\top p(y)]=p(y)-(p(y)+B(y)y)
=-B(y)y$, the chain rule yields
\[
 D\bigl[p(y)^\top\alpha\bigr]v=a(x)\alpha^\top B(y)u,\qquad
 D\bigl[G(y)-y^\top p(y)\bigr]v=-a(x)y^\top B(y)u,
\]
while $D\beta(x)v=v^\top Dz(x)v/R^2=a(x)(\norm v^2/R^2-\beta^2)$. Combining the
terms gives
\[
 D^2\Phi_U(x)[v,v]
 =a(x)(\alpha^\top-\beta y^\top)B(y)u
  +a(x)\bigl[G(y)-y^\top p(y)\bigr]\Bigl(\frac{\norm v^2}{R^2}-\beta^2\Bigr),
\]
which is \eqref{gl:eq:perspective-hess}.

Finally, when $G$ is three times differentiable at $y$, differentiate
\eqref{gl:eq:perspective-hess} once more along $v$. Using
$Da(x)v=-a(x)^2Dw(x)v=-a(x)^2\beta$ and
\[
 Du(x)v=-(D\beta(x)v)y-\beta(Dy(x)v)
 =-a(x)\Bigl(\frac{\norm v^2}{R^2}-\beta^2\Bigr)y-a(x)\beta u,
\]
the derivative of $a(x)u^\top B(y)u$ along $v$ equals
\[
 -a(x)^2\beta\,u^\top B(y)u+a(x)^2D^3G(y)[u,u,u]
 -2a(x)^2\Bigl(\frac{\norm v^2}{R^2}-\beta^2\Bigr)u^\top B(y)y
 -2a(x)^2\beta\,u^\top B(y)u.
\]
Similarly, since
\[
 D\Bigl[a(x)\Bigl(\frac{\norm v^2}{R^2}-\beta^2\Bigr)\Bigr]v
 =-a(x)^2\beta\Bigl(\frac{\norm v^2}{R^2}-\beta^2\Bigr)
  +a(x)\Bigl(-2\beta a(x)\Bigl(\frac{\norm v^2}{R^2}-\beta^2\Bigr)\Bigr)
 =-3a(x)^2\beta\Bigl(\frac{\norm v^2}{R^2}-\beta^2\Bigr),
\]
the derivative of $a(x)[G(y)-y^\top p(y)](\norm v^2/R^2-\beta^2)$ along $v$
equals
\[
 -a(x)^2\Bigl(\frac{\norm v^2}{R^2}-\beta^2\Bigr)u^\top B(y)y
 -3a(x)^2\beta\bigl[G(y)-y^\top p(y)\bigr]\Bigl(\frac{\norm v^2}{R^2}-\beta^2\Bigr).
\]
Adding the two expressions yields \eqref{gl:eq:perspective-d3}.
\end{proof}

\subsubsection{$C^\infty$ global perspective witnesses with a Lipschitz Hessian}
\label{gl:subsec:smooth-wrapper}

Now let $\widetilde g_m:\R^m\to\R$ be the $C^\infty$ chain of
Lemma~\ref{x:lem:smooth-chain} formed from the convolved gates
$\widetilde\varphi,\widetilde\psi$ of \eqref{x:eq:gates} with $\eta=1/24$,
$a_0=3/8$, and $a_1=23/24$. Define
\begin{equation}
 \widehat{\widetilde g}_m(y):=\widetilde g_m(y)+\frac94m+\frac14,\qquad
 \widetilde F^{\rm g}_U(x):=w(x)\,\widehat{\widetilde g}_m\bigl(U^\top\rho_R(x)\bigr)-\frac14.
 \label{gl:eq:smooth-wrapper}
\end{equation}
Because $w$ and $\rho_R$ are $C^\infty$ on all of $\R^d$ and $\widetilde g_m$
is $C^\infty$ on $\R^m$, $\widetilde F^{\rm g}_U$ is $C^\infty$ on $\R^d$.

\begin{lemma}[Analytic bounds for the $C^\infty$ global perspective wrapper]
\label{gl:lem:smooth-perspective-wrapper}
Let $m\ge1$ and $R^2\ge64m$. For every orthonormal $m$-frame $U$ in $\R^d$,
the function $\widetilde F^{\rm g}_U$ in \eqref{gl:eq:smooth-wrapper} is $C^\infty$,
coercive, nonconvex, globally $58$-smooth, and globally $12\sqrt m$-Lipschitz,
with initial gap $\widetilde F^{\rm g}_U(0)-\inf_{x\in\R^d}\widetilde F^{\rm g}_U(x)\le
\frac94(m-\frac12)$ and
$\Lip(\nabla^2\widetilde F^{\rm g}_U)<2016$. At every point $x\in\R^d$, its Hessian
satisfies the two-sided global matrix bound
\begin{equation}
 -H^{\rm g}_U\preceq\nabla^2\widetilde F^{\rm g}_U(x)\preceq H^{\rm g}_U
 :=56UU^\top+\frac{84m}{R^2}I,\qquad
 \norm{H^{\rm g}_U}_{\rm op}<58,
 \label{gl:eq:smooth-H}
\end{equation}
and hence at every point $x\in\R^d$,
\begin{equation}
 \tr\bigl[(\nabla^2\widetilde F^{\rm g}_U(x))_+\bigr]
 \le\norm{\nabla^2\widetilde F^{\rm g}_U(x)}_*
 \le\tr H^{\rm g}_U=56m+\frac{84dm}{R^2}.
 \label{gl:eq:smooth-nuclear}
\end{equation}
If any coordinate of $y=U^\top\rho_R(x)$ has magnitude below $a_1=23/24$, then
$\norm{\nabla\widetilde F^{\rm g}_U(x)}>4/5$. The prefix hypotheses of
Lemma~\ref{lem:stage} hold on all of $\R^d$ with $z=\rho_R$, $\tau=a_0=3/8$,
and $c_j=\gamma_{j-1}$, where $\gamma_0=1$ and
$\gamma_{j-1}(x)=\widetilde\psi(-y_{j-1})+\widetilde\psi(y_{j-1})\in[0,1]$ for
$j>1$.
\end{lemma}

\begin{proof}
In \eqref{eq:chain} with the gates $\widetilde\varphi,\widetilde\psi$, one has
$0\le\widetilde\varphi\le9/4$, $0\le\widetilde\psi\le1$, and at most one of
$\widetilde\psi(-t),\widetilde\psi(t)$ is nonzero, so each of the $m$
summands of $\widetilde g_m$ lies in $[-9/4,9/4]$ and the initial summand
$-\widetilde\varphi(y_1)$ lies in $[-9/4,0]$. Hence
$1/4\le\widehat{\widetilde g}_m(y)<\frac92m$ for all $y\in\R^m$. Since
$w(x)\ge1$, we get $\widetilde F^{\rm g}_U(x)\ge(w(x)-1)/4\ge0$, which proves
coercivity and $\inf\widetilde F^{\rm g}_U\ge0$. Because $w(0)=1$ and
$\widetilde g_m(0)=-\widetilde\varphi(0)=-9/8$, we have
$\widetilde F^{\rm g}_U(0)=\frac94(m-\frac12)$.

Write $p=\nabla\widetilde g_m(y)$ and $B=\nabla^2\widetilde g_m(y)$. By
Lemma~\ref{x:lem:smooth-chain}, $\norm p\le10\sqrt m$, $\norm B_{\rm op}\le42$,
$\norm{D^3\widetilde g_m}_{\rm op}\le1872$, $\norm{By}\le44\sqrt m$,
$|y^\top By|\le44m$, and $|y^\top p|\le8m$. Therefore
\[
 \bigl|\widehat{\widetilde g}_m(y)-y^\top p\bigr|
 \le\Bigl(\frac92+8\Bigr)m=\frac{25}2m.
\]
For $R\ge8\sqrt m\ge8$, identity \eqref{gl:eq:perspective-grad} gives
\[
 \norm{\nabla\widetilde F^{\rm g}_U(x)}
 \le10\sqrt m+\frac{25m}{2R}
 \le\Bigl(10+\frac{25}{16}\Bigr)\sqrt m<12\sqrt m.
\]
For any $v\in\R^d$, put $\alpha=U^\top v$ and $\beta=z(x)^\top v/R^2$. Since
$\norm{z(x)}^2/R^2=1-a(x)^2$, Cauchy--Schwarz gives
$\beta^2\le(1-a(x)^2)\norm v^2/R^2$, and hence
\begin{equation}\label{gl:eq:abeta}
 a(x)\beta^2\le a(x)\bigl(1-a(x)^2\bigr)\frac{\norm v^2}{R^2}
 \le\frac2{3\sqrt3}\frac{\norm v^2}{R^2}
 <\frac7{18}\frac{\norm v^2}{R^2}.
\end{equation}
By the bounds on $B$ above and Young's inequality with parameter $\gamma=7/22$,
\[
 a(x)\bigl|(\alpha-\beta y)^\top B(\alpha-\beta y)\bigr|
 \le\bigl(42+44\gamma\bigr)\norm\alpha^2+\Bigl(\frac{44}{\gamma}+44\Bigr)ma(x)\beta^2
 \le56\norm{U^\top v}^2+\frac{638m}{9R^2}\norm v^2.
\]
Adding the second term of \eqref{gl:eq:perspective-hess}, which has magnitude at
most $\frac{25m}{2R^2}\norm v^2$, and noting that
$638/9+25/2=1501/18<84$, we obtain \eqref{gl:eq:smooth-H} and
\eqref{gl:eq:smooth-nuclear} at every $x\in\R^d$. Since $R^2\ge64m$,
$\norm{H^{\rm g}_U}_{\rm op}\le56+84/64<58$.

If some $|y_i|<a_1$, the smallest such index $j$ satisfies $p_j\le-1$ by
Lemma~\ref{x:lem:smooth-chain}. Projecting \eqref{gl:eq:perspective-grad} onto
$u_j$ gives
\[
 u_j^\top\nabla\widetilde F^{\rm g}_U(x)
 =p_j+\bigl[\widehat{\widetilde g}_m(y)-y^\top p\bigr]\frac{y_j}{R^2}
 \le-1+\frac{25m}{2R^2}
 \le-1+\frac{25}{128}=-\frac{103}{128}<-\frac45.
\]
For nonconvexity, choose $x=tu_1$ ($t<0$) so that $y_1=-9/8$ and
$y_2=\dots=y_m=0$. In a neighborhood of this point, $\widetilde\psi(-y_1)=1$
and $\widetilde\psi(y_1)=0$ are constant, because $9/8>a_1$ and $-9/8<a_0$,
so $B_{11}=-\widetilde\varphi''(-9/8)=-4$; indeed $w_\eta$ is supported in
$[-\eta,\eta]$ and $\varphi''=4$ on $[-5/4,-1]$, which contains
$[-9/8-\eta,-9/8+\eta]$. Since $R\ge8$, $1-y_1^2/R^2\ge1-81/4096>31/32$ and
$a(x)=\sqrt{1-y_1^2/R^2}>31/32$. Applying \eqref{gl:eq:perspective-hess} with
$v=u_1$ ($\alpha=e_1$, $\beta=y_1/R^2$) gives
\[
 u_1^\top\nabla^2\widetilde F^{\rm g}_U(x)u_1
 =-4a(x)\Bigl(1-\frac{y_1^2}{R^2}\Bigr)^2
  +a(x)\bigl[\widehat{\widetilde g}_m(y)-y_1p_1\bigr]\frac{1-y_1^2/R^2}{R^2}
 \le-4\Bigl(\frac{31}{32}\Bigr)^5+\frac{25}{128}<-3<0.
\]

For the third derivative, fix a unit vector $v\in\R^d$ and write
$u=\alpha-\beta y$ and $s=\norm{z(x)}/R=\sqrt{1-a(x)^2}\in[0,1]$. Let
$I(y)=\{i:|y_i|\le4/3\}$ be the active coordinate set of
Lemma~\ref{x:lem:smooth-chain}, and let $u_I=P_{I(y)}u$. Since
$\norm{P_{I(y)}y}\le(4/3)\sqrt m$ and $|\beta|\le s/R\le s/(8\sqrt m)$, we have
$\norm{u_I}\le\norm\alpha+|\beta|\norm{P_{I(y)}y}\le1+s/6$, and hence
\[
 a(x)^2\norm{u_I}^3
 \le(1-s^2)\Bigl(1+\frac{s}{6}\Bigr)^3
 <\frac{17}{16},
\]
because on $[0,1]$ the middle expression increases while $5s^2+12s<3$ and
decreases afterwards, and the root of $5s^2+12s=3$ lies in $[0.228,0.229]$,
where the expression is at most $(1-0.228^2)(1+0.229/6)^3<1.061$.
Because $B$ and $D^3\widetilde g_m(y)$ are supported on $I(y)$, identity
\eqref{gl:eq:perspective-d3} gives
\begin{align*}
 \bigl|D^3\widetilde F^{\rm g}_U(x)[v,v,v]\bigr|
 &\le1872\cdot\frac{17}{16}
   +\frac3R\Bigl(56+\frac{638m}{9R^2}\Bigr)
   +\frac3{R^2}\cdot\frac76\cdot44\sqrt m
   +\frac3R\cdot\frac{25m}{2R^2}<2016,
\end{align*}
using $R\ge8\sqrt m\ge8$. Since $D^3\widetilde F^{\rm g}_U(x)$ is a symmetric
trilinear form on $\R^d$, its operator norm equals
$\sup_{\norm v=1}|D^3\widetilde F^{\rm g}_U(x)[v,v,v]|$
\citep{banach1938homogene}, so
$\norm{D^3\widetilde F^{\rm g}_U(x)}_{\rm op}<2016$, which integrates
to $\Lip(\nabla^2\widetilde F^{\rm g}_U)<2016$.

Finally, suppose $y_j=0$ and $|y_k|\le a_0$ for all $k>j$, and put
\[
 r_j(x)=w(x)\widehat{\widetilde g}_m(y_1,\ldots,y_{j-1},0,\ldots,0)-1/4.
\]
The gates $\widetilde\psi(\pm y_k)$ vanish with their derivatives for all $k\ge j$,
so $\widehat{\widetilde g}_m(y)=\widehat{\widetilde g}_m(y_1,\ldots,y_{j-1},0,\ldots,0)$,
$\widetilde F^{\rm g}_U(x)=r_j(x)$, $p_k=0$ for $k>j$, and
$p_j=-\gamma_{j-1}(x)$. Because $y_j=0$, the product $y_jp_j$ vanishes, so
$y^\top p$ agrees with the corresponding inner product for
$(y_1,\ldots,y_{j-1},0,\ldots,0)$. Subtracting \eqref{gl:eq:perspective-grad}
for $r_j$ from \eqref{gl:eq:perspective-grad} for $\widetilde F^{\rm g}_U$
cancels the $z(x)/R^2$ term and leaves
$\nabla\widetilde F^{\rm g}_U(x)-\nabla r_j(x)=-\gamma_{j-1}(x)u_j$. Both $r_j$
and $\gamma_{j-1}$ depend only on $u_1,\ldots,u_{j-1}$, and $y_j=0<a_1$ gives
$\norm{\nabla\widetilde F^{\rm g}_U(x)}>4/5$ by the barrier above, so the
hypotheses of Lemma~\ref{lem:stage} hold with $z=\rho_R$ and $\tau=a_0$.
\end{proof}

\begin{theorem}[$C^\infty$ multistage lower bound on the global fixed-matrix class]
\label{gl:thm:global-stages}
Let $m\ge2$ satisfy
\begin{equation}
 204m\le T,\qquad
 28560\,m^2\log(e+m)\le ET,\qquad
 6144\,m\log(e+m)\le d.
 \label{gl:eq:smooth-stage-conditions}
\end{equation}
Then, for either oracle, there are $C^\infty$, globally $L$-smooth, coercive,
nonconvex witnesses in the global fixed-matrix class $\cF^d_{L,\Delta,E}$ (and
hence in $\cP^d_{L,\Delta,E}$, $\cB^d_{L,\Delta,E}$, and $\cA^d_{L,\Delta,E}$)
giving $N_{\rm det}^{\rm val}\ge N_{\rm det}^{\rm lin}\ge\lfloor d/2\rfloor m$. The
witnesses are globally $2\sqrt{L\Delta}$-Lipschitz, satisfy the global
nuclear norm bound $\norm{\nabla^2f(x)}_*\le LE$ at every $x\in\R^d$, and have
\begin{equation}
 \Lip(\nabla^2f)
 <\frac12\Bigl(\frac{204m}{T}\Bigr)^2\frac{L^2}{\eps}.
 \label{gl:eq:smooth-rho}
\end{equation}
\end{theorem}

\begin{proof}
Set $\ell=\log(e+m)$, $R^2=d/(96\ell)$, $L'=204mL/T$, and $E'=140m\ell$.
Conditions \eqref{gl:eq:smooth-stage-conditions} imply $R^2\ge64m$,
$m\le d/128$, $L'\le L$, and $L'E'=28560m^2\ell L/T\le LE$. By
\eqref{gl:eq:smooth-H}, since $\ell\ge\log(e+2)>3/2$,
\[
 \frac{\tr H^{\rm g}_U}{58}
 =\frac{56m+84\cdot96m\ell}{58}
 <\frac{12152}{87}m\ell<140m\ell=E'.
\]
Put $h=\lfloor d/2\rfloor$. Then $p=d-m-h+1\ge63d/128$ and $M/p\le64m/63$ for
$M\le hm$. Since $\log(e+64m/63)\le(64/63)\ell$, Lemma~\ref{lem:strip-existence}
gives a squared projection bound at most
$(144/25)(128/63)(64/63)/96<(3/8)^2=a_0^2$.
Lemma~\ref{lem:stage} with $z=\rho_R$, $\tau=a_0$, $b=4/5$, and $Q=hm$
therefore applies to $\widetilde F^{\rm g}_U$.

Scale $\widetilde F^{\rm g}_U$ via Lemma~\ref{a:lem:scaling} with $\Lambda=58$ and
$b=4/5$:
\[
 \sigma=\frac{58\eps}{L'b},\qquad
 \alpha=\frac{58\eps^2}{L'b^2},\qquad
 f(x)=\alpha\widetilde F^{\rm g}_U(x/\sigma).
\]
Since $58(9/4)/b^2=6525/32<204$, the initial gap is at most
$204(m-1/2)\eps^2/L'\le\Delta$. The global upper matrix $(L'/58)H^{\rm g}_U$ has
operator norm at most $L'\le L$ and trace at most $L'E'\le LE$ on all of
$\R^d$, and \eqref{gl:eq:smooth-nuclear} gives $\norm{\nabla^2f(x)}_*\le LE$ for
all $x\in\R^d$. Thus $f\in\cF^d_{L,\Delta,E}$, and at the algorithm's output
$\norm{\nabla f(\widehat x)}>\eps$. The Hessian Lipschitz constant of $f$ is at
most
\[
 \frac{\alpha}{\sigma^3}2016
 =\frac{2016b}{58^2}\frac{(L')^2}{\eps}
 =\frac{2016}{4205}\Bigl(\frac{204m}{T}\Bigr)^2\frac{L^2}{\eps}
 <\frac12\Bigl(\frac{204m}{T}\Bigr)^2\frac{L^2}{\eps},
\]
which proves \eqref{gl:eq:smooth-rho}. Finally, $\norm{\nabla f(x)}\le
(12/b)\sqrt m\,\eps=15\sqrt m\,\eps\le2\sqrt{L\Delta}$ because
$T\ge204m$.
\end{proof}

\begin{corollary}[Rate on the global class with a prescribed Hessian Lipschitz constant]
\label{gl:cor:global-rate}
For any $\rho\ge0$, on the $C^\infty$ members of $\cF^d_{L,\Delta,E}$ (and hence
of $\cP^d_{L,\Delta,E}$ and $\cB^d_{L,\Delta,E}$) satisfying
$\norm{\nabla^2f(x)}_*\le LE$ everywhere on $\R^d$ and $\Lip(\nabla^2f)\le\rho$,
if $ET\le d$, then for either oracle,
\begin{equation}
 N_{\rm det}^{\rm val}\ge N_{\rm det}^{\rm lin}\ge
 \max\left\{d,\;\frac{d}{612}
 \min\left\{T,\;\frac{T\sqrt{2\rho\eps}}{L},\;
 \sqrt{\frac{ET}{\log(e+T)}}\right\}\right\}.
 \label{gl:eq:global-rate}
\end{equation}
In particular, if $\rho\ge L^2/(2\eps)$, the middle term in the minimum is
redundant and
\[
 N_{\rm det}\bigl(\cF^d_{L,\Delta,E}\bigr)
 \ge\max\left\{d,\;\frac{d}{612}
 \min\left\{T,\;\sqrt{\frac{ET}{\log(e+T)}}\right\}\right\}
 \ge\frac{d}{672}\min\left\{T,\;\sqrt{\frac{ET}{\log(e+T)}}\right\}.
\]
\end{corollary}

\begin{proof}
Let $M$ denote the minimum of three terms in \eqref{gl:eq:global-rate}, put
$z=M/204$, and set $m=\lfloor z\rfloor$. When $z<3$, the rank-one quadratic
dimension floor $N_{\rm det}^{\rm lin}(\cF^d_{L,\Delta,E})\ge d$ of
Lemma~\ref{lem:dimension-floor} (whose witness is $C^\infty$ with constant
Hessian $Luu^\top$, so $\Lip(\nabla^2f)=0$ and $\norm{\nabla^2f}_*=L\le LE$)
exceeds $dz/3=dM/612$. When $z\ge3$, we have $m\ge3$, $204m\le M\le T$,
$28560m^2\log(e+m)\le204^2m^2\log(e+T)\le ET$, and
$6144m\log(e+m)<2\cdot204^2m\log(e+m)\le m\cdot204^2\cdot m\log(e+T)\le ET
\le d$. Moreover, $(204m/T)L\le\sqrt{2\rho\eps}$, so \eqref{gl:eq:smooth-rho}
gives $\Lip(\nabla^2f)<\rho$. Theorem~\ref{gl:thm:global-stages} yields
$N_{\rm det}^{\rm lin}\ge\lfloor d/2\rfloor m$. Since $m/z\ge3/4$ and
$\lfloor d/2\rfloor/d\ge15/32$ for $d\ge6144$, we have
$\lfloor d/2\rfloor m\ge(45/128)dz>dz/3=dM/612$.
\end{proof}

\begin{corollary}[The $\Omega(dT)$ threshold on the global classes]
\label{gl:cor:global-window}
On the $C^\infty$ subclass of Corollary~\ref{gl:cor:global-rate} with
$\rho\ge L^2/(2\eps)$, and hence on $\cF^d_{L,\Delta,E}$ and
$\cP^d_{L,\Delta,E}$, whenever $E\ge T\log(e+T)$ and $E\le d$ (without
requiring $ET\le d$), both deterministic oracle costs satisfy
\begin{equation}
 N_{\rm det}^{\rm val}\ge N_{\rm det}^{\rm lin}
 \ge\frac{dT}{18432}.
 \label{gl:eq:global-all-window}
\end{equation}
If in addition $T\log(e+T)\le E\le d/T$, then on $\cF^d_{L,\Delta,E}$,
\begin{equation}
 \frac{dT}{612}\le N_{\rm det}\le10\,dT,\qquad
 \frac{ET}{12}\le N_{\rm rand}^{\rm cap}\le23\,ET,
 \label{gl:eq:global-matched-window}
\end{equation}
and the cost ratio for the same oracle $N_{\rm det}/N_{\rm rand}^{\rm cap}$ lies between
$d/(14076\,E)$ and $120\,d/E$.
\end{corollary}

\begin{proof}
For \eqref{gl:eq:global-all-window}, set $z=T/6144$ and $m=\lfloor z\rfloor$.
If $z<3$, Lemma~\ref{lem:dimension-floor} gives $N_{\rm det}\ge d>dz/3
=dT/18432$. If $z\ge3$, then $m\ge3$, $204m\le T$,
$28560m^2\log(e+m)\le T^2\log(e+T)\le ET$, and $6144m\log(e+m)\le
T\log(e+T)\le E\le d$, so Theorem~\ref{gl:thm:global-stages} gives
$N_{\rm det}\ge\lfloor d/2\rfloor m\ge dz/3=dT/18432$. When
$T\log(e+T)\le E\le d/T$, we also have $ET\le d$ and
$\min\{T,\sqrt{ET/\log(e+T)}\}=T$, so Corollary~\ref{gl:cor:global-rate} gives
$N_{\rm det}\ge dT/612$, while coordinate differences,
Theorem~\ref{b:thm:stp}, and Corollary~\ref{cor:quadratic-baseline} give the
remaining bounds in \eqref{gl:eq:global-matched-window}.
\end{proof}

\section{Further Results for Randomized Methods}
\label{rb:sec:main}
This appendix shows that the hard functions of
Theorem~\ref{thm:linear-lower} do not extend the deterministic bound
$\Omega(dT)$ to randomized methods (Appendix~\ref{rb:sec:parity}), and
combines our bounds with the lower bound of
\citet{xia2026global} into a wider window with matching bounds for exact
values (Appendix~\ref{rb:sec:xy}).

\subsection{The unrestricted hard functions are easy for randomized methods}
\label{rb:sec:parity}
Theorem~\ref{thm:linear-lower} uses the functions
$\tilde F_{m,U}(x)=g_m(U^\top\rho_R(x))+q(x)$ with $R^2=5m$, where
$\rho_R(x)=x/\sqrt{1+\norm x^2/R^2}$ and the public potential $q$ vanishes
on $\norm x\le r_0=2\sqrt m$. A deterministic method must pay order $d$
calls for each of the $m$ hidden directions. A randomized method can do
better on this family, because the minimizer depends only on the sum of the
hidden directions, and this sum can be read in batches.

\begin{lemma}[Stages of one parity are read together]
\label{rb:lem:parity}
Let $U=(u_1,\ldots,u_m)$ have orthonormal columns, and let
$B\subseteq\{1,\ldots,m-1\}$ be a nonempty set of indices of the same
parity. Suppose $z_B\in\R^d$ satisfies
\begin{equation}
 \ip{u_i}{z_B}\ge\frac{13}{8}\quad(i\in B),
 \qquad
 |\ip{u_k}{z_B}|\le\frac38\quad(k\notin B).
 \label{rb:eq:trigger}
\end{equation}
Then for every $z$ with $\norm{z-z_B}\le1/16$,
\[
 g_m(U^\top z)=c_B-\Bigl\langle u_1\mathbf 1_{\{1\notin B\}}
 +\sum_{i\in B}u_{i+1},\,z\Bigr\rangle,
\]
where $c_B=-\frac98\mathbf 1_{\{1\notin B\}}-\frac94\mathbf 1_{\{1\in B\}}
-\frac98|B|$. Moreover, at $z_*=\frac54\sum_{j=1}^mu_j$ the chain attains
its minimum $-\frac94m$, and its gradient in $z$ vanishes.
\end{lemma}
\begin{proof}
Indices of the same parity are never consecutive, so $B$ and
$\{i+1:i\in B\}$ are disjoint. For $\norm{z-z_B}\le1/16$, the coordinates
$y=U^\top z$ satisfy $y_i\ge25/16>5/4$ for $i\in B$ and $|y_k|\le7/16<1/2$
for $k\notin B$. In \eqref{eq:chain}, the first term is $-9/4$ if $1\in B$
and $-(y_1+9/8)$ otherwise. For $r\ge2$, the $r$th term vanishes when
$r-1\notin B$, because then $\psi(\pm y_{r-1})=0$. When $r-1\in B$, we have
$r\notin B$, so $\psi(y_{r-1})=1$, $\psi(-y_{r-1})=0$, and
$\varphi(y_r)=y_r+9/8$. Summing gives the displayed affine function. At
$z_*$ every coordinate equals $5/4$, where $\varphi=9/4$ and $\psi=1$ and
both have zero derivative, so each of the $m$ terms equals its minimum
$-9/4$.
\end{proof}

\begin{proposition}[A randomized method on the unrestricted hard functions]
\label{rb:prop:parity}
Let $m\ge2$ and $d\ge2+80m\log(24m^2)$, and consider the scaled functions
$f_U(x)=\alpha\tilde F_{m,U}(x/\sigma)$ of the proof of
Theorem~\ref{thm:linear-lower}. There is a randomized method using exact values that,
for every orthonormal frame $U$, uses at most
\[
 2m+17(d+1)\sqrt{m\log(24m^2)}
\]
values and, with probability at least $5/6$, outputs an unqueried exact
global minimizer $\widehat x$ of $f_U$, so that $\nabla f_U(\widehat x)=0$.
If $500<T\le d/\log(ed)$ and $m$ is the number of stages used in
Theorem~\ref{thm:linear-lower}, these hypotheses hold and the bound is
$O(d\sqrt{T\log(e+T)})$, whereas every deterministic method needs
$\Omega(dT)$ values on the same functions.
\end{proposition}
Algorithm~\ref{alg:parity} collects the steps of the following proof.

\begin{algorithm}[t]
\caption{Randomized method on the hard functions of
Proposition~\ref{rb:prop:parity}. As in the proof, a query at a point $z$ with
$\norm z<R$ is made at $\rho_R^{-1}(z)$ and returns $g(z)=g_m(U^\top z)$ after
the public value $q$ is subtracted. For a point $z$ and an orthonormal list
$W$, $\textsc{Diff}(z,W)=\sum_{v\in W}\delta^{-1}[g(z)-g(z+\delta v)]\,v$ uses
$1+|W|$ values; $e_1,\ldots,e_d$ are the coordinate vectors.}
\label{alg:parity}
\begin{algorithmic}[1]
\Require $m$, $d$, the public $\alpha,\sigma,R,q,\rho_R$, and exact values of $f_U$
\State $\delta\gets1/16$, $R_{\rm eff}\gets\frac34R$,
$K\gets\lfloor\sqrt{m/(40\log(24m^2))}\rfloor$,
$b\gets\lceil48dK\log(24m^2)/R_{\rm eff}^2\rceil$
\State $u_1\gets\textsc{Diff}(0,(e_1,\ldots,e_d))$ \hfill{\itshape Step 0}
\If{$K=0$} \hfill{\itshape small $m$}
  \For{$j=2,\ldots,m$}
    \State $u_j\gets\textsc{Diff}(2u_{j-1},(e_1,\ldots,e_d))-u_1\mathbf 1_{\{j\ne2\}}$
  \EndFor
  \State $s\gets\sum_{j=1}^mu_j$
\Else \hfill{\itshape Phase 1}
  \State split the odd and the even indices of $\{1,\ldots,m-1\}$ into batches of at most $K$ indices
  \State for each batch, independently, draw a uniformly random orthonormal basis of $u_1^\perp$ and give each index $i\ne1$ of the batch its own block $S_i$ of $b$ consecutive basis vectors
  \State $\hat p_1\gets u_1$
  \For{$j=2,\ldots,m-1$}
    \State $g_j\gets\textsc{Diff}(R_{\rm eff}\hat p_{j-1},\text{basis of }S_j)$
    \State $\hat p_j\gets g_j/\norm{g_j}$, or a fixed unit vector of $S_j$ if $g_j=0$
  \EndFor
  \State $s\gets u_1$ \hfill{\itshape Phase 2}
  \For{each batch $B$}
    \State $z_B\gets(R_{\rm eff}/\sqrt{|B|})\sum_{i\in B}\hat p_i$
    \State $s\gets s+\textsc{Diff}(z_B,(e_1,\ldots,e_d))-u_1\mathbf 1_{\{1\notin B\}}$
  \EndFor
\EndIf
\State $z_*\gets\frac54s$
\State \Return $\widehat x=\sigma\rho_R^{-1}(z_*)$ if $\norm{z_*}<R$, and $\widehat x=0$ otherwise
\end{algorithmic}
\end{algorithm}

\begin{proof}
The constants $\alpha,\sigma>0$ are public, so we work with $\tilde F_{m,U}$.
At a query point $x$ with $\rho_R(x)=z$, subtracting the public value $q(x)$
returns $g_m(U^\top z)$, and $\rho_R^{-1}(z)=z/\sqrt{1-\norm z^2/R^2}$ for
$\norm z<R$. Put $\delta=1/16$,
\[
 R_{\rm eff}=\tfrac34R,\qquad
 K=\Bigl\lfloor\sqrt{\tfrac{m}{40\log(24m^2)}}\Bigr\rfloor,\qquad
 b=\Bigl\lceil\frac{48dK\log(24m^2)}{R_{\rm eff}^2}\Bigr\rceil .
\]
The assumption on $d$ is equivalent to
$R_{\rm eff}\le\frac3{16}\sqrt{(d-2)/\log(24m^2)}$. All points $z$ used
below satisfy $\norm z\le R_{\rm eff}+\delta<R$. If $B$ and $z_B$ satisfy
the hypotheses of Lemma~\ref{rb:lem:parity}, the $d+1$ values at $z_B$ and
$z_B+\delta e_a$, $a=1,\ldots,d$, give the difference quotients
$\delta^{-1}[g_m(U^\top z_B)-g_m(U^\top(z_B+\delta e_a))]$, which are
exactly the coordinates of $u_1\mathbf 1_{\{1\notin B\}}+\sum_{i\in B}u_{i+1}$.

\emph{Step 0.} Near $z=0$ every $|y_i|\le\delta<1/2$, so the chain equals
$-(y_1+9/8)$ there. The $d+1$ values at $z=0$ and $z=\delta e_a$,
$a=1,\ldots,d$, give $u_1$ exactly.

\emph{Small $m$.} If $K=0$, then $m<40\log(24m^2)$. For $j=2,\ldots,m$ in
turn, the point $2u_{j-1}$ satisfies \eqref{rb:eq:trigger} with $B=\{j-1\}$,
so its $d+1$ values give $u_1\mathbf 1_{\{j\ne2\}}+u_j$, and hence $u_j$.
This deterministic method finds $\sum_ju_j$ exactly with
$m(d+1)<(d+1)\sqrt{40m\log(24m^2)}$ values and outputs $\widehat x$ as in
Phase~2 below. Assume from now on that $K\ge1$, so that
$K^2\le m/(40\log(24m^2))$ and $m\ge645$. Since
$R_{\rm eff}^2=\frac{45}{16}m$, we have $Kb\le\frac{32}{75}d+K\le(d-1)/2$.

\emph{Phase 1.} Split the odd and the even indices of $\{1,\ldots,m-1\}$
into batches of at most $K$ indices each; there are at most
$\lceil(m-1)/(2K)\rceil+\lceil(m-2)/(2K)\rceil\le\min\{m-1,\,m/K+2\}$ batches. Within each batch, take
mutually orthogonal $b$-dimensional subspaces $S_i$ of $u_1^\perp$, one for
each index $i\ne1$, as consecutive blocks of a uniformly random orthonormal
basis of $u_1^\perp$, independently across batches; this is possible because
$Kb\le(d-1)/2$. Put $\hat p_1=u_1$. For $j=2,\ldots,m-1$ in turn, take
$z^{(j)}=R_{\rm eff}\hat p_{j-1}$, evaluate at $z^{(j)}$ and at
$z^{(j)}+\delta v$ for an orthonormal basis $v$ of $S_j$, put
\[
 g_j=\sum_v\delta^{-1}\bigl[g_m(U^\top z^{(j)})
 -g_m(U^\top(z^{(j)}+\delta v))\bigr]\,v\in S_j,
\]
and set $\hat p_j=g_j/\norm{g_j}$, or a fixed unit vector of $S_j$ if
$g_j=0$.

\emph{Probability.} Fix $U$, put $\eta=\sqrt{4\log(24m^2)/(d-2)}$, so that
$R_{\rm eff}\eta\le3/8$, and for $i\ge2$ let
$\hat p_i^{\rm true}=P_{S_i}u_i/\norm{P_{S_i}u_i}$. Consider three kinds of
events.
\begin{enumerate}[label=(\roman*),leftmargin=1.6em,itemsep=1pt,topsep=2pt]
\item \emph{Signal}, for $2\le i\le m-1$: $\norm{P_{S_i}u_i}^2\ge b/(3d)$.
Each $S_i$ is uniform among the $b$-dimensional subspaces of the
$(d-1)$-dimensional space $u_1^\perp$, which contains $u_i$, so
$\norm{P_{S_i}u_i}^2$ has the law of $X/(X+Y)$ for independent
$X\sim\chi^2_b$ and $Y\sim\chi^2_{d-1-b}$. The event holds if $X\ge b/2$ and
$X+Y\le\frac32(d-1)$. By Chernoff bounds for $\chi^2$ variables, these fail
with probability at most $e^{-b/11}+e^{-(d-1)/22}\le1/(12m)$, since
$b\ge11\log(24m)$ and $d-1\ge22\log(24m)$.
\item \emph{Triggers}, for $2\le i\le m-2$ and $k\notin\{1,i\}$:
$|\ip{u_k}{\hat p_i^{\rm true}}|\le\eta$. The law of $S_i$ is invariant
under the rotations of $u_1^\perp$ that fix $u_i$, so the component of
$\hat p_i^{\rm true}$ orthogonal to $u_i$ has a uniform direction in the
$(d-2)$-dimensional space $\{u_1,u_i\}^\perp$, which contains $u_k$. The
tail bound in the proof of Lemma~\ref{a:lem:cap-union} shows that the event
fails with probability at most $2e^{-(d-2)\eta^2/2}=1/(288m^4)$.
\item \emph{Batches}, for each batch $B$ and each $k\ge2$:
$|\ip{u_k}{w_{B,k}}|\le\eta\sqrt{|B|}$, where
$w_{B,k}=\sum_{i\in B\setminus\{1,k\}}\hat p_i^{\rm true}$. The joint law of
the subspaces of $B$ is invariant under the rotations of $u_1^\perp$ that fix
every $u_i$ with $i\in B\setminus\{1,k\}$, and these rotations act on each
such $\hat p_i^{\rm true}$ by rotation. Hence the component of $w_{B,k}$
orthogonal to $A=\operatorname{span}\{u_1,u_i:i\in B\setminus\{1,k\}\}$ has a
uniform direction in $A^\perp$, which contains $u_k$ and has dimension at
least $d-1-K\ge(d-2)/2$. As $\norm{w_{B,k}}\le\sqrt{|B|}$, the event fails
with probability at most $2e^{-(d-2)\eta^2/4}=1/(12m^2)$.
\end{enumerate}
There are $m-2$ signal events, fewer than $m^2$ trigger events, and at most
$m(m-1)$ batch events, so all of them hold with probability at least
$1-\frac{2m-3}{12m}-\frac1{288m^2}>\frac56$.

On this event, $\hat p_j=\hat p_j^{\rm true}$ for $2\le j\le m-1$, by
induction on $j$. Indeed, $z^{(j)}$ satisfies \eqref{rb:eq:trigger} with
$B=\{j-1\}$: its inner product with $u_{j-1}$ is $R_{\rm eff}\ge2$ if $j=2$
and otherwise
$R_{\rm eff}\norm{P_{S_{j-1}}u_{j-1}}\ge R_{\rm eff}\sqrt{b/(3d)}
\ge4\sqrt{K\log(24m^2)}\ge2$, and its other inner products vanish or, by
the trigger events, are at most $R_{\rm eff}\eta\le3/8$ in absolute value. By
Lemma~\ref{rb:lem:parity} the chain is affine near $z^{(j)}$ with gradient
$-u_2$ if $j=2$ and $-(u_1+u_j)$ otherwise. Every $v\in S_j$ is orthogonal
to $u_1$, so the difference quotients are exact, $g_j=P_{S_j}u_j\ne0$, and
$\hat p_j=\hat p_j^{\rm true}$. For each batch $B$ put
\[
 z_B=\frac{R_{\rm eff}}{\sqrt{|B|}}\sum_{i\in B}\hat p_i,
\]
which has norm $R_{\rm eff}$, because the $\hat p_i$ of one batch are
orthonormal. For $i\in B$ with $i\ge2$, the signal and batch events give
$\ip{u_i}{z_B}\ge R_{\rm eff}\sqrt{b/(3dK)}-R_{\rm eff}\eta\ge2-\frac38=\frac{13}8$;
if $1\in B$, then $\ip{u_1}{z_B}=R_{\rm eff}/\sqrt{|B|}\ge2$; and for
$k\notin B$, $\ip{u_k}{z_B}$ vanishes if $k=1$ and otherwise has absolute
value at most $R_{\rm eff}\eta\le\frac38$ by the batch events.

\emph{Phase 2.} For each batch $B$, the $d+1$ values at $z_B$ and
$z_B+\delta e_a$ give $u_1\mathbf 1_{\{1\notin B\}}+\sum_{i\in B}u_{i+1}$,
and hence $\sum_{i\in B}u_{i+1}$. The batches partition
$\{1,\ldots,m-1\}$, so $u_1$ plus the sum of all batch vectors equals
$\sum_{j=1}^mu_j$ exactly. Let $z_*$ be $\frac54$ times this computed sum,
and output $\widehat x=\sigma\rho_R^{-1}(z_*)$, or $\widehat x=0$ if
$\norm{z_*}\ge R$. On the event above, $z_*=\frac54\sum_ju_j$,
$\norm{z_*}^2=\frac{25}{16}m$, and
\[
 \norm{\rho_R^{-1}(z_*)}^2=\frac{25m/16}{1-5/16}=\frac{25}{11}m<4m=r_0^2,
\]
so $q$ and $\nabla q$ vanish there, and Lemma~\ref{rb:lem:parity} shows that
$\widehat x$ is a global minimizer with zero gradient.

\emph{Cost.} The method uses at most
$d+1+(m-2)(b+1)+(d+1)(m/K+2)$ values. Here
$(m-2)(b+1)\le\frac{256}{15}dK\log(24m^2)+2m
\le\frac{256}{15\sqrt{40}}\,d\sqrt{m\log(24m^2)}+2m$
and $m/K\le2\sqrt{40m\log(24m^2)}$, which give the stated bound since
$\sqrt{m\log(24m^2)}\ge100$.

\emph{Range.} If $500<T\le d/\log(ed)$, then
Theorem~\ref{thm:linear-lower} uses $m=\lfloor T/275+1/2\rfloor\ge2$, so
$m\le\frac43(m-\frac12)\le d/(206\log(ed))$. Hence $24m^2<d^2$ and
$80m\log(24m^2)<\frac{160}{206}d\le d-2$, because $d>4700$. Finally
$2m\le d$ and $\sqrt{m\log(24m^2)}=O(\sqrt{T\log(e+T)})$, and
Theorem~\ref{thm:linear-lower} gives the deterministic bound $dT/500$.
\end{proof}

The hard functions of Theorem~\ref{thm:linear-lower}, on which deterministic
methods pay $\Omega(dT)$, are solved with $O(d\sqrt{T\log(e+T)})$ values by
the proposition, so they cannot give a randomized lower bound of that order.
For exact values, \citet{xia2026global} prove a randomized lower bound of order $dT$ with
a different construction when $T\ge c$ and $d\ge CT\log(2+T)$ for absolute constants $c,C>0$;
whether such a bound also holds under gradient projections is open.

\subsection{Exact values: a wider window with matching bounds}
\label{rb:sec:xy}
For exact values, the concurrent and independent lower bound of
\citet[Theorem~2.3]{xia2026global} extends the randomized bound
$\Omega(ET)$ beyond $ET\le d$. Combined with our deterministic bounds, it
removes the condition $E\le d/T$ from Corollary~\ref{cor:main-window} when
$T$ is at most of order $d/\log(ed)$. Their bound concerns exact values
only, so the rank-priced oracle is not covered, and the constants are no
longer explicit.

\begin{corollary}[A wider window for exact values]
\label{rb:cor:xy-window}
There are absolute constants $c_1,c_2>0$ such that, for exact values on every
class between $\cF^d_{L,\Delta,E}$ and $\cB^d_{L,\Delta,E}$, if
$ET\le c_1d^2/\log(ed)$, then $N_{\rm rand}\ge c_2\,ET$. Consequently, if
$T\log(e+T)\le E\le d$ and $T\le c_1d/\log(ed)$, then
\[
 \frac{dT}{18432}\le N_{\rm det}\le10\,dT,\qquad
 c_2\,ET\le N_{\rm rand}\le23\,ET.
\]
\end{corollary}

\begin{proof}
The deterministic lower bound is Corollary~\ref{gl:cor:global-window}, which
requires only $T\log(e+T)\le E\le d$. The two upper bounds and the randomized
lower bound $\min\{d,ET\}/12$ are in Theorem~\ref{a:thm:main}. Lower bounds
proved on $\cF^d_{L,\Delta,E}$ and upper bounds proved on $\cB^d_{L,\Delta,E}$
hold on every class between them. It remains to bound $N_{\rm rand}$ from
below when $ET>d$.

Put $\ell=LE/d\le L$ and $\tau=\ell\Delta/\eps^2=ET/d>1$. Every globally
$\ell$-smooth function with $f(0)-\inf f\le\Delta$ lies in
$\cF^d_{L,\Delta,E}$, with the upper model $\ell I_d$, whose norm is
$\ell\le L$ and whose trace is $LE$. Theorem~2.3 of \citet{xia2026global}
uses the same starting point $0$, the same gap $f(0)-\inf f$, a cap on the
number of exact values, and success probability at least $2/3$ on every
function. We apply it with smoothness $\ell$, gap $\Delta$, and a Lipschitz
constant $L_0\ge\sqrt{\ell\Delta}$, so that $\min\{\ell\Delta,L_0^2\}=\ell\Delta$.
It provides absolute constants $c,C,c_0>0$ such that, if $\tau\ge1/c_0$ and
$d\ge C\tau\log(2+\tau)$, every randomized algorithm needs at least
$c\,d\tau=c\,ET$ exact values on its hard functions; these are globally
$\ell$-smooth with gap at most $\Delta$, and hence lie in
$\cF^d_{L,\Delta,E}$. If $1<\tau<1/c_0$, the bound $\min\{d,ET\}/12$ gives
$N_{\rm rand}\ge d/12>c_0ET/12$. Finally, let $c_1=\min\{1,1/(2C)\}$. Then
$\tau=ET/d\le c_1d/\log(ed)\le d$, and $2+d\le(ed)^2$ gives
$\tau\log(2+\tau)\le c_1d\log(2+d)/\log(ed)\le2c_1d\le d/C$. Hence the
corollary holds with $c_2=\min\{1/12,\,c_0/12,\,c\}$, and the cost ratio
$N_{\rm det}/N_{\rm rand}$ is of order $d/E$.
\end{proof}

\section{Experimental Protocol}\label{app:empirical}
These experiments illustrate the mechanisms behind the main theorems and the
plausibility of the curvature assumption. All synthetic runs use
$L=\Delta=1$, $x_0=0$, and $\eps=T^{-1/2}$, and they charge every function
value, including difference probes, rejected proposals, line search centers,
and final checks. Three metrics are reported separately: the offline first-hit
cost, that is, the number of calls before the first retained iterate with
$\norm{\nabla f}\le\eps$, computed by an observer that does not affect the run;
whether the output is $\eps$-stationary; and the total call count.

\paragraph{Randomized versus deterministic.}
The first family is the rank-one quadratic
$f_u(x)=\frac12(\ip ux-\sqrt2)^2$, with $E=1$. The second uses
$\phi(t)=0.1t^2+0.8(1-\cos t)$ and
\[
 f(x)=a^2\sum_{i=1}^dh_i\phi\bigl((Q^\top x)_i/a+\theta_i\bigr),\qquad
 \theta=(2,0,\ldots,0),\qquad a=\frac1{\sqrt{\phi(2)}},
\]
with $\max_ih_i=1$ and $\sum_ih_i=E$. Since $\phi''\in[-0.6,1]$, one has
$\nabla^2f\preceq Q\operatorname{diag}(h)Q^\top$ everywhere, so $f$ lies in the
fixed-matrix class; it is nonconvex and coercive, and $f(0)-\inf f=1$. At
$x_0=0$ only the first rotated coordinate is away from its minimizer, since
$\phi'(0)=0$, so along the line through $x_0$ in direction $Qe_1$ the gradient
stays parallel to $Qe_1$; random directions still pay the curvature of the
other coordinates. The
concentrated spectrum sets $h_1=1$, assigns $0.1/(d-1)$ to each other
eigenvalue, and distributes the remaining trace in order without exceeding
one; the diffuse spectrum sets $h_1=1$ and $h_i=(E-1)/(d-1)$ for $i>1$. The
rotation is the Householder reflection $Q=I-2vv^\top$ with
$v=(e_1-u)/\norm{e_1-u}$ and $u$ uniform on the sphere. The grid uses
$d\in\{10^3,10^4,10^5\}$, $T\in\{2,8,32\}$, and $E\in\{2,4,8,\ldots,128\}$ for
the nonconvex family, keeps the cells with $ET\le d/4$, and has five rotation
seeds per cell. The three-point method uses $h=4\eps/(5L(E+1))$,
$N=\lceil10(E+1)T\rceil$ steps, an output drawn uniformly and independently from the iterates before the updates, and
the hard cap $1+2N$ of Theorem~\ref{x:thm:stp}; it is also run with a fixed
seed. The deterministic methods are coordinate central differences and
finite-difference L-BFGS on both families, and Krylov conjugate gradients on
the quadratic family.

On the regular grid ($575$ runs per method, $45$ for Krylov), every method
reached the target offline in every run. The returned output was $\eps$-stationary in $524$ of
$575$ runs of the three-point method, in $509$ of $575$ with a fixed seed, and
in all runs of the deterministic methods. At $d=10^5$, the log--log slopes of
the first-hit cost of the three-point method range from $0.54$ to $0.75$ in $T$
and from $0.90$ to $0.95$ in $E$. The $27$ targeted runs comprise nine each of
three-point method with a fixed seed, coordinate differences, and finite-difference L-BFGS.
Each uses a rank-one quadratic chosen orthogonal to all of the run's queries, a
property certified in exact integer arithmetic ($15$ distinct functions). All
$27$ runs ended at gradient norm $\sqrt2$ without an offline hit and are
reported as right-censored, whereas $20$ fresh seeds on each of these functions succeeded in
$17$--$20$ runs.

\paragraph{Stronger baselines.}
On the $43$ cells of the regular grid with $E\in\{1,4,16\}$ ($215$
instances, with the same functions and rotation seeds), we also ran two
further methods. The Gaussian two-point method uses
$g_k=[f(x_k+\mu a_k)-f(x_k-\mu a_k)]a_k/(2\mu)$ with $a_k\sim N(0,I_d)$, step
$1/(L(E+2))$, $N=\lceil(6E+13)T\rceil$ steps,
$\mu=\eps/(4L(6E+13)\sqrt{d(d+2)(d+4)})$, and an output drawn uniformly from
the iterates before the updates, for $2N$ calls in total. The restarted
subspace search uses the capacity given below.
The two-point method returned $\eps$-stationary outputs in $203$ of $215$ runs
and the subspace search in all $215$; paired with the three-point method on the
same functions and seeds, the subspace search reached the target later in all
$43$ cells, by a median factor between $8.7$ and $1.8\times10^4$. As in
the targeted runs above, the subspace search with budgets $32$, $64$, and $128$
below $d$ ended at gradient norm $\sqrt2$ on six rank-one quadratics orthogonal
to all of its queries and its output, whereas $20$ fresh seeds of the
three-point method per function succeeded in $114$ of $120$ runs.

\paragraph{Monotone two-point method and wall functions.}
We also ran the monotone two-point method of
Theorem~\ref{b:thm:monotone} on the same $215$ instances, with the step, the
number of updates, and the difference radius of the two-point method above,
forward differences, and $2N+1$ calls. Its first hit came after a median of
$31$ calls, against $30$ for the two-point method; in the cells of
Figure~\ref{fig:empirical-mechanism}(a) it trailed the two-point method by about
one call, its initial query. Its uniformly drawn output was $\eps$-stationary in
$203$ of $215$ runs, as for the two-point method, and its last iterate in all
$215$. We also built wall functions $f(x)=h(Px)+\beta(\|(I-P)x\|^2)$,
where $P$ projects onto a random subspace of dimension $E_0$, $h$ is a sum of
rescaled one-dimensional profiles with $|h''|\le L$ and $h(0)-\inf h=\Delta/2$,
and $\beta(s)=LE_0s/(2d)$ for $s\le s_1=d\Delta/(LE_0)$, beyond which $\beta'$
rises smoothly to $L/4$ with $2\beta'+4s\beta''\le L$. Below $f(0)$ one has
$s<s_1$, so these functions lie in $\cB^d_{L,\Delta,2E_0}$, whereas above $f(0)$
the Hessian trace reaches about $Ld/2$. On $45$ instances ($9$ cells with
$d\in\{10^3,10^4,10^5\}$, $T\in\{8,32\}$, $E_0\in\{4,16\}$, and $8E_0T\le d$,
five seeds each), the last iterate of the monotone two-point method was
$\eps$-stationary in all runs, and the uniformly drawn outputs of the two-point
and three-point methods in $44$ and $42$; their median first hits were $69$,
$72$, and $173$ calls. No method
exceeded $f(0)$. The two-point method with step $1/(L(d+2))$, which is safe
without an acceptance test, reached the target in none of the $45$ runs within
the same budget.

\paragraph{Subspace search.}
The restarted subspace search uses capacity
$r=\lceil\sqrt{dET/(d+T)}\rceil$, a variant without restarts uses $r=d$, and
coordinate central differences follow Appendix~\ref{a:sec:upper}. On the two
families above with $d\in\{10^3,10^4\}$, $T\in\{2,8,32,128,512\}$, $E=1$ for
the quadratic, and $E\in\{2,4,8,16\}$ for the nonconvex family ($1{,}800$
runs including the three-point method), all
deterministic runs returned $\eps$-stationary outputs and the three-point method
did so in $425$ of $450$ runs. Every subspace run used exactly one expansion and
no restart, consistent with the gradient staying along $Qe_1$; on the nonconvex family, the paired median ratio of total calls
between the subspace search and coordinate differences is about $0.67$ for
$T\le128$ and $0.50$ at $T=512$.

The chain has $E\in\{8,16\}$, $m=2E$,
$a=\sqrt{8(m+1)/m}$, a fixed Householder reflection $R$, $z=(Rx)_{1:m}$, and
\[
 f(x)=\frac18\Bigl[(z_1-a)^2+\sum_{i=1}^{m-1}(z_i-z_{i+1})^2+z_m^2\Bigr]
 -\frac1m .
\]
Its global upper model $H=\frac14R^\top\operatorname{diag}(K_m,0)R$, where $K_m$
is tridiagonal with diagonal $2$ and off-diagonal $-1$, satisfies
$\norm H_{\rm op}<1$ and $\operatorname{tr}H=E$; moreover $f(0)=1$ and
$\inf f=0$. At a point supported on the first $j$ active coordinates, the
gradient adds only coordinate $j+1$, so each expansion can add at most one
direction. With $d\in\{10^3,10^4\}$, $T\in\{2,8,32,128,512\}$, and five seeds,
all $300$ runs returned $\eps$-stationary outputs; $240$ of them satisfy
$ET\le d$. The number of expansions was $1,2,3,6,10$ at the five accuracies,
with no restart. At $d=10^4$, the paired median ratio of total calls was about
$1.00$ for $T\le8$, $0.57$ at $T=32$, $0.39$ to $0.41$ at $T=128$, and $0.25$ to
$0.26$ at $T=512$. Over the five accuracies, the log--log slope of the total
cost is about $0.31$ for the subspace search and $0.57$ for coordinate
differences, both below the exponents $1/2$ and $1$ of the worst-case upper
bounds.

\paragraph{Local curvature of GPT-2.}
GPT-2 uses the GELU activation, so its loss is smooth in the parameters. We
ran MeZO on SST-2 and RTE with three seeds each and $1{,}024$ updates, reaching
final accuracies of $76.6$--$85.2\%$ and $50.8$--$53.9\%$, and at updates
$0,256,512,1024$ measured the Hessian $G$ of a fixed training batch in
all $d=124{,}439{,}808$ parameters, in FP32. Since
$\operatorname{tr}(G_+)=(\operatorname{tr}G+\|G\|_*)/2$ and
$\|G\|_*\le\sqrt d\,\|G\|_F$, any $u_1\ge\operatorname{tr}G$,
$u_2\ge\|G\|_F^2$ and $\ell\in(0,\norm G_{\rm op}]$ give
\[
\frac{\operatorname{tr}(G_+)}{d\norm G_{\rm op}}\le
\frac{u_1/d+\sqrt{u_2/d}}{2\ell}.
\]
We take for $\ell$ the absolute Rayleigh quotient after a $32$-step power
iteration. From $m=128$ Rademacher probes, one Hessian-vector product each, we
form the Hutchinson estimates $\hat\sigma_1$ of $\operatorname{tr}G$ and
$\hat\sigma_2$ of $\operatorname{tr}(G^2)=\|G\|_F^2$
\citep{hutchinson1989stochastic}, and set $u_2=\hat\sigma_2/(1-\rho)$ with
$\rho=\sqrt{6\log(2/\delta)/m}\approx0.595$ and
$u_1=\hat\sigma_1+\sqrt{2u_2/(m\delta)}$, where $\delta=0.025/24$. As in the
proof of Theorem~1 of \citet{roosta2014trace}, Markov's inequality, convexity
and $\mathbb E e^{-tz^2}\le1-t+3t^2/2$ for $z=a^\top w$ with $\|a\|=1$ and
Rademacher $w$ give $\Prob(u_2<\|G\|_F^2)\le e^{mt(1-\rho)}(1-t+3t^2/2)^m$,
which is below $1.3\times10^{-4}$ at $t=0.22$. Each Rademacher sample of
$\operatorname{tr}G$ has variance at most $2\|G\|_F^2$, so by Chebyshev's
inequality $u_2\ge\|G\|_F^2$ and $u_1<\operatorname{tr}G$ hold together with
probability at most $\delta$. A union bound over the $24$ points shows that,
with probability at least $0.95$, every plotted value bounds the ratio from
above. These bounds range from $6.70\times10^{-5}$ to $9.19\times10^{-5}$ on
SST-2 and from $6.83\times10^{-5}$ to $1.16\times10^{-4}$ on RTE, and decrease
along every run.

\paragraph{Low-dimensional diagnostics.}
Comparisons on a related separable family at $d\in\{16,64,128\}$ had
$ET>d$ in every cell, outside the regime where randomized costs are known to
be of order $ET$. There, at the isotropic endpoint
$E=d$ the coordinate methods were faster, random directions were slower in
most cells at $d=16$, and diffuse spectra gave heterogeneous results.
Curvature measurements on OPT models use the ReLU activation, so their Hessian
products concern one activation branch only.

\section{Lean Formalization of the Main Bounds}
\label{app:lean}
Theorems~\ref{thm:main-random} and~\ref{thm:main-det} and
Corollary~\ref{cor:main-window} on $\cB^d_{L,\Delta,E}$, together with the
randomized lower bound of Corollary~\ref{cor:quadratic-baseline} on
$\cF^d_{L,\Delta,E}$, are formalized, in the form stated below, in Lean~4
(version 4.19.0) with the Mathlib library (revision \texttt{c44e0c8}) and
checked by the Lean kernel. \leanavailability{} Its README lists the Lean
name of each statement below. Beyond propositional extensionality, the axiom
of choice, and the soundness of quotients, on which Mathlib rests, the proofs
use no axioms or unproved assumptions.

\paragraph{Model.}
Queries, outputs, and costs follow Section~\ref{sec:model}. A call $(x,V)$
returns $(f(x),V^\top\nabla f(x))$ at cost $1+\operatorname{rank}V$, and a
call with exact values only has $V=0$. A deterministic algorithm maps each
finite transcript to its next call or to an output point; the output is
unrestricted and uncharged. $N_{\rm det}$ is the smallest worst-case cost of
an algorithm that returns an $\eps$-stationary point on every function of the
class. A randomized algorithm draws its seed from a law fixed before the
function and chooses each call by a policy that is jointly measurable in the
seed and the transcript. It must return an $\eps$-stationary point with
probability at least $2/3$ on each function of the class; the number of calls
is not bounded in advance, and a run that does not terminate counts as a
failure with infinite cost. $N_{\rm rand}$ is the smallest cap on the cost of
all runs, and $N^{\rm exp}_{\rm rand}$ is the smallest worst-case expected
cost. The class $\cB^d_{L,\Delta,E}$ is defined as in~\eqref{eq:main-class}:
one fixed $H\succeq0$ with $\norm H_{\rm op}\le L$ and
$\operatorname{tr}H\le LE$ dominates the derivative of $\nabla f$ almost
everywhere on $\{f<f(0)\}$. Lemma~\ref{a:lem:hessian-segment}, which
identifies this class with the segment form used in the proofs, is formalized
as well. The class $\cF^d_{L,\Delta,E}$ is formalized through the quadratic
upper model~\eqref{eq:trace-class} on all of $\R^d$.

\paragraph{Formalized statements.}
Let $d\ge1$, $L,\Delta,\eps>0$, $1\le E\le d$, and $T\ge1$. For the value
oracle and for the rank-priced oracle, $N_{\rm det}(\cB^d_{L,\Delta,E})$ is
finite and satisfies
\begin{gather*}
 d\le N_{\rm det}\le\min\bigl\{10dT,\;100\sqrt{dET(d+T)}\bigr\},\\
 ET\le d\ \Longrightarrow\
 \frac d{2250}\min\Bigl\{T,\sqrt{\frac{ET}{\log(e+T)}}\Bigr\}
 \le N_{\rm det}\le100\sqrt2\,d\sqrt{ET},\\
 E\ge T\log(e+T)\ \Longrightarrow\ N_{\rm det}\ge\frac{dT}{2359296},
 \qquad
 T\log(e+T)\le E\le\frac dT\ \Longrightarrow\ N_{\rm det}\ge\frac{dT}{2250}.
\end{gather*}
For randomized algorithms and both oracles, $N_{\rm rand}(\cB^d_{L,\Delta,E})$
is finite, and
\begin{gather*}
 N_{\rm rand}(\cF^d_{L,\Delta,E})\ge\frac{\min\{d,ET\}}{12},\qquad
 N^{\rm exp}_{\rm rand}(\cF^d_{L,\Delta,E})\ge\frac{\min\{d,ET\}}{144},\\
 \frac{\min\{d,ET\}}{12}\le N_{\rm rand}(\cB^d_{L,\Delta,E})\le23ET,
\end{gather*}
where the upper bound is attained by the three-point method of
Theorem~\ref{b:thm:stp}. If $T\log(e+T)\le E\le d/T$, then for the same oracle
\[
 \frac d{51750\,E}\le\frac{N_{\rm det}(\cB^d_{L,\Delta,E})}{N_{\rm rand}(\cB^d_{L,\Delta,E})}\le\frac{120\,d}E.
\]
These statements imply Theorems~\ref{thm:main-random} and~\ref{thm:main-det}
and Corollary~\ref{cor:main-window} on $\cB^d_{L,\Delta,E}$, whose constants
are not specified. The randomized lower bounds have the constants of
Corollary~\ref{cor:quadratic-baseline}. The formalized deterministic lower
bounds use more conservative constants than Theorem~\ref{a:thm:main}
($1/2250$ in place of $1/672$), including the constant $9$ in place of $13/3$
in the selection Lemma~\ref{a:lem:selection} and cruder chain and wrapper
bounds.

\paragraph{Not formalized.}
The lower bound of Theorem~\ref{thm:linear-lower}, the
variant~\eqref{u:eq:unknown} without $E$ or $H$, the three-point bound on
$\cA^d_{L,\Delta,E}$, the coercivity, nonconvexity, and Lipschitz properties
of the hard functions, the deterministic lower bounds on
$\cF^d_{L,\Delta,E}$, and the results of Appendices~\ref{app:further}
and~\ref{rb:sec:main} are not formalized.

\end{document}